%% file: main_0416.tex
\documentclass{article}

\PassOptionsToPackage{numbers, compress}{natbib}
\usepackage[main, final]{neurips_2026}

\usepackage[utf8]{inputenc} % allow utf-8 input
\usepackage[T1]{fontenc}    % use 8-bit T1 fonts
\usepackage{hyperref}       % hyperlinks
\usepackage{url}            % simple URL typesetting
\usepackage{booktabs}       % professional-quality tables
\usepackage{amsfonts}       % blackboard math symbols
\usepackage{nicefrac}       % compact symbols for 1/2, etc.
\usepackage{microtype}      % microtypography
\usepackage{subcaption}
\usepackage{wrapfig}
\usepackage{multirow}
\usepackage{makecell}
\usepackage{array}
\usepackage{graphicx}
\usepackage{amsmath}
\usepackage{bbm}
\usepackage{tabularx}
\usepackage[table]{xcolor}
\usepackage{rotating}

\usepackage[normalem]{ulem}
\newcommand{\bestm}[1]{$\mathbf{#1}$}
\newcommand{\secondm}[1]{$\underline{#1}$}

\title{A Locally Tokenized Generative Model \\for Robust Time-Series Watermarking}

\author{%
  Dongbin Kim\textsuperscript{1 \dag},\quad Geonwoo Shin\textsuperscript{1 \dag},\quad Yujin Choi\textsuperscript{1 2},\quad Soyeon Park\textsuperscript{1},\quad Jaewook Lee\textsuperscript{1 *} \\[1ex]
  \textsuperscript{1}Seoul National University \qquad \textsuperscript{2}Nanyang Technological University \\[1ex]
  \texttt{\{dongbin413,shin0621,uznhigh,soyeon2,jaewook\}@snu.ac.kr}
}

\begin{document}

\maketitle

\renewcommand{\thefootnote}{\fnsymbol{footnote}}
\footnotetext[2]{Equal contribution.}
\footnotetext[1]{Corresponding author.}
\renewcommand{\thefootnote}{\arabic{footnote}}
\setcounter{footnote}{0}

\begin{abstract}
Watermarking is a central tool for provenance in generative models, yet its application to multivariate time series remains hindered by reliability failures under post-editing attacks. We show that existing detectors, which rely on globally coupled re-encoding, suffer from bidirectional drift of the null distribution: post-editing attacks can shift the $z$-score of non-watermarked samples in either direction, invalidating clean-calibrated thresholds. We argue that this instability is a property of the re-encoding, and that reliable detection requires each recovered unit to depend only on a bounded temporal neighborhood. Guided by this principle, we propose L-VQVAE, a generative model in which each discrete token is produced from a short contiguous window, and LVQMark, a watermarking method over this token space that combines logit-bias injection with robust re-encoding for attack-time detection. Experiments on four benchmarks spanning finance, energy, and neuroimaging show that our approach preserves generation quality while stabilizing both detection power and false-positive behavior under post-editing attacks.

\end{abstract}

\section{Introduction}

Reliable provenance verification is essential as generative models increasingly produce high-quality synthetic data~\citep{brown2020language, rombach2022high, yuan2024diffusionts}. Among existing approaches---provenance metadata, forensic detection, model fingerprinting, and watermarking~\citep{england2021amp, ricker2024aeroblade, yu2021artificial}---generation-time watermarking is particularly suited to synthetic time series~\citep{kirchenbauer2023watermark, soi2025timewak}: it embeds verification signals during sampling, requires no retraining or external metadata, and remains detectable under downstream transformations~\citep{wen2023tree, yang2024gaussian}.

%문단 2 : 제기한 문제점
Despite these advances, most watermarking methods assess robustness through detectability of watermarked samples after post-editing attack, typically reporting true positive rate (TPR) at a fixed false positive rate (FPR) (TPR@X\%FPR) calibrated on clean negative samples~\cite{zhaoprovable, wen2023tree, soi2025timewak, sanderwatermark}. Yet for provenance verification, false positives are also critical: attacked non-watermarked samples may be misdetected as watermarked, leading to erroneous attribution and reduced reliability of the provenance guarantee. While recent image watermarking and forgery-aware studies discuss such false-attribution risks~\cite{muller2025black, arabi2025seal}, they remain largely unexplored in time-series watermarking.

We address this gap by enforcing \emph{locality} in the recovered token representation. We first introduce \textbf{L-VQVAE}, a generative model for multivariate time series in which each discrete token is recovered from a bounded temporal neighborhood of the observed signal, rather than through global inversion. L-VQVAE comprises a local tokenizer, a global decoder, and an autoregressive transformer, all sharing a single codebook. We then build \textbf{LVQMark}, a generation-time watermarking method designed over this locally recoverable token interface: it embeds a red--green logit bias during autoregressive sampling, where green-set tokens receive a positive sampling bias at each position, and introduces a robust encoder that maps attacked continuous signals back to their clean token assignments for detection. Because generation, insertion, and detection are all defined over the same locally recoverable token representation, perturbations affect only a bounded subset of recovered evidence, preserving both detection power and false-positive control.

We evaluate LVQMark on four time-series datasets under varying sequence lengths and post-editing attacks. Experiments show that LVQMark preserves high generation quality while achieving robust watermark detection and stable false-positive behavior on attacked non-watermarked samples.

Our main contributions are summarized as follows:
\begin{itemize}
\item We formulate false-positive reliability in time-series watermarking 
as a \emph{re-encoding stability} problem. We show that when the re-encoding 
map is globally coupled, post-editing attacks can shift the null distribution 
of the detection statistic for non-watermarked samples, invalidating 
clean-calibrated thresholds.
\item We introduce L-VQVAE, a generative model for multivariate time series in which each token depends only on a short temporal neighborhood, preventing perturbations from inducing global drift in the recovered representation.
\item We build LVQMark, a generation-time watermarking method over the L-VQVAE token space, combining red--green logit-bias insertion with scheduling modifications tailored to L-VQVAE and robust re-encoding for attack-time detection.
\end{itemize}

\section{Watermarking in Time Series and Robustness}
\label{sec:preliminary}

\paragraph{Generalized detection framework.}
Generation-time watermarks \cite{kirchenbauer2023watermark, huo2024token, soi2025timewak} share a common statistical template. Let $\mathbf{x}\in\mathbb{R}^{T\times D}$ denote an observed sample, let $\mathcal{X}$ denote the space of recovered units (e.g.\ a finite codebook index set), and let $\Phi(\mathbf{x})\in\mathcal{X}^N$ denote the internal representation recovered by the detector. For each position $n\in\{1,\dots,N\}$, a secret key $\kappa$ specifies a position-wise target region $\mathcal{T}_n\subset\mathcal{X}$. Under the null hypothesis $H_0$, the recovered units $\{\Phi(\mathbf{x})_n\}_{n=1}^{N}$ are independent across positions and each lies in $\mathcal{T}_n$ with probability $\mu_0$; watermarking biases the generative process so that this event occurs more frequently. The detector computes the agreement rate and its standardized score
\begin{equation}
\hat{g}(\mathbf{x}) = \frac{1}{N}\sum_{n=1}^{N} \mathbbm{1}\!\left[\Phi(\mathbf{x})_n \in \mathcal{T}_n\right],
\qquad
z(\mathbf{x}) = \frac{\hat{g}(\mathbf{x}) - \mu_0}{\sigma_0},
\qquad
\sigma_0 = \sqrt{\tfrac{\mu_0(1-\mu_0)}{N}},
\label{eq:zscore}
\end{equation}
and declares $\mathbf{x}$ watermarked whenever $z(\mathbf{x})>\eta$ for a fixed threshold $\eta>0$. In the balanced setting considered throughout this paper, $\mu_0=1/2$, so that $\sigma_0=\tfrac{1}{2\sqrt{N}}$. This formulation subsumes both token-level generation-time watermarks for language models, such as KGW~\cite{kirchenbauer2023watermark}, and time-series methods such as TimeWak~\cite{soi2025timewak}: although these approaches adopt different choices of the recovered representation $\Phi$ and target regions $\mathcal{T}_n$, they share an identical statistical structure.

\paragraph{Red-green watermarking and time-series departure.}
KGW~\cite{kirchenbauer2023watermark} recovers $\Phi(\mathbf{x})$ trivially from the observed token sequence: detection simply reproduces the green list $G_n$ from the secret seed $\kappa$ and applies Equation~\eqref{eq:zscore} directly. Time-series watermarking~\cite{soi2025timewak} adheres to the same statistical logic but departs in one essential respect: the detector must first re-encode the continuous signal into $\Phi(\mathbf{x})$. This re-encoding step is benign under clean conditions, but, as we show next, becomes the primary source of robustness degradation under post-editing attacks.

\begin{table}[h]
\vspace{-0.8em}
\centering
\caption{Intended interpretation of $z$ under a clean-calibrated
detector. The validity of this interpretation requires re-encoding
stability of $\Phi$ under the attack family; in its absence,
null drift bidirectionally invalidates the threshold.}
\label{tab:zscore_cases}
\begin{tabular}{lccc}
\toprule
 & $z\gg 0$ & $z\approx 0$ & $z\ll 0$ \\
\midrule
Watermarked     & Correct detection & Miss               & Statistical anomaly \\
Non-watermarked & False positive    & Correct rejection  & Statistical anomaly \\
\bottomrule
\end{tabular}
\end{table}

\paragraph{Re-encoding stability.}
To formalize this failure mode, we call a re-encoding map $\Phi$ \emph{stable} under an attack family $\mathcal{F}$ when, for every non-watermarked
$\mathbf{x}$ and every $\mathcal{A}\in\mathcal{F}$, the distribution
of $\Phi(\mathcal{A}(\mathbf{x}))$ stays close to that of
$\Phi(\mathbf{x})$, keeping
$\mathbb{E}_{\mathcal{H}_0}[\hat{g}(\mathcal{A}(\mathbf{x}))]$ near
$\mu_0$. We refer to violations of this property as \emph{re-encoding
instability}. Its observable symptom is the drift of
$\mathbb{E}[\hat{g}(\mathcal{A}(\mathbf{x}))]$ away from $\mu_0$ on
non-watermarked samples, which we call \emph{null drift}. We refer to
the property that the clean-calibrated threshold continues to control
the FPR under attack as \emph{false-positive reliability}. Under this
view, false-positive reliability is a direct consequence of
re-encoding stability.

\begin{figure}[t]
    \centering
    \begin{subfigure}[t]{0.54\linewidth}
        \centering
        \includegraphics[width=\linewidth]{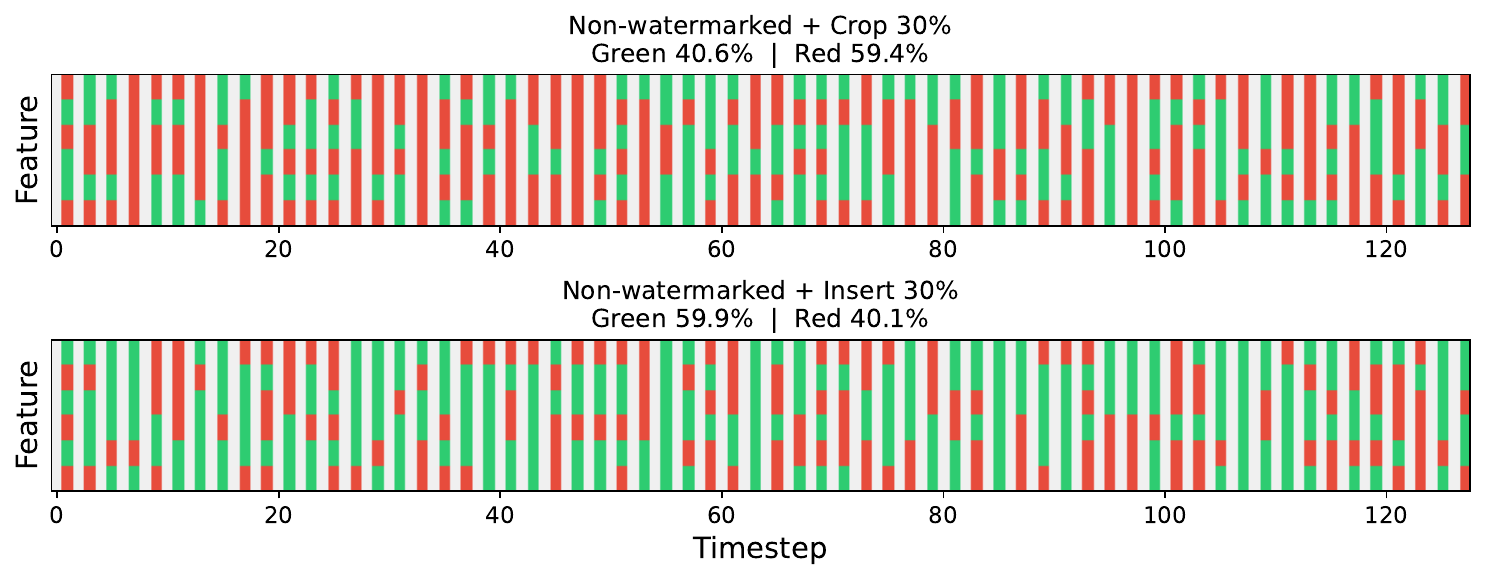}%
        \caption{Recovered green/red assignments.}
    \end{subfigure}
    \hfill
    \begin{subfigure}[t]{0.45\linewidth}
        \centering
        \includegraphics[width=\linewidth]{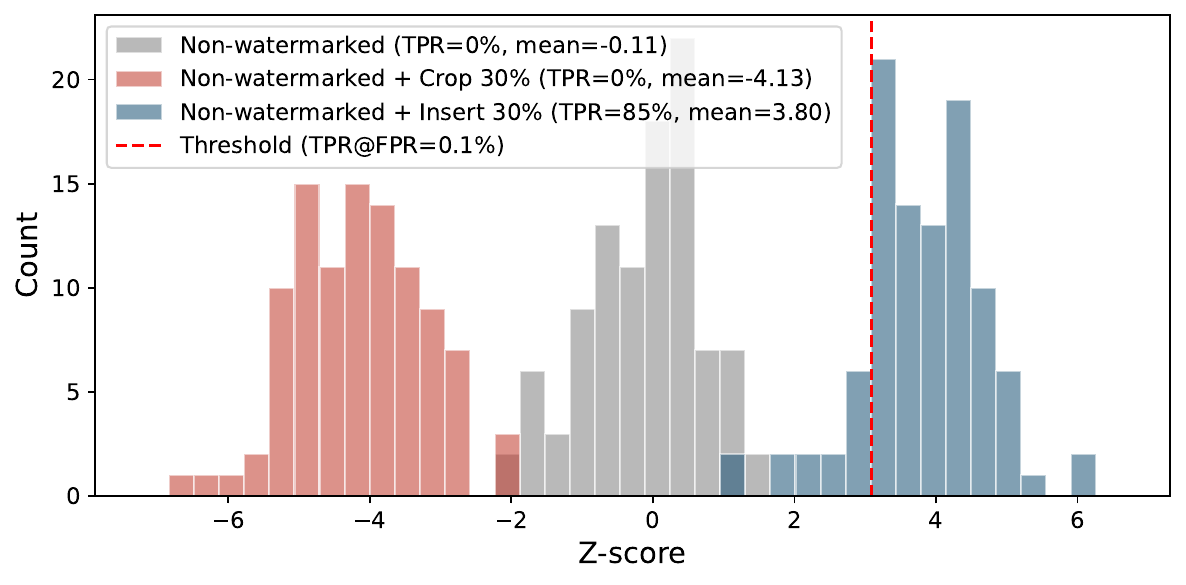}
        \caption{$z$-score distributions.}
        \label{fig:nonwm_drift_zscore}
    \end{subfigure}
    \caption{Bidirectional drift of the null on \emph{non-watermarked} samples
    (TimeWak~\cite{soi2025timewak}, Stocks). Crop and insert push the recovered tokens (a) and the resulting
    $z$-score (b) in opposite directions, with insert crossing the clean-calibrated
    threshold and yielding an $85\%$ false-positive rate.}
    \label{fig:nonwm_drift}
\end{figure}

\paragraph{Bidirectional null drift under post-editing attacks.}
Existing methods fail to satisfy this property. As Figure~\ref{fig:nonwm_drift} illustrates on a representative method and Table~\ref{tab:main_128} confirms across baselines, $\hat{g}$ concentrates near $\mu_0=1/2$ in the clean setting (as Equation~\eqref{eq:zscore} requires) but drifts away under attack in a method- and attack-dependent manner. Since the drift direction is governed by the interaction between $\Phi$ and $\mathcal{A}$ rather than by the watermark, attacks can either trigger false positives ($z\gg 0$) or induce severe \emph{statistical anomalies} ($z\ll 0$), and the clean-calibrated threshold loses its intended meaning under both regimes of Table~\ref{tab:zscore_cases}.

\paragraph{A representation-level interpretation.}
We attribute null drift to $\Phi$ rather than to the statistic itself: existing methods instantiate $\Phi$ through \emph{globally coupled} computations, where each recovered position depends on essentially every time step of $\mathbf{x}$, so a localized perturbation propagates to every position. We address this in Section~\ref{sec:method} by making each unit of $\Phi$ depend only on a bounded temporal neighborhood---a property we call \emph{locality}---confining attacks to a bounded subset of recovered units.

\section{Proposed Method}
\label{sec:method}

\begin{figure}[t!]
    \centering
    \includegraphics[width=\textwidth]{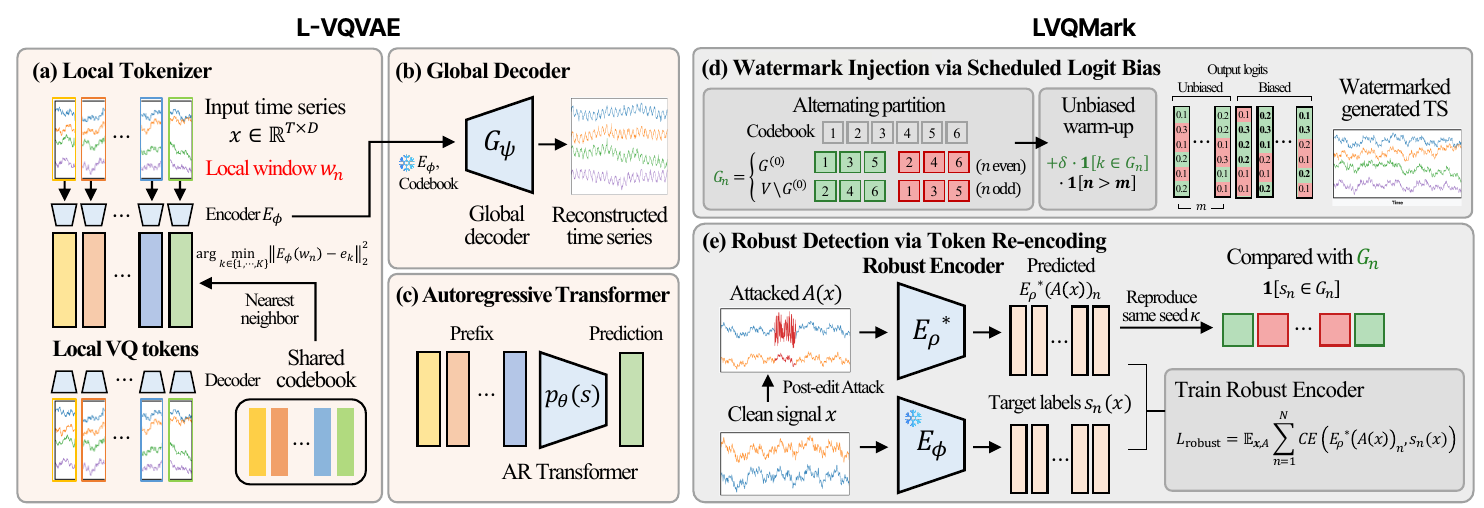}
    \caption{Overview of L-VQVAE and LVQMark. \textbf{(a)} The local tokenizer encodes each temporal window into a discrete VQ token. \textbf{(b)} The global decoder reconstructs the time series from the token sequence. \textbf{(c)} An autoregressive transformer models the token distribution. \textbf{(d)} Watermark injection via scheduled logit bias with alternating partition and unbiased warm-up. \textbf{(e)} Robust detection via token re-encoding, where $E_\rho^*$ recovers clean token assignments from the attacked signal for the green-ratio test.}

    \label{fig:overview}
\end{figure}

Motivated by the observation that globally coupled representations propagate post-editing attacks beyond the attacked region, we design a generative model with a deliberately 
\emph{local} internal representation (\textbf{L-VQVAE}) and a watermarking method that exploits 
this locality (\textbf{LVQMark}). A schematic is shown in Figure~\ref{fig:overview}.

\subsection{L-VQVAE: Locally Tokenized Generative Model}
\label{sec:lvqvae}
 
L-VQVAE generates multivariate time series through three stages: a local tokenizer encodes each short temporal window into a discrete codebook entry, a global decoder reconstructs a full time series from the resulting token sequence, and an autoregressive transformer models the distribution over token sequences for sampling.
 
\subsubsection{Local Tokenization}
\label{sec:local-tokenizer}

Let $\mathbf{x}\in\mathbb{R}^{T\times D}$ denote a multivariate time series with $T$ time steps and $D$ variables. The local tokenizer produces a discrete token sequence $\mathbf{s}=(s_1,\dots,s_N)\in\{1,\dots,K\}^N$ from $\mathbf{x}$ in three steps: sliding-window decomposition, per-window summarization, and vector quantization against a shared codebook $\mathcal{C}=\{\mathbf{e}_k\}_{k=1}^K$.

\paragraph{Sliding-window decomposition.} We partition $\mathbf{x}$ into overlapping local windows with receptive field $R$ and stride $s$ (with $s$ chosen so that $(T-R)/s$ is an integer):
\begin{equation}
\mathbf{w}_n = \mathbf{x}_{t_n+1:\,t_n+R} \in \mathbb{R}^{R\times D}, \qquad t_n = (n-1)\,s, \qquad n=1,\dots,N,
\end{equation}
where $N=(T-R)/s+1$ is the number of windows. By construction, $\mathbf{w}_n$ depends only on the time steps within $[t_n+1,\,t_n+R]$, so a temporal post-editing attack overlaps at most $\lceil R/s \rceil + 1$ consecutive windows and can therefore alter at most that many tokens downstream.

\paragraph{Window summarization and quantization.} Each window is summarized into a single latent vector by a shared cross-attention encoder $E_\phi$ and then quantized against the codebook. We linearly project $\mathbf{w}_n$ to dimension $d$ and add learnable positional embeddings $\mathbf{P}\in\mathbb{R}^{R\times d}$ (shared across windows) to obtain $\mathbf{H}_n\in\mathbb{R}^{R\times d}$. A single learnable query vector $\mathbf{q}_0\in\mathbb{R}^{1\times d}$, shared across all windows, is then refined by a stack of $L$ cross-attention blocks: for $\ell=1,\dots,L$,
\begin{align}
\tilde{\mathbf{q}}^{(\ell)}_n &= \mathrm{LN}\!\left(\mathbf{q}^{(\ell-1)}_n + \mathrm{CrossAttn}(\mathbf{q}^{(\ell-1)}_n,\mathbf{H}_n,\mathbf{H}_n)\right), \\
\mathbf{q}^{(\ell)}_n &= \mathrm{LN}\!\left(\tilde{\mathbf{q}}^{(\ell)}_n + \mathrm{FFN}(\tilde{\mathbf{q}}^{(\ell)}_n)\right),
\end{align}
with $\mathbf{q}^{(0)}_n = \mathbf{q}_0$ for all $n$. The output $\mathbf{q}^{(L)}_n\in\mathbb{R}^{1\times d}$ is squeezed and linearly projected to obtain the window latent $E_\phi(\mathbf{w}_n)\in\mathbb{R}^{d_c}$, which is mapped to its nearest codebook entry:
\begin{equation}
s_n = \arg\min_{k\in\{1,\dots,K\}}\|E_\phi(\mathbf{w}_n)-\mathbf{e}_k\|_2^2, \qquad \mathbf{z}^q_n = \mathbf{e}_{s_n},
\end{equation}
with gradients propagated through the $\arg\min$ by the straight-through estimator~\cite{van2017neural}. Because keys and values are derived exclusively from $\mathbf{w}_n$, any post-editing attack of $\mathbf{x}$ that does not overlap $\mathbf{w}_n$ leaves $E_\phi(\mathbf{w}_n)$ and $s_n$ unchanged. For brevity, we write $\mathbf{z}_n := E_\phi(\mathbf{w}_n)$ for the continuous (pre-quantization) encoding of window $\mathbf{w}_n$, and $\mathbf{z}^q_n$ for its quantized counterpart.

\paragraph{Local pretraining.} Training $(E_\phi,\mathcal{C})$ jointly with the global decoder from scratch leads to poor codebook utilization: the global decoder can compensate for inaccurate codes by drawing on neighboring tokens, weakening the gradient signal to the codebook and undermining the per-token locality on which detection-time recovery relies. We therefore first pretrain $(E_\phi,\mathcal{C})$ together with a \emph{local} decoder $D^{\mathrm{loc}}_{\tilde\psi}$ that reconstructs $\mathbf{w}_n$ from the single code $\mathbf{z}^q_n$ alone. The encoder and local decoder are trained by minimizing
\begin{equation}
\mathcal{L}_{\text{VQ}} = \mathbb{E}_n\!\left[\,\underbrace{\|\mathbf{w}_n - D^{\mathrm{loc}}_{\tilde\psi}(\mathbf{z}^q_n)\|_2^2}_{\text{reconstruction}} \;+\; \beta \underbrace{\|\mathbf{z}_n - \mathrm{sg}[\mathbf{z}^q_n]\|_2^2}_{\text{commitment}}\,\right],
\end{equation}
where $\mathrm{sg}[\cdot]$ denotes the stop-gradient operator and $\mathbf{z}_n = E_\phi(\mathbf{w}_n)$. The codebook entries $\{\mathbf{e}_k\}_{k=1}^{K}$ are not updated by gradient descent; instead, each $\mathbf{e}_k$ is maintained as an exponential moving average of the encoder outputs $\mathbf{z}_n$. Under this local bottleneck, each code must fully describe its window without relying on neighboring context. After pretraining, $D^{\mathrm{loc}}_{\tilde\psi}$ is discarded, and $(E_\phi,\mathcal{C})$ are frozen for the subsequent training of the global decoder.

\subsubsection{Global Decoder and Autoregressive Transformer}
\label{sec:decoder-prior}
 
Locality is imposed only on the encoding side, which is the sole component re-executed at detection time and therefore exposed to post-editing attacks. Since the decoder and AR Transformer are invoked only during generation, they need not be local and can model long-range dependencies without compromising robustness.
 
\paragraph{Global decoder.} The decoder $G_\psi$ maps the full token sequence to a time series in a single forward pass. Let $\mathbf{E}\in\mathbb{R}^{N\times d}$ denote the embedded token sequence with positional embeddings, and let $\mathbf{Q}_t\in\mathbb{R}^{T\times d}$ be a learnable set of time-step queries. A stack of cross-attention blocks updates $\mathbf{Q}_t$ using $\mathbf{E}$ as keys and values,
\begin{equation}
\mathbf{Q}_t \leftarrow \mathrm{LN}(\mathbf{Q}_t + \mathrm{CrossAttn}(\mathbf{Q}_t,\mathbf{E},\mathbf{E})), \qquad \mathbf{Q}_t \leftarrow \mathrm{LN}(\mathbf{Q}_t + \mathrm{FFN}(\mathbf{Q}_t)),
\end{equation}
and a linear projection yields $\hat{\mathbf{x}} = G_\psi(\mathbf{s}) \in \mathbb{R}^{T\times D}$. $G_\psi$ is trained with $E_\phi$ and the codebook frozen, under $\mathcal{L}_{\text{dec}} = \|\mathbf{x}-G_\psi(\mathbf{s}(\mathbf{x}))\|_2^2$.
 
\paragraph{Autoregressive Transformer.} The token sequence is modeled by a decoder-only autoregressive transformer, $p_\theta(\mathbf{s})=\prod_{n=1}^{N} p_\theta(s_n\mid s_{<n})$, trained with next-token cross-entropy. Causal self-attention is restricted to a fixed lookback window, consistent with the locality principle of the tokenizer. Together, the local tokenizer, the global decoder, and the autoregressive transformer constitute L-VQVAE as a standalone generative model: sampling from $p_\theta$ and decoding through $G_\psi$ produces a synthetic time series without any watermark.

\subsection{LVQMark: Watermarking over Local VQ Tokens}
\label{sec:lvqmark}
 
LVQMark embeds a watermark into the L-VQVAE generation process by biasing the autoregressive sampling step, and provides a detection mechanism based on robust re-encoding.

\subsubsection{Watermark Injection via Scheduled Logit Bias}
\label{sec:watermark-injection}

\paragraph{Logit-bias watermarking.}
At each autoregressive position $n$, a secret seed $\kappa$ partitions the codebook $\mathcal{V}=\{1,\dots,K\}$ into a green subset $G_n\subseteq\mathcal{V}$ and its complementary red subset $R_n = \mathcal{V}\setminus G_n$, with $|G_n|=|R_n|=K/2$. Let $\boldsymbol{\ell}_n\in\mathbb{R}^{K}$ denote the logits produced by the autoregressive transformer at position $n$. A positive bias $\delta>0$ is added to the logits of green tokens before sampling:
\begin{equation}
\tilde{\ell}_{n,k} = \ell_{n,k} + \delta\,\mathbbm{1}[k\in G_n], \qquad s_n \sim \mathrm{Categorical}(\mathrm{softmax}(\tilde{\boldsymbol{\ell}}_n)),
\label{eq:logit-bias}
\end{equation}
This adapts the red--green scheme of KGW~\citep{kirchenbauer2023watermark} (Section~\ref{sec:preliminary}) to the L-VQVAE codebook. At detection time, the observed time series is re-encoded into tokens by the robust encoder $E_\rho^\star$ (Section~\ref{sec:robust-encoder}), the green sets $\{G_n\}$ are reconstructed from the same seed $\kappa$, and the green-ratio test of Section~\ref{sec:preliminary} is applied with target regions $\mathcal{T}_n = G_n$.

\paragraph{Context-independent partition.}
In KGW for language models, the green set $G_n$ typically depends on the preceding token~$s_{n-1}$, which diversifies the partition across positions. We instead fix the partition independently of the recovered token context. The reason is locality: if $G_n$ depended on recovered tokens, a single re-encoding error at position~$n$ would propagate to the partition at position~$n{+}1$, coupling re-encoding instability with watermark scheduling. A context-independent rule confines each re-encoding error to the position where it occurs.

\paragraph{Failure modes of a na\"ive fixed partition.}
A context-independent partition interacts with two structural properties of L-VQVAE token sequences. First, adjacent windows largely overlap, so the AR Transformer frequently emits runs of the same token; a position-invariant $G_n$ then applies $+\delta$ uniformly across the run, driving the green ratio toward~$1$ or~$0$ regardless of watermark presence (\emph{bias accumulation}). Second, at the first few autoregressive steps the context is empty and logits are nearly uniform, so a fixed $\delta$ causes near-deterministic selection of the same green token, producing a stereotyped prefix that conditions all subsequent steps (\emph{bias dominance}).

\paragraph{Alternating partition.}
To address bias accumulation, we make the partition depend on position parity. Let $G^{(0)}\subset\mathcal{V}$ be a fixed green set determined by the secret seed $\kappa$, with $|G^{(0)}|=K/2$, and define
\begin{equation}
  G_n =
  \begin{cases}
    G^{(0)}, & n \text{ even},\\
    \mathcal{V}\setminus G^{(0)}, & n \text{ odd},
  \end{cases}
  \qquad
  R_n = \mathcal{V}\setminus G_n.
\end{equation}
Under this rule, a token $k$ is biased ($+\delta$) at one parity and unbiased at the other. Consequently, when the autoregressive model emits a run of identical tokens, the bias is applied at only half of the positions in the run, rather than uniformly across the entire run. This bounds the contribution of any single token value to the green-ratio statistic and prevents repeated-token runs from saturating it toward $1$ or $0$. Since the detector reconstructs $G_n$ from the same seed and parity rule, no detection power is lost.

\paragraph{Unbiased warm-up.}
To address bias dominance at early steps, we suppress the logit bias for the first $m$ positions (we use $m=3$). Combining this with the alternating partition above, the watermarked logits become
\begin{equation}
  \tilde{\ell}_{n,k} = \ell_{n,k} + \delta\cdot\mathbbm{1}[k\in G_n]\cdot\mathbbm{1}[n>m],
  \qquad
  s_n \sim \mathrm{Categorical}(\mathrm{softmax}(\tilde{\boldsymbol{\ell}}_n)),
\end{equation}
so that positions $n\le m$ are sampled from the unmodified autoregressive distribution. The detector excludes these same positions when computing the green-ratio statistic, so no detection signal is lost: at steps where the autoregressive context is empty, $\delta$ would have caused near-deterministic selection of a green token regardless of input, providing no usable evidence for the test.

\subsubsection{Robust Detection via Token Re-encoding}
\label{sec:robust-encoder}
Local tokenization confines the effect of a post-editing attack to a bounded set of tokens, but does not guarantee that those tokens are recovered to their clean codebook entries: the clean encoder $E_\phi$ may reassign perturbed windows to different codes. We therefore introduce a \emph{robust encoder} $E_\rho^\star$ whose objective is to recover, from an attacked input $A(\mathbf{x})$, the token assignments that $E_\phi$ would have produced on $\mathbf{x}$. The two components are complementary: without local tokenization, a globally coupled representation propagates the post-editing attack to every recovered token, leaving $E_\rho^\star$ no stable reference to recover (Table~\ref{tab:robust_encoder_64}); without $E_\rho^\star$, locally affected tokens remain misclassified.

$E_\rho^\star$ shares the windowed cross-attention architecture of $E_\phi$ but replaces vector quantization with a per-token classifier over the codebook vocabulary. It is trained on pairs $(\mathbf{x},A(\mathbf{x}))$ with $\mathbf{x}$ sampled from L-VQVAE and $A$ drawn from a fixed family of attacks (including the identity), excluding pairs in which $A$ removes the entire span of some window:
\begin{equation}
\mathcal{L}_{\text{robust}} = \mathbb{E}_{\mathbf{x},A}\sum_{n=1}^{N}\mathrm{CE}\!\left(E_\rho^\star(A(\mathbf{x}))_n,\; s_n(\mathbf{x})\right),
\end{equation}
where $s_n(\mathbf{x})$ are the codes produced by the frozen $E_\phi$ on the clean input. Joint training over multiple attack types yields a single encoder that recovers clean assignments without specializing to any one perturbation. At detection time, $E_\rho^\star$ replaces $E_\phi$ in the green-ratio test of Section~\ref{sec:preliminary}.

\section{Experiments}
\label{sec:exp}

\subsection{Experimental Setup}

\paragraph{Datasets and baselines.}
We evaluate LVQMark on four multivariate time-series datasets: Stocks~\cite{yoon2019time}, ETTh~\cite{zhou2021informer}, Energy~\cite{candanedo2017appliances}, and fMRI~\cite{smith2011network}, allowing us to assess the generality of the proposed method under heterogeneous temporal patterns. We compare against representative watermarking baselines, including Tree-Ring (TR)~\cite{wen2023tree}, Gaussian Shading (GS)~\cite{yang2024gaussian}, and TimeWak~\cite{soi2025timewak}, which cover both context-independent and context-dependent watermarking strategies. To analyze watermarking performance together with backbone choice, we consider diffusion- and VQ-based generative backbones: DiffusionTS~\cite{yuan2024diffusionts} for diffusion-based watermarking methods, and TimeVQVAE~\cite{lee2023vector} and SDFormer~\cite{chen2024sdformer} as VQ-based generative baselines. TimeVQVAE uses discrete latent tokens with a transformer prior, while SDformer is a recent VQ-based model designed for strong time-series generation quality. 

\paragraph{Attack setting and metrics.} 

We evaluate robustness under three post-editing attacks---\textit{offset}, \textit{crop}, and \textit{insertion}---at two strengths, 5\% and 30\%. The main paper reports results for the stronger 30\% setting, while the 5\% results are deferred to Appendix~\ref{app:mild_attack}. We assess generation quality using Context-FID~\cite{jeha2022psa}, Correlational~\cite{liao2020conditional}, Discriminative~\cite{yoon2019time}, and Predictive score~\cite{yoon2019time}. For watermark reliability, we treat watermarked samples as the positive class and report TPR@0.1\%FPR, where the threshold is calibrated on clean non-watermarked samples to yield 0.1\% FPR, and TPR is measured on watermarked samples at test time. To directly evaluate false-positive stability under post-editing attack, we additionally report the empirical FPR on \textit{attacked} non-watermarked samples under the same clean-calibrated detector.

All reported results are averaged over five runs conducted with fixed random seeds to ensure reproducibility. Additional details on datasets, preprocessing, attack implementation, metrics, and hyperparameters are deferred to Appendix~\ref{app:exp_details}.

\subsection{Watermark Detection and Generation Quality}

\begin{table*}[t!]
\centering
\setlength{\tabcolsep}{3pt}
\renewcommand{\arraystretch}{1.13}
\caption{Results of synthetic time series watermark detection and quality. Watermarked (TPR) and Non-watermarked (FPR, mean $z$-score) detections are evaluated under 30\% attacks. Quality metrics are for 64-length. Best results are in bold, and second-best are underlined. $z$-scores with $|z|>3.09$ are marked in \textcolor{red}{red}.}
\label{tab:main_64}
\resizebox{\textwidth}{!}{%
\begin{tabular}{c|l|l|ccc|cr|cr|cr|cccc}
\hline
\hline
\rowcolor{gray!20}
\multicolumn{3}{c|}{\textbf{Setting}}
& \multicolumn{3}{c|}{\textbf{Watermark (TPR $\uparrow$)}}
& \multicolumn{6}{c|}{\textbf{Non-watermarked (FPR $\downarrow$ | $z$-score)}}
& \multicolumn{4}{c}{\textbf{Quality Metric ($\downarrow$)}} \\
\hline
\rowcolor{gray!20}
\multicolumn{1}{c|}{\textbf{Dataset}}& \multicolumn{1}{c|}{\textbf{Model}} & \multicolumn{1}{c|}{\textbf{Method}}
& \textbf{Offset} & \textbf{Crop} & \textbf{Insert}
& \multicolumn{2}{c|}{\textbf{Offset}}
& \multicolumn{2}{c|}{\textbf{Crop}}
& \multicolumn{2}{c|}{\textbf{Insert}}
& \textbf{C-FID} & \textbf{Corr.} & \textbf{Disc.} & \textbf{Pred.} \\
\hline

\multirow{5}{*}{{Stocks}}
& \multirow{3}{*}{DiffusionTS}
  & TR
  & 0.00 & \bestm{1.00} & \bestm{1.00}
  & \bestm{0.00} & $+0.21$
  & 1.00 & \textcolor{red}{$+69.43$}
  & 1.00 & \textcolor{red}{$+8.11$}
  & 1.52 & 0.07 & 0.15 & \bestm{0.04} \\
& & GS
  & \bestm{1.00} & 0.00 & \bestm{1.00}
  & \bestm{0.00} & $-1.40$
  & \bestm{0.00} & \textcolor{red}{$-14.87$}
  & 0.05 & $+1.52$
  & 1.50 & \secondm{0.02} & 0.23 & \bestm{0.04} \\
& & TimeWak
  & \bestm{1.00} & \bestm{1.00} & \bestm{1.00}
  & \bestm{0.00} & $-0.19$
  & \bestm{0.00} & \textcolor{red}{$-4.32$}
  & 0.05 & $+1.35$
  & 0.29 & \bestm{0.01} & 0.13 & \bestm{0.04} \\
\cline{2-16}
& SDformer  & LVQMark
  & 0.00 & 0.01 & 0.00
  & \bestm{0.00} & $-0.16$
  & 0.02 & $+1.19$
  & \bestm{0.00} & $+0.31$
  & \secondm{0.08} & \bestm{0.01} & \bestm{0.01} & \bestm{0.04} \\
\cline{2-16}
& L-VQVAE   & LVQMark
  & \bestm{1.00} & \bestm{1.00} & \bestm{1.00}
  & \bestm{0.00} & $-0.12$
  & 0.01 & $-0.05$
  & 0.01 & $+0.35$
  & \bestm{0.07} & \secondm{0.02} & \secondm{0.06} & \bestm{0.04} \\
\hline
\hline
\multirow{5}{*}{{ETTh}}
& \multirow{3}{*}{DiffusionTS}
  & TR
  & \bestm{1.00} & \bestm{1.00} & \bestm{1.00}
  & \bestm{0.00} & $+0.89$
  & 1.00 & \textcolor{red}{$+17.16$}
  & 1.00 & \textcolor{red}{$+38.55$}
  & 2.17 & 0.22 & 0.29 & \secondm{0.14} \\
& & GS
  & \bestm{1.00} & \bestm{1.00} & \bestm{1.00}
  & \bestm{0.00} & $-1.53$
  & \bestm{0.00} & \textcolor{red}{$-5.44$}
  & \bestm{0.00} & \textcolor{red}{$-5.82$}
  & 3.43 & 0.25 & 0.36 & 0.16 \\
& & TimeWak
  & \bestm{1.00} & 0.61 & \bestm{1.00}
  & 0.04 & $+1.17$
  & \bestm{0.00} & $-0.28$
  & 0.03 & $+1.20$
  & 0.37 & 0.13 & \secondm{0.11} & \bestm{0.12} \\
\cline{2-16}
& SDformer  & LVQMark
  & 0.00 & 0.00 & 0.00
  & \bestm{0.00} & $-0.94$
  & 0.02 & $+0.09$
  & \bestm{0.00} & $-0.09$
  & \secondm{0.04} & \bestm{0.05} & \bestm{0.01} & \bestm{0.12} \\
\cline{2-16}
& L-VQVAE   & LVQMark
  & \bestm{1.00} & \bestm{1.00} & \bestm{1.00}
  & 0.01 & $-0.12$
  & \bestm{0.00} & $-0.22$
  & \bestm{0.00} & $+0.35$
  & \bestm{0.03} & \secondm{0.06} & \bestm{0.01} & \bestm{0.12} \\
\hline
\hline

\multirow{5}{*}{{Energy}}
& \multirow{3}{*}{DiffusionTS}
  & TR
  & 0.00 & \bestm{1.00} & \bestm{1.00}
  & \bestm{0.00} & $+1.18$
  & 1.00 & \textcolor{red}{$+53.73$}
  & 1.00 & \textcolor{red}{$+58.55$}
  & 0.58 & 1.98 & 0.43 & \secondm{0.28} \\
& & GS
  & \bestm{1.00} & 0.38 & \bestm{1.00}
  & 1.00 & \textcolor{red}{$+14.51$}
  & \bestm{0.00} & $+0.87$
  & 1.00 & \textcolor{red}{$+11.77$}
  & 1.78 & 2.72 & 0.48 & 0.31 \\
& & TimeWak
  & \bestm{1.00} & 0.97 & \bestm{1.00}
  & \bestm{0.00} & \textcolor{red}{$-9.69$}
  & \bestm{0.00} & $-0.42$
  & \bestm{0.00} & $-1.57$
  & \secondm{0.14} & 1.52 & \secondm{0.14} & \bestm{0.25} \\
\cline{2-16}
& SDformer  & LVQMark
  & \bestm{1.00} & 0.02 & 0.58
  & \bestm{0.00} & $-0.02$
  & \bestm{0.00} & $+0.34$
  & \bestm{0.00} & $+0.34$
  & \bestm{0.04} & \secondm{1.05} & \bestm{0.08} & \bestm{0.25} \\
\cline{2-16}
& L-VQVAE   & LVQMark
  & \bestm{1.00} & \bestm{1.00} & \bestm{1.00}
  & \bestm{0.00} & $-0.06$
  & \bestm{0.00} & $-0.25$
  & 0.01 & $+0.04$
  & \bestm{0.04} & \bestm{0.95} & 0.15 & \bestm{0.25} \\
\hline
\hline

\multirow{5}{*}{{fMRI}}
& \multirow{3}{*}{DiffusionTS}
  & TR
  & \bestm{1.00} & 0.02 & \bestm{1.00}
  & \bestm{0.00} & $+0.05$
  & 1.00 & $+2.27$
  & 1.00 & \textcolor{red}{$+12.14$}
  & 3.63 & 12.83 & 0.40 & 0.14 \\
& & GS
  & \bestm{1.00} & \bestm{1.00} & \bestm{1.00}
  & \bestm{0.00} & $-1.78$
  & \bestm{0.00} & \textcolor{red}{$-32.25$}
  & 0.06 & $-0.01$
  & 0.74 & 8.31 & 0.50 & 0.11 \\
& & TimeWak
  & \bestm{1.00} & \bestm{1.00} & \bestm{1.00}
  & \bestm{0.00} & $+0.08$
  & \bestm{0.00} & $-2.74$
  & \bestm{0.00} & $+0.18$
  & \secondm{0.45} & 1.87 & \secondm{0.25} & \secondm{0.10} \\
\cline{2-16}
& SDformer  & LVQMark
  & 0.25 & 0.02 & 0.04
  & \bestm{0.00} & $+0.03$
  & 0.02 & $+0.51$
  & \bestm{0.00} & $-0.40$
  & \bestm{0.13} & \secondm{1.19} & \bestm{0.12} & \bestm{0.09} \\
\cline{2-16}
& L-VQVAE   & LVQMark
  & \bestm{1.00} & \bestm{1.00} & \bestm{1.00}
  & \bestm{0.00} & $-0.03$
  & \bestm{0.00} & $-0.24$
  & \bestm{0.00} & $-0.17$
  & \bestm{0.13} & \bestm{1.14} & 0.27 & \bestm{0.09} \\
\hline
\hline
\end{tabular}%
}
\end{table*}

\paragraph{Stable false-positive behavior under post-editing attacks.}
\label{sec:exp-drift}

Table~\ref{tab:main_64} reveals two distinct failure modes among existing baselines.
\textit{Spurious false positives.} Tree-Ring under crop produces mean $z$ values of $+69.4$ on Stocks, $+17.2$ on ETTh, and $+53.7$ on Energy, driving the non-watermarked FPR to $1.00$; Gaussian Shading shows the same pattern under offset on Energy ($z=+14.5$, FPR $=1.00$). The attacked null has drifted past the clean-calibrated threshold $\eta$, causing non-watermarked samples to be systematically misattributed.
\textit{Severe negative drift.} Other baselines drift in the opposite direction: Gaussian Shading reaches $z=-32.2$ on fMRI under crop, and TimeWak reaches $z=-9.7$ on Energy under offset. These scores stay below $\eta$ and yield zero empirical FPR, but the drift magnitude shows the null has been distorted well beyond its nominal scale---calibration here is accidental, not principled.
FPR alone is therefore insufficient: a detector can appear calibrated when drift falls on the negative side while remaining unreliable in absolute terms. By contrast, L-VQVAE with LVQMark keeps the mean attacked $z$ within $[-0.25, +0.35]$ across all twelve (dataset, attack) configurations, with FPR $\leq 0.01$ and TPR $=1.00$ in eleven of twelve cases.

\paragraph{L-VQVAE achieves generation quality comparable to state-of-the-art backbones.}
Table~\ref{tab:quality_64} shows that L-VQVAE remains highly competitive in generation quality, achieving the best or second-best result on most metrics across the four datasets. By comparison, TimeVQVAE performs substantially worse on several datasets. TimeVQVAE uses separate pathways to model coarse temporal structure and fine local detail. Although this helps represent different scales of variation, cross-variable interactions are handled less explicitly, which may limit generation quality on datasets with strong inter-variable dependencies, such as Energy and fMRI.

\begin{table*}[h!]
\centering
\scriptsize
\setlength{\tabcolsep}{3pt}
\renewcommand{\arraystretch}{1.13}
\caption{Results of synthetic time series quality for 64-length sequences. Best results are in bold, and second-best are underlined.}
\label{tab:quality_64}
\resizebox{\textwidth}{!}{%
\begin{tabular}{l|cccc|cccc|cccc|cccc}
\hline
\hline
\rowcolor{gray!20}
 & \multicolumn{4}{c|}{\textbf{Stocks}} & \multicolumn{4}{c|}{\textbf{Energy}} & \multicolumn{4}{c|}{\textbf{ETTh}} & \multicolumn{4}{c}{\textbf{fMRI}} \\
\rowcolor{gray!20}
\hline
\textbf{Model} & \textbf{C-FID} & \textbf{Corr.} & \textbf{Disc.} & \textbf{Pred.} & \textbf{C-FID} & \textbf{Corr.} & \textbf{Disc.} & \textbf{Pred.} & \textbf{C-FID} & \textbf{Corr.} & \textbf{Disc.} & \textbf{Pred.} & \textbf{C-FID} & \textbf{Corr.} & \textbf{Disc.} & \textbf{Pred.} \\
\hline
TimeVQVAE   & 0.439           & 0.079           & 0.368           & 0.056           & 5.423           & 7.189           & 0.466           & 0.293           & 2.432           & 0.238           & 0.453           & 0.244           & 18.643          & 38.410          & 0.275           & 0.255 \\
DiffusionTS & 0.294           & 0.013           & 0.126           & \secondm{0.037} & 0.143           & 1.519           & 0.135           & \secondm{0.251} & 0.370           & 0.134           & 0.114           & 0.117           & 0.446           & 1.869           & 0.253           & 0.100 \\
SDformer    & \secondm{0.067} & \bestm{0.007}   & \secondm{0.043} & \bestm{0.036}   & \bestm{0.029}   & \secondm{1.009} & \bestm{0.081}   & \bestm{0.248}   & \secondm{0.030} & \bestm{0.052}   & \bestm{0.008}   & \bestm{0.112}   & \bestm{0.089}   & \secondm{1.129} & \bestm{0.114}   & \secondm{0.084} \\
\hline
L-VQVAE     & \bestm{0.039}   & \secondm{0.008} & \bestm{0.018}   & \bestm{0.036}   & \secondm{0.033} & \bestm{0.955}   & \secondm{0.127} & \bestm{0.248}   & \bestm{0.024}   & \secondm{0.062} & \secondm{0.010} & \secondm{0.115} & \secondm{0.096} & \bestm{1.118}   & \secondm{0.163} & \bestm{0.083} \\
\hline
\hline
\end{tabular}%
}
\end{table*}

\subsection{Component Analysis and Ablations}

\paragraph{Both scheduling components are necessary.}
\label{sec:exp-abl}

\begin{figure}[h!]
\centering
\begin{subfigure}[b]{0.48\textwidth}
  \centering
  \includegraphics[width=\textwidth]{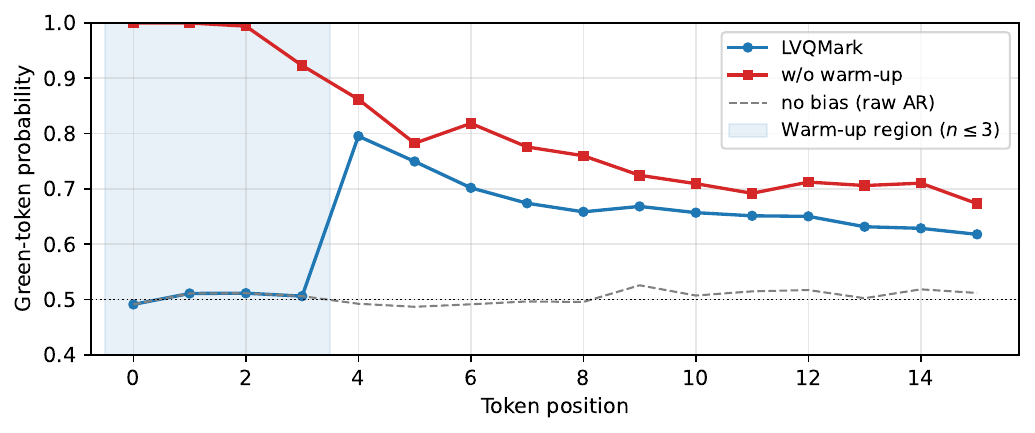}
  \caption{ETTh}
  \label{fig:green_prob_etth}
\end{subfigure}
\hfill
\begin{subfigure}[b]{0.48\textwidth}
  \centering
  \includegraphics[width=\textwidth]{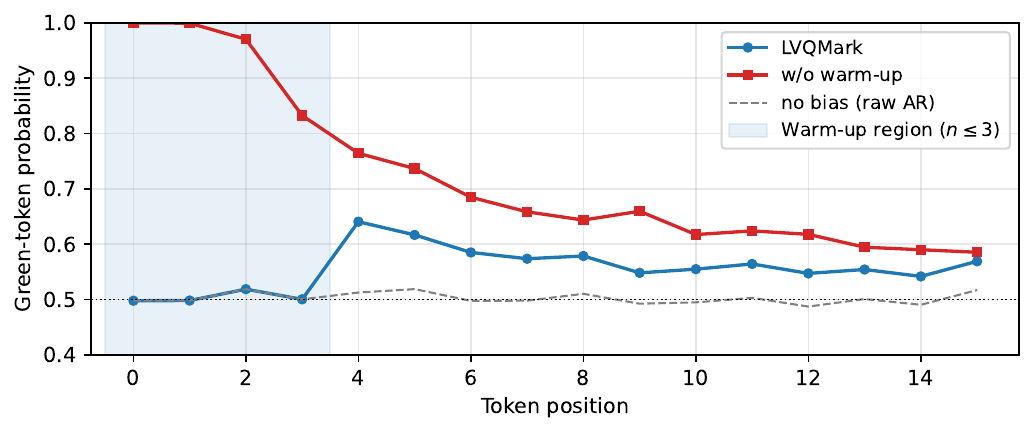}
  \caption{Energy}
  \label{fig:green_prob_energy}
\end{subfigure}
\caption{Per-position green-token probability ($T{=}64$, $\delta{=}10$). Shaded region: warm-up positions ($n \leq 3$) where the logit bias is suppressed.}
\label{fig:green_prob}
\end{figure}

We ablate the two scheduling modifications introduced in Section~\ref{sec:watermark-injection}: the alternating partition and the unbiased warm-up. Figure~\ref{fig:green_prob} shows the per-position green-token probability with and without the warm-up. Without it, the first few positions are sampled near-deterministically from the green set, creating a stereotyped prefix that propagates into later steps. The warm-up suppresses the bias during positions $n \leq m$, keeping early tokens close to the unbiased baseline and allowing the autoregressive context to diversify before the watermark signal is introduced.

Table~\ref{tab:sampling_method_64} quantifies the impact on generation quality. Removing the warm-up alone degrades Context-FID moderately; removing both modifications together leads to further deterioration across most datasets, confirming that the two components address complementary failure modes.

\begin{table*}[h!]
\centering
\setlength{\tabcolsep}{3pt}
\renewcommand{\arraystretch}{1.13}
\caption{Ablation of watermark scheduling components for 64-length sequences (LVQMark). \textsc{no-warmup}: unbiased warm-up removed. \textsc{no-alt-warmup}: both alternating partition and unbiased warm-up removed. Best results are in bold, and second-best are underlined.}
\label{tab:sampling_method_64}
\resizebox{\textwidth}{!}{%
\begin{tabular}{l|cccc|cccc|cccc|cccc}
\hline
\hline
\rowcolor{gray!20}
 & \multicolumn{4}{c|}{\textbf{Stocks}} & \multicolumn{4}{c|}{\textbf{Energy}} & \multicolumn{4}{c|}{\textbf{ETTh}} & \multicolumn{4}{c}{\textbf{fMRI}} \\
\rowcolor{gray!20}
\hline
\textbf{Variant} & \textbf{C-FID} & \textbf{Corr.} & \textbf{Disc.} & \textbf{Pred.} & \textbf{C-FID} & \textbf{Corr.} & \textbf{Disc.} & \textbf{Pred.} & \textbf{C-FID} & \textbf{Corr.} & \textbf{Disc.} & \textbf{Pred.} & \textbf{C-FID} & \textbf{Corr.} & \textbf{Disc.} & \textbf{Pred.} \\
\hline
\textsc{no-alt-warmup} & 0.222           & \bestm{0.007}   & \secondm{0.094} & \bestm{0.036}   & 0.114           & 1.060           & 0.203           & \secondm{0.250} & 0.172           & 0.098           & 0.027           & 0.119           & 0.324           & 1.362           & \secondm{0.199} & 0.093 \\
\textsc{no-warmup}      & \secondm{0.194} & 0.020           & 0.107           & \bestm{0.036}   & \secondm{0.071} & \bestm{0.924}   & \secondm{0.181} & \bestm{0.248}   & \secondm{0.061} & \bestm{0.060}   & \bestm{0.007}   & \bestm{0.114}   & \secondm{0.255} & \secondm{1.186} & \bestm{0.170}   & \secondm{0.089} \\
LVQMark                 & \bestm{0.068}   & \secondm{0.015} & \bestm{0.055}   & \bestm{0.036}   & \bestm{0.035}   & \secondm{0.945} & \bestm{0.151}   & \bestm{0.248}   & \bestm{0.028}   & \secondm{0.061} & \secondm{0.010} & \secondm{0.116} & \bestm{0.133}   & \bestm{1.135}   & 0.267           & \bestm{0.086} \\
\hline
\hline
\end{tabular}%
}
\end{table*}

Across the four datasets, LVQMark consistently achieves the best or second-best quality on most metrics. Removing the unbiased warm-up alone degrades Context-FID and Correlational scores moderately, while removing both modifications together leads to further deterioration, particularly on ETTh, Energy, and fMRI. This incremental pattern confirms that the two modifications address complementary failure modes: the alternating partition mitigates the statistical artifact induced by repeated-token runs, while the unbiased warm-up prevents the stereotyped prefix from distorting downstream generation. We adopt the full configuration as the default in all main experiments. Additional ablations for other sequence lengths are provided in Appendix~\ref{app:abl-short}.

\begin{table*}[h!]
\centering
\setlength{\tabcolsep}{3pt}
\renewcommand{\arraystretch}{1.13}
\caption{Results of watermark detection under attack for 64-length sequences. LVQMark is applied to different variants, and watermark detection performance (TPR) is evaluated under 30\% attacks. Best results are in bold, and second-best are underlined.}
\label{tab:robust_encoder_64}
\resizebox{\textwidth}{!}{%
\begin{tabular}{c|c|ccc|ccc|ccc|ccc}
\hline
\hline
\rowcolor{gray!20}
 &  & \multicolumn{3}{c|}{\textbf{Stocks}} & \multicolumn{3}{c|}{\textbf{Energy}} & \multicolumn{3}{c|}{\textbf{ETTh}} & \multicolumn{3}{c}{\textbf{fMRI}} \\
\rowcolor{gray!20}
\hline
\textbf{Model} & \textbf{Type} & \textbf{Offset} & \textbf{Crop} & \textbf{Insert} & \textbf{Offset} & \textbf{Crop} & \textbf{Insert} & \textbf{Offset} & \textbf{Crop} & \textbf{Insert} & \textbf{Offset} & \textbf{Crop} & \textbf{Insert} \\
\hline
\multirow{2}{*}{SDformer} & w/o Robust & \secondm{0.00} & \secondm{0.01} & \secondm{0.00} & \bestm{1.00} & 0.02 & 0.58 & 0.00 & 0.00 & 0.00 & \bestm{1.00} & 0.02 & \bestm{1.00} \\
 & w/ Robust & \secondm{0.00} & 0.00 & 0.00 & \bestm{1.00} & \secondm{0.66} & \secondm{0.80} & \secondm{0.13} & \secondm{0.08} & \secondm{0.03} & \secondm{0.05} & \secondm{0.28} & \secondm{0.03} \\
\hline
\multirow{2}{*}{L-VQVAE} & w/o Robust & \bestm{1.00} & 0.00 & \bestm{1.00} & \bestm{1.00} & 0.00 & \bestm{1.00} & \bestm{1.00} & 0.01 & \bestm{1.00} & \bestm{1.00} & \bestm{1.00} & \bestm{1.00} \\
 & w/ Robust & \bestm{1.00} & \bestm{1.00} & \secondm{0.98} & \bestm{1.00} & \bestm{1.00} & \bestm{1.00} & \bestm{1.00} & \bestm{1.00} & \bestm{1.00} & \bestm{1.00} & \bestm{1.00} & \bestm{1.00} \\
\hline
\hline
\end{tabular}%
}
\end{table*}

% 2. robust encoder를 일반 vqvae에 적용했을 때 성능이 나오는가 / table
\paragraph{Robust encoder requires local tokenization.}
\label{sec:local_tokenization}
Table~\ref{tab:robust_encoder_64} compares SDformer and L-VQVAE with and without the attack-augmented robust encoder used in our framework. The robust encoder is itself effective: paired with L-VQVAE, it lifts Crop detection from near-zero to near-perfect across all datasets while preserving Offset and Insert. With SDformer, however, gains are uneven—substantial on Energy, marginal on ETTh, absent on Stocks—and on fMRI the encoder actively \emph{degrades} detection, collapsing Offset from $1.00$ to $0.05$ and Insert from $1.00$ to $0.03$. Robust detection thus hinges not on attack-aware training alone, but on a local representation that supports stable per-region recovery.
Additional experiments for other sequence lengths are provided in Appendix~\ref{app:short}.

\begin{figure}[h!]
    \centering
    \begin{subfigure}{0.24\textwidth}
        \centering
        \includegraphics[width=\linewidth]{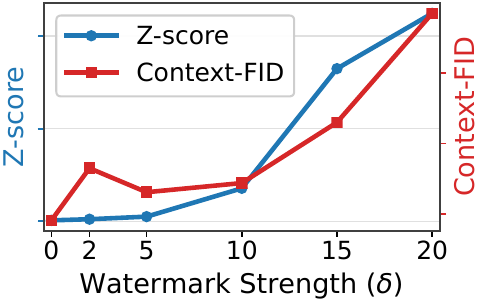}
        \caption{Stocks}
    \end{subfigure}\hfill
    \begin{subfigure}{0.24\textwidth}
        \centering
        \includegraphics[width=\linewidth]{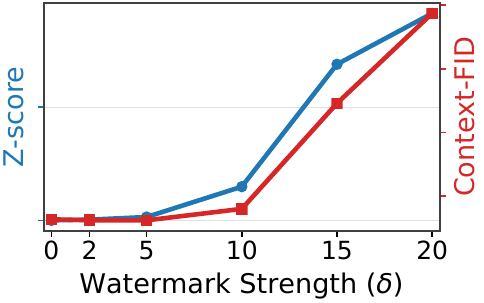}
        \caption{Energy}
    \end{subfigure}\hfill
    \begin{subfigure}{0.24\textwidth}
        \centering
        \includegraphics[width=\linewidth]{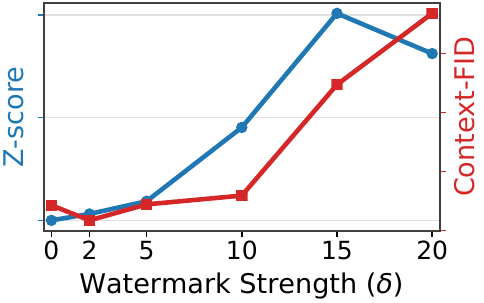}
        \caption{ETTh}
    \end{subfigure}\hfill
    \begin{subfigure}{0.24\textwidth}
        \centering
        \includegraphics[width=\linewidth]{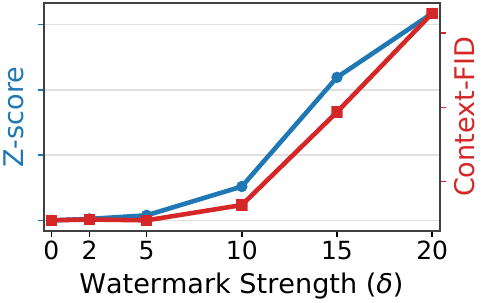}
        \caption{fMRI}
    \end{subfigure}
    \caption{$z$-score and Context-FID across watermark strength $\delta$ for 64-length}
    \label{fig:strength}
\end{figure}

\paragraph{Detectability–quality trade-off in $\delta$.}
Figure~\ref{fig:strength} shows how watermark strength $\delta$ affects the trade-off between watermark detectability and generation quality under our LVQMark, where $\delta=0$ denotes the non-watermarked baseline. As $\delta$ increases, watermark evidence becomes substantially stronger across Energy, ETTh, and fMRI, while Context-FID remains relatively stable at moderate strengths and increases more clearly only at larger values. This pattern indicates a clear trade-off: stronger bias improves detectability, but excessive bias can harm sample quality. % In practice, $\delta=10$ provides a favorable balance for Energy, ETTh, and fMRI, delivering strong watermark evidence while keeping quality close to the non-watermarked baseline. For Stocks, we use a larger value, $\delta=20$, in the final configuration.

\begin{table*}[h!]
\centering
\setlength{\tabcolsep}{3pt}
\renewcommand{\arraystretch}{1.13}
\caption{Effect of the receptive field $R$ of L-VQVAE under the 30\% crop attack for 64-length sequences. Windows are non-overlapping ($s=R$) and the codebook size and token count are fixed; $R=4$ is the default. Best results are in bold. $z$-scores with $|z|>3.09$ are marked in \textcolor{red}{red}.}
\label{tab:rf}
\resizebox{\textwidth}{!}{%
\begin{tabular}{c|ccccc|cr|cr|cr|cr|cr}
\hline
\hline
\rowcolor{gray!20}
& \multicolumn{5}{c|}{\textbf{WM (TPR $\uparrow$)}} & \multicolumn{10}{c}{\textbf{No-WM (FPR $\downarrow$ | $z$)}} \\
\hline
\rowcolor{gray!20}
\multicolumn{1}{c|}{\textbf{Dataset}} & $R{=}4$ & $R{=}8$ & $R{=}16$ & $R{=}32$ & $R{=}64$
& \multicolumn{2}{c|}{$R{=}4$} & \multicolumn{2}{c|}{$R{=}8$} & \multicolumn{2}{c|}{$R{=}16$} & \multicolumn{2}{c|}{$R{=}32$} & \multicolumn{2}{c}{$R{=}64$} \\
\hline
Stocks & \bestm{1.00} & 0.00 & 0.00 & 0.00 & \bestm{1.00}
& 0.01 & $-0.05$ & \bestm{0.00} & \textcolor{red}{$-7.65$} & \bestm{0.00} & $-1.48$ & \bestm{0.00} & $+0.04$ & 1.00 & \textcolor{red}{$+8.93$} \\
ETTh & \bestm{1.00} & \bestm{1.00} & \bestm{1.00} & \bestm{1.00} & 0.00
& \bestm{0.00} & $-0.22$ & 0.17 & $+2.06$ & \bestm{0.00} & $+0.54$ & \bestm{0.00} & $+0.71$ & \bestm{0.00} & \textcolor{red}{$-6.01$} \\
Energy & \bestm{1.00} & \bestm{1.00} & \bestm{1.00} & \bestm{1.00} & \bestm{1.00}
& \bestm{0.00} & $-0.25$ & \bestm{0.00} & $-0.43$ & \bestm{0.00} & $-1.63$ & 0.03 & $+1.49$ & 0.96 & \textcolor{red}{$+4.91$} \\
fMRI & \bestm{1.00} & \bestm{1.00} & \bestm{1.00} & 0.00 & 0.01
& \bestm{0.00} & $-0.24$ & 0.95 & \textcolor{red}{$+4.76$} & \bestm{0.00} & $-0.32$ & \bestm{0.00} & $-2.52$ & \bestm{0.00} & $-0.11$ \\
\hline
\hline
\end{tabular}%
}
\end{table*}

\paragraph{Sensitivity to the receptive field.} Table~\ref{tab:rf} varies the receptive field $R$ from the default of 4 to 64 within L-VQVAE, with non-overlapping windows, a fixed codebook size and token count, and all other settings unchanged (Appendix~\ref{app:rf}), so that locality is the only design axis that moves. Two degradation patterns appear as $R$ grows: the attacked null drifts away from zero, in several settings far enough to raise the false-positive rate, and the watermark evidence weakens until detection fails. Only $R=4$ satisfies both objectives on all four datasets, and the practical rule is to keep $R$ small relative to $T$: locality binds only the encoder, while long-range structure is handled by the global decoder and the autoregressive transformer at generation time.

\subsection{Deployment Cost}
\label{sec:cost}

\begin{wraptable}{r}{0.46\textwidth}
\vspace{-13pt}
\centering
\setlength{\tabcolsep}{4pt}
\renewcommand{\arraystretch}{1.13}
\caption{Generation and verification cost on ETTh ($T=64$), single GPU, batch size 1 (memory at 64).}
\label{tab:cost}
\small
\begin{tabular}{l|cc}
\hline
\hline
\rowcolor{gray!20}
\multicolumn{1}{c|}{\textbf{Metric}} & \textbf{TimeWak} & \textbf{Ours} \\
\hline
Gen. throughput (samples/s) & 350 & \bestm{4{,}922} \\
Verif. FLOPs (GFLOP) & 9.59 & \bestm{1.83} \\
Verif. GPU latency (ms) & 1,010 & \bestm{0.72} \\
Verif. peak memory (MB) & 781 & \bestm{259} \\
\hline
\hline
\end{tabular}
\vspace{-8pt}
\end{wraptable}
Table~\ref{tab:cost} compares the recurring cost of LVQMark with TimeWak on ETTh; the full measurements, including the higher one-time training cost of LVQMark, are in Appendix~\ref{app:cost}. Autoregressive sampling generates an order of magnitude faster than diffusion sampling, and verification is a single forward pass of the local encoder followed by a counting test, three orders of magnitude faster than the sequential diffusion inversion of TimeWak. This profile suits the intended deployment, in which generation is a one-time operation at the data provider and verification is the recurring one, performed at high volume and often on modest hardware.

\subsection{Robustness Beyond Local Edits}
\label{sec:beyond_local}

The attacks of Table~\ref{tab:main_64} are shift, mask, and replace operations. Table~\ref{tab:nonlocal} evaluates two transformations that alter every window at once, Gaussian noise and smoothing (Appendix~\ref{app:attacks}), with the original robust encoder and with one retrained on both attacks. The original encoder transfers to every setting except Gaussian noise on Stocks, with a small loss on ETTh that retraining the encoder alone closes. The non-watermarked $z$-score stays near zero and the false-positive rate at or below 0.01 throughout, so null stability extends to attacks that touch every window. Stocks under Gaussian noise fails with either encoder: financial series have a low signal-to-noise ratio~\cite{liu2022finrl}, so additive noise obscures the local patterns on which token recovery relies, while false positives remain controlled.

\begin{table*}[h!]
\centering
\setlength{\tabcolsep}{3pt}
\renewcommand{\arraystretch}{1.13}
\caption{Results of watermark detection under non-local attacks for 64-length sequences. Gaussian noise ($\alpha=0.3$) and smoothing ($w=3$) perturb every time step. Orig.: the robust encoder of the main experiments. Retr.: retrained with both attacks added to its augmentation set, with the generator and the watermark schedule unchanged. Best results are in bold.}
\label{tab:nonlocal}
\resizebox{\textwidth}{!}{%
\begin{tabular}{c|c|cc|cr|cr||c|cc|cr|cr}
\hline
\hline
\rowcolor{gray!20}
& & \multicolumn{2}{c|}{\textbf{WM (TPR $\uparrow$)}} & \multicolumn{4}{c||}{\textbf{No-WM (FPR $\downarrow$ | $z$)}}
& & \multicolumn{2}{c|}{\textbf{WM (TPR $\uparrow$)}} & \multicolumn{4}{c}{\textbf{No-WM (FPR $\downarrow$ | $z$)}} \\
\hline
\rowcolor{gray!20}
\multicolumn{1}{c|}{\textbf{Enc.}} & \multicolumn{1}{c|}{\textbf{Dataset}} & \textbf{Noise} & \textbf{Smooth} & \multicolumn{2}{c|}{\textbf{Noise}} & \multicolumn{2}{c||}{\textbf{Smooth}}
& \multicolumn{1}{c|}{\textbf{Dataset}} & \textbf{Noise} & \textbf{Smooth} & \multicolumn{2}{c|}{\textbf{Noise}} & \multicolumn{2}{c}{\textbf{Smooth}} \\
\hline
Orig. & \multirow{2}{*}{Stocks} & \bestm{0.02} & \bestm{1.00} & \bestm{0.00} & $-0.83$ & \bestm{0.00} & $-0.06$
      & \multirow{2}{*}{ETTh}   & 0.93 & \bestm{1.00} & \bestm{0.00} & $-0.07$ & \bestm{0.00} & $+0.05$ \\
Retr. & & 0.00 & \bestm{1.00} & \bestm{0.00} & $-0.24$ & \bestm{0.00} & $-0.53$
      & & \bestm{1.00} & \bestm{1.00} & 0.01 & $-0.17$ & \bestm{0.00} & $-0.22$ \\
\hline
Orig. & \multirow{2}{*}{Energy} & \bestm{1.00} & \bestm{1.00} & \bestm{0.00} & $-0.07$ & \bestm{0.00} & $-0.21$
      & \multirow{2}{*}{fMRI}   & \bestm{1.00} & \bestm{1.00} & \bestm{0.00} & $-0.02$ & \bestm{0.00} & $-0.06$ \\
Retr. & & \bestm{1.00} & \bestm{1.00} & \bestm{0.00} & $-0.06$ & \bestm{0.00} & $+0.08$
      & & \bestm{1.00} & \bestm{1.00} & \bestm{0.00} & $-0.01$ & \bestm{0.00} & $-0.02$ \\
\hline
\hline
\end{tabular}%
}
\end{table*}

\section{Conclusion and Limitations}
In this work, we argued that false-positive reliability in time-series watermarking is fundamentally a re-encoding stability problem. Under post-editing attacks, watermark detection must first re-encode the observed continuous signal into an internal token representation, and when this re-encoding is globally coupled, post-editing attacks can distort the recovered evidence far beyond the attacked region. Motivated by this failure mode, we introduced L-VQVAE, a generative model in which each token depends only on a short temporal neighborhood, and built LVQMark on top of this locally recoverable interface with robust re-encoding and generation-time watermark injection. Across four datasets, L-VQVAE with LVQMark preserves competitive generation quality while substantially improving false-positive stability under post-editing attacks.
 
\paragraph{Limitations.} Our experiments cover four multivariate time-series datasets and three sequence lengths; extending the evaluation to additional domains and longer horizons is a natural direction for future work. Robustness is assessed under three local post-editing attacks at two strength levels and two non-local transformations. Finally, the multi-stage training pipeline involves more steps than single-stage baselines, though each stage is standard and can be executed independently.

\section*{Acknowledgments}
This work was supported by the Basic Science Research Program through the National Research Foundation of Korea (NRF) funded by the Ministry of Education (No. RS-2025-25420430). This work was also partly supported by the NRF grant funded by the Korean government (MSIT) (No. RS-2024-00338859), and by the Rural Development Administration of Korea (No. RS-2026-25504956).

\bibliography{references}
\bibliographystyle{plainnat}

%%%%%%%%%%%%%%%%%%%%%%%%%%%%%%%%%%%%%%%%%%%%%%%%%%%%%%%%%%%%
\newpage
\appendix

\section{Related Work}
\label{app:extended_related_work}

\paragraph{Discrete interfaces for time-series generation.}
Vector Quantization (VQ) \cite{van2017neural} provides a discrete interface for continuous signals by mapping encoder outputs to entries in a learned codebook. The resulting code indices can be decoded back to the signal domain or modeled by a sequence prior. TimeVQVAE~\cite{lee2023vector} and SDformer~\cite{chen2024sdformer} show that such discrete token-based generators can be competitive with continuous-domain time-series generators~\cite{yuan2024diffusionts} and preserve information useful for downstream classification or forecasting. However, these representations are not designed for re-encoding stability, or robustness to post-processing corruptions. These properties are essential for watermark detection, where a modified continuous signal must be re-encoded into a consistent token sequence before the watermark statistic can be computed.

\paragraph{Token-level generation-time watermarking.}
Generation-time watermarking is attractive because it embeds the watermark by steering the sampling process itself, rather than by directly perturbing the generated sample. This distinction is important because post-processing-based watermarking can directly alter generated samples and degrade quality~\cite{yang2024gaussian}; in time-series generation, such perturbations may further distort temporal or cross-variate dependencies that are central to synthetic-data utility~\cite{yoon2019time, yuan2024diffusionts}. The red-green watermarking framework~\cite{kirchenbauer2023watermark} is a standard template for token-level generation-time watermarking in language models. At each sampling step, it partitions the vocabulary into a \textit{green} set and its complement \textit{red} set using a secret key, and adds a positive logit bias to the tokens in green set. This alters the sampling distribution, so that generated token sequences contain a detectable statistical bias. Subsequent methods~\cite{huo2024token, huunbiased, wu2026an} refine this mechanism by adapting the bias to token-level context or by designing unbiased rules that better preserve the original sampling distribution. We build on this distributional view, but instantiate it for time series over a local VQ token space. In contrast, our work focuses on detector-side re-encoding stability and false-positive behavior on attacked non-watermarked samples.

\section{Experiment Details}
\label{app:exp_details}
\subsection{Implementation Details}
Tables~\ref{tab:hp-lvqvae}--\ref{tab:hp-vqvae} summarize the hyperparameters used for each model.
All models share the same data splits and preprocessing across datasets for fair comparison. Regarding the random replacement probability, during AR Transformer training each input token is independently replaced with a random codebook index with probability $p$, while ground-truth labels remain unchanged. This forces the model to predict the next token correctly even under corrupted context, preventing over-reliance on preceding tokens and improving generalization. For the standard VQ-VAE, which attends over the full token history, we set $p=0.3$ for ETTh and Stocks — datasets with stronger temporal autocorrelation — and $p=0.1$ for Energy and fMRI. For Local VQ-VAE, the AR Transformer restricts attention to a fixed lookback window of 6 tokens, which structurally limits long-range dependency; accordingly, $p=0.1$ suffices across all datasets.

\begin{table}[h!]
\centering
\caption{L-VQVAE hyperparameter settings.
  Stride 2 is used for window $T{=}24$; stride 4 for $T{=}64,128$.
  \textsuperscript{$\dagger$}\,Robust encoder is trained with $\textbf{crop}_\text{var}$.}
\label{tab:hp-lvqvae}
\begin{tabular}{lcccc}
\toprule
Parameter & Energy & ETTh & fMRI & Stocks \\
\midrule
\multicolumn{5}{l}{\textit{Encoder / Decoder}} \\
Feature size $D$           & 28    & 7     & 50    & 6   \\
$d_\text{model}$           & 512   & 512   & 512   & 128 \\
Attention heads            & 8     & 8     & 8     & 4   \\
Encoder layers             & 5     & 5     & 5     & 5   \\
Local decoder layers       & 5     & 5     & 5     & 5   \\
Global decoder layers      & 5     & 5     & 5     & 5   \\
Receptive field $r$        & 4     & 4     & 4     & 4   \\
Stride ($T{=}24$)          & 2     & 2     & 2     & 2   \\
Stride ($T{=}64,128$)      & 4     & 4     & 4     & 4   \\
Codebook size $K$          & 4096  & 512   & 1024  & 1024 \\
Codebook dim               & 512   & 512   & 512   & 128  \\
Commitment weight          & 0.25  & 0.25  & 0.25  & 0.25 \\
Dropout                    & 0.0   & 0.0   & 0.0   & 0.0  \\
\midrule
\multicolumn{5}{l}{\textit{AR Transformer}} \\
$n_\text{embd}$            & 1024  & 1024  & 1024  & 1024 \\
$n_\text{head}$            & 8     & 8     & 8     & 8    \\
$n_\text{layer}$           & 2     & 2     & 2     & 2    \\
Random replace prob        & \multicolumn{4}{c}{0.1} \\
Lookback                   & \multicolumn{4}{c}{6}   \\
Watermark $\delta$         & 10    & 10    & 10    & 20  \\
\midrule
\multicolumn{5}{l}{\textit{Training}} \\
Optimizer                  & \multicolumn{4}{c}{AdamW, $\beta{=}(0.9,\,0.99)$} \\
Batch size                 & \multicolumn{4}{c}{128} \\
Learning rate              & \multicolumn{4}{c}{$3\times10^{-4}$} \\
Gradient clip              & \multicolumn{4}{c}{1.0} \\
Pretrain steps (stage 1)   & \multicolumn{4}{c}{30{,}000} \\
Decoder steps (stage 2)    & \multicolumn{4}{c}{10{,}000} \\
Transformer steps (stage 3)& \multicolumn{4}{c}{30{,}000} \\
\midrule
\multicolumn{5}{l}{\textit{Robust Encoder}\textsuperscript{$\dagger$}} \\
$d_\text{model}$           & 1024  & 1024  & 1024  & 256  \\
Attention heads            & 16    & 16    & 16    & 4    \\
Layers                     & \multicolumn{4}{c}{3} \\
Receptive field $r$        & \multicolumn{4}{c}{4} \\
Optimizer                  & \multicolumn{4}{c}{AdamW, $\beta{=}(0.9,\,0.99)$} \\
Learning rate              & \multicolumn{4}{c}{$1\times10^{-4}$} \\
Gradient clip              & \multicolumn{4}{c}{1.0} \\
Training steps             & \multicolumn{4}{c}{50{,}000} \\
Val interval               & \multicolumn{4}{c}{5{,}000} \\
% Train/val split            & \multicolumn{4}{c}{90/10 (seed 42)} \\
% Gen.\ seeds                & \multicolumn{4}{c}{4 (1, 12, 123, 1234)} \\
\midrule
\multicolumn{5}{l}{\textit{Attack augmentation (robust encoder training)}} \\
Clean (no attack)          & \multicolumn{4}{c}{-} \\
Offset                     & \multicolumn{4}{c}{factor $\in \{0.05,\, 0.30\}$} \\
$\textbf{crop}_\text{var}$\textsuperscript{$\dagger$} & \multicolumn{4}{c}{factor $\in \{0.05,\, 0.30\}$} \\
Insertion                  & \multicolumn{4}{c}{factor $\in \{0.05,\, 0.30\}$} \\
% Total chunks               & \multicolumn{4}{c}{7 (clean $+$ 3 attacks $\times$ 2 factors)} \\
% Attack seed                & \multicolumn{4}{c}{123} \\
Gaussian noise (retrained $E^\star_\rho$ only) & \multicolumn{4}{c}{$\alpha = 0.3$} \\
Smoothing (retrained $E^\star_\rho$ only)      & \multicolumn{4}{c}{$w = 3$} \\
\bottomrule
\end{tabular}
\end{table}

% -------------------------------------------------------
% SDformer
% -------------------------------------------------------
\begin{table}[h!]
\centering
\caption{SDformer hyperparameter settings.}
\label{tab:hp-vqvae}
\begin{tabular}{lcccc}
\toprule
Parameter & Energy & ETTh & fMRI & Stocks \\
\midrule
\multicolumn{5}{l}{\textit{Encoder / Decoder}} \\
Feature size $D$           & 28    & 7     & 50    & 6   \\
Hidden dim                 & 512   & 512   & 512   & 512 \\
Residual blocks            & 2     & 2     & 2     & 2   \\
Downsample rate            & 4     & 4     & 4     & 4   \\
EMA decay                  & 0.99  & 0.99  & 0.99  & 0.99 \\
Embed loss weight          & 0.01  & 0.5   & 0.01  & 2.0 \\
Quantizer type             & \multicolumn{4}{c}{cosine} \\
Codebook size $K$          & 512   & 512   & 512   & 512 \\
Codebook dim               & 512   & 512   & 512   & 256 \\
\midrule
\multicolumn{5}{l}{\textit{AR Transformer}} \\
$n_\text{embd}$            & 1024  & 1024  & 1024  & 1024 \\
$n_\text{head}$            & 8     & 8     & 8     & 8    \\
$n_\text{layer}$           & 2     & 6     & 2     & 2    \\
Random replace prob        & 0.1   & 0.3   & 0.1   & 0.3  \\
Watermark $\delta$         & \multicolumn{4}{c}{10} \\
\midrule
\multicolumn{5}{l}{\textit{Training}} \\
Optimizer                  & \multicolumn{4}{c}{AdamW, $\beta{=}(0.9,\,0.99)$} \\
Batch size                 & 64    & 128   & 64    & 128  \\
Learning rate              & \multicolumn{4}{c}{$3\times10^{-4}$} \\
Gradient clip              & \multicolumn{4}{c}{1.0} \\
VQ-VAE steps (stage 1)     & \multicolumn{4}{c}{50{,}000} \\
Transformer steps (stage 2)& \multicolumn{4}{c}{50{,}000} \\
\bottomrule
\end{tabular}
\end{table}

\subsection{Dataset Details}

We use four multivariate time-series datasets spanning diverse domains.
All datasets are segmented into overlapping windows of size $T \in \{24, 64, 128\}$ with stride 1,
yielding a set of fixed-length samples.
Each sample is normalized per-variable to $[-1, 1]$ via MinMax scaling.
The resulting samples split 80\%/20\% into train and test sets.
Table~\ref{tab:dataset_details} summarizes the dataset statistics.

\begin{table}[h!]
\centering
\caption{Details of datasets used in experiments.}
\label{tab:dataset_details}
\begin{tabular}{lrrl}
\toprule
Dataset & \# Timesteps & \# Features & Source \\
\midrule
Stocks  & 3,685  & 6  & \url{https://finance.yahoo.com} \\
ETTh    & 17,420 & 7  & \url{https://github.com/zhouhaoyi/ETDataset} \\
Energy  & 19,735 & 28 & \url{https://archive.ics.uci.edu} \\
fMRI    & 10,000 & 50 & \url{https://www.fmrib.ox.ac.uk/datasets} \\
\bottomrule
\end{tabular}
\end{table}

\subsection{Attack Settings}
\label{app:attacks}

We evaluate detection robustness under three post-editing attacks: offset, insert, and crop. Each attack is applied with strength factor $\alpha \in \{0.05,0.30\}$.

The \textbf{offset} attack shifts each variable by $\alpha \times \bar{x}_d$, where $\bar{x}_d$ is the temporal mean of variable $d$ in the sample. The \textbf{insert} attack replaces $\lfloor \alpha T \rfloor$ randomly selected time steps with uniform random values drawn from the observed range of each variable. The \textbf{crop} attack removes a random rectangular region in the $(T,D)$ space, retaining a window of size $(1-\alpha)T \times (1-\alpha)D$ at a randomly chosen position. The removed region is filled with the per-variable midpoint, defined as $(\min + \max)/2$ computed over the surviving window. Under standard min-max normalization followed by rescaling to $[-1,1]$, this midpoint reduces to zero, making the fill semantically neutral. All model families use the same randomly drawn crop window at evaluation time.

For robust encoder training, we use an additional augmentation, $\textbf{crop}_\text{var}$, which retains a contiguous block of $(1-\alpha)D$ variables over the full temporal axis and fills the remaining $\lfloor \alpha D \rfloor$ variables with the per-variable midpoint.   $\textbf{crop}_\text{var}$ differs from \textbf{crop} in that it preserves the temporal dimension and perturbs only the variable axis, thereby exposing the encoder to feature-level corruption without removing local temporal context entirely.

For the non-local evaluation in Table~\ref{tab:nonlocal}, the Gaussian noise attack adds zero-mean Gaussian noise to every observation with standard deviation $\alpha$ times the within-sample standard deviation of each variable, using $\alpha=0.3$. The smoothing attack replaces each observation by the average of a centered window of $w=3$ consecutive time steps. % [INSERT: boundary handling at the first and last time step, if not plain truncation]
The retrained robust encoder of Table~\ref{tab:nonlocal} uses the augmentation set of Table~\ref{tab:hp-lvqvae} extended with these two attacks, and only the robust-encoder stage is retrained. The local tokenizer, the global decoder, the autoregressive transformer, and the watermark schedule are those of the main experiments.

\subsection{Evaluation Metrics}

\paragraph{Context-FID}
Context-FID is an evaluation metric proposed by Jeha et al.~\cite{jeha2022psa} to measure the distributional gap between real and generated time series. It can be viewed as a time-series-oriented extension of the conventional Fréchet Inception Distance (FID), which is widely used in image generation tasks.

The key difference from image-based FID lies in the feature representation. While standard FID extracts image embeddings using an Inception network, Context-FID employs TS2Vec~\cite{yue2022ts2vec}, a representation model specifically designed for time series. Since TS2Vec captures temporal characteristics of sequential data, the resulting distance is more suitable for assessing synthetic time series quality.

Yue et al.~\cite{yue2022ts2vec} further observed that generative models achieving lower Context-FID scores often show stronger downstream forecasting performance. This suggests that Context-FID reflects not only superficial similarity but also useful temporal distributional alignment. Accordingly, a smaller Context-FID value indicates that the generated time series more closely match the real data distribution.

\paragraph{Correlational score}
The correlational score evaluates whether synthetic time series reproduce the inter-variable dependency patterns observed in real data. To this end, we first compute the covariance between the $i$-th and $j$-th variables over the temporal dimension. Following~\cite{liao2020conditional}, the covariance is defined as:

$$
\mathrm{Cov}_{i,j}
=
\frac{1}{W}
\sum_{t=1}^{W} K_i^t K_j^t
-
\left(
\frac{1}{W}
\sum_{t=1}^{W} K_i^t
\right)
\left(
\frac{1}{W}
\sum_{t=1}^{W} K_j^t
\right).
$$

Here, $W$ is the sequence length, and $K_i^t$ and $K_j^t$ denote the values of the $i$-th and $j$-th features at time step $t$. The covariance is obtained by subtracting the product of the two feature means from the mean of their element-wise products.

After computing covariance matrices for both real and synthetic data, we compare their normalized correlation structures using the following discrepancy measure~\cite{yuan2024diffusionts}:

$$
\frac{1}{10}
\sum^d_{i,j}
\left|
\frac{\mathrm{Cov}_{i,j}^{R}}
{\sqrt{\mathrm{Cov}_{i,i}^{R}\mathrm{Cov}_{j,j}^{R}}}
-
\frac{\mathrm{Cov}_{i,j}^{S}}
{\sqrt{\mathrm{Cov}_{i,i}^{S}\mathrm{Cov}_{j,j}^{S}}}
\right|.
$$

In the equation above, superscripts $R$ and $S$ indicate real and synthetic data, respectively, and $d$ denotes the number of features. The summation is computed over all feature pairs. A smaller correlational score means that the synthetic data better preserves the pairwise correlation structure of the real data.

\paragraph{Discriminative score}

The discriminative score measures the distinguishability between real and synthetic time series. Specifically, a classifier is trained to predict whether each sample comes from the real dataset or the generated dataset, and the score is calculated as $|\mathrm{accuracy} - 0.5|$.

If the generated samples are highly realistic, the classifier should not be able to reliably distinguish them from real samples, and its accuracy should approach $0.5$. Therefore, lower discriminative scores indicate better generation quality. Following the evaluation protocol of TimeGAN \cite{yoon2019time}, we use a two-layer GRU classifier for this evaluation.

\paragraph{Predictive score}

The predictive score assesses whether generated time series preserve temporal patterns that are useful for forecasting. Rather than directly measuring visual or statistical similarity, this metric evaluates the utility of generated sequences in a prediction task.

Following the evaluation protocol of TimeGAN~\cite{yoon2019time}, we use a one-layer GRU-based predictor. We report the mean absolute error (MAE) between the predicted values and the ground-truth values on the test set. A lower predictive score indicates that the synthetic data better preserves the temporal dynamics required for sequence prediction.
%\footnote{\url{https://github.com/zwzhang123/TimeGAN-pytorch/tree/main}}

\paragraph{$z$-score}

The $z$-score is used to evaluate the detectability of watermarks. It measures how far the detection statistic of watermarked samples deviates from that of non-watermarked samples. This metric is naturally connected to hypothesis testing. The null hypothesis $H_0$ assumes that the sample does not contain a watermark generated by the corresponding watermarking method.

When the $z$-score is close to zero, the watermarked and non-watermarked samples are statistically difficult to separate. A sufficiently large positive $z$-score, in contrast, provides evidence against $H_0$ and supports the conclusion that a watermark is present.

We consider methods that produce a scalar per-sample detection statistic $x_i$, such as bit accuracy for TimeWak \cite{soi2025timewak} or green-list ratio for LVQMark. For these methods, the $z$-score is computed at the population level via a subsampled-mean construction. Let $\mu_{NW}$ and $\sigma_{NW}$ denote the mean and standard deviation of $x_i$ over a non-watermarked, attack-free reference set generated by the same model. Following TimeWak \cite{soi2025timewak}, we draw $B=100$ subsamples of size $n=1000$ without replacement. For each replicate $b$, we compute

$$
Z_b
=
\frac{
\bar{x}^{(b)} - \mu_{NW}
}{
\sigma_{NW}/\sqrt{n}
}.
$$

We then report $\overline{Z}=\mathrm{mean}(Z_b)$, $\sigma_{Z}=\mathrm{std}(Z_b)$, and the empirical decision threshold $\Pr\!\left(Z_b > \Phi^{-1}(0.999)\right) \approx 3.0902$.

\paragraph{TPR@X\%FPR}
TPR@X\%FPR is the true positive rate at a threshold calibrated to yield an X\% false positive rate on clean non-watermarked samples. It measures detection power under a fixed false-positive constraint, which is particularly important in provenance verification, where false positives can lead to incorrect attribution of synthetic content. In our setting, this metric allows us to compare how reliably different methods detect watermarked samples when they are required to satisfy the same nominal false-positive budget.

\subsection{Compute Resources}

All experiments were conducted on a server with three GPUs: one NVIDIA H100 PCIe (80 GB) and two NVIDIA RTX PRO 6000 Blackwell Server Edition (96 GB each), an Intel Xeon Gold 6444Y dual-socket CPU (64 logical cores, up to 4.0 GHz).

\subsection{Computational Cost Measurement}
\label{app:cost}

All cost measurements are taken on ETTh with $T=64$ on a single NVIDIA RTX PRO 6000 GPU, with TimeWak over DiffusionTS as the baseline. Training time is the wall-clock time to train each model as configured for the main experiments, with all four stages of LVQMark included (Table~\ref{tab:hp-lvqvae}). Generation throughput is the number of samples produced per second on the GPU, where TimeWak runs its diffusion sampler and L-VQVAE samples 16 tokens autoregressively and decodes them with the global decoder in one pass.

The verification stage is measured end to end, from the observed series to the detection decision. For LVQMark it is one forward pass of the robust encoder $E^\star_\rho$ (three windowed cross-attention layers, $d_{\text{model}}=1024$, 38.3M parameters for ETTh) followed by the green-ratio count, without the generator or the autoregressive transformer. For TimeWak it is the 249-step BDIA inversion of the DiffusionTS model (0.38M parameters, applied sequentially) followed by its value-domain statistical test. FLOPs per sample count the dense operations of each pipeline, 114 MFLOP per window over 16 windows for LVQMark and 38.5 MFLOP per diffusion step over 249 steps for TimeWak, whose statistical test adds a negligible amount. GPU latency is reported at batch size 1, GPU throughput and peak memory at batch size 64. CPU measurements fix a single thread (\texttt{torch.set\_num\_threads(1)}) as a core-count-independent proxy for edge devices. Table~\ref{tab:cost_full} lists all measurements.

\begin{table}[h]
\centering
\setlength{\tabcolsep}{5pt}
\renewcommand{\arraystretch}{1.13}
\caption{Full cost measurements on ETTh ($T=64$) on a single GPU. Best results are in bold.}
\label{tab:cost_full}
\resizebox{\textwidth}{!}{%
\begin{tabular}{l|l|cc}
\hline
\hline
\rowcolor{gray!20}
\textbf{Stage} & \textbf{Metric} & \textbf{TimeWak} & \textbf{LVQMark (ours)} \\
\hline
Training (once) & Wall-clock time (min) & \bestm{9.3} & 41.8 \\
\hline
Generation & Throughput, GPU (samples/s) & 350 & \bestm{4{,}922} \\
\hline
\multirow{6}{*}{Verification} & FLOPs per sample (GFLOP) & 9.59 & \bestm{1.83} \\
 & Latency, GPU, batch 1 (ms/sample) & 1,010 & \bestm{0.72} \\
 & Throughput, GPU, batch 64 (samples/s) & 47 & \bestm{19{,}090} \\
 & Peak GPU memory, batch 64 (MB) & 781 & \bestm{259} \\
 & Latency, 1 CPU thread, batch 1 (ms/sample) & 892 & \bestm{23.4} \\
 & Throughput, 1 CPU thread, batch 64 (samples/s) & 3.8 & \bestm{85.9} \\
\hline
\hline
\end{tabular}
}
\end{table}

\section{Effect of Partition Rules on Robust Detection}
\label{app:partition_rule}

Tables~\ref{tab:partition_zscore_30} and \ref{tab:partition_zscore_5} compare the context-dependent KGW partition with our context-independent partition under 5\% and 30\% attacks, respectively. Under the KGW rule, the green set at position $n$ is determined by the recovered token at $n{-}1$, so a single re-encoding error can affect the partition at the next position and propagate a local perturbation beyond the attacked region. In contrast, our context-independent partition depends only on the position index and the secret key, eliminating this cross-position coupling and confining each re-encoding error to its originating position. As a result, our partition consistently yields higher watermark $z$-scores across datasets and attack types under both attack strengths. This shows that the robustness benefit of preserving locality is systematic rather than specific to the stronger 30\% setting. Combined with parity alternation and the bias-free warm-up (Section~\ref{sec:watermark-injection}), the proposed design leads to more stable recovered evidence and stronger overall watermark detectability.

\begin{table*}[h!]
\centering
\setlength{\tabcolsep}{3pt}
\renewcommand{\arraystretch}{1.13}
\caption{Partition rule comparison. $z$-scores under 30\% attacks. Best results are in bold.}
\label{tab:partition_zscore_30}
\resizebox{\textwidth}{!}{%
\begin{tabular}{c|ccc|ccc|ccc|ccc}
\hline
\hline
\rowcolor{gray!20}
 & \multicolumn{3}{c|}{\textbf{Stocks}} & \multicolumn{3}{c|}{\textbf{Energy}} & \multicolumn{3}{c|}{\textbf{ETTh}} & \multicolumn{3}{c}{\textbf{fMRI}} \\
\rowcolor{gray!20}
\hline
\textbf{Partition} & \textbf{Offset} & \textbf{Crop} & \textbf{Insert.} & \textbf{Offset} & \textbf{Crop} & \textbf{Insert.} & \textbf{Offset} & \textbf{Crop} & \textbf{Insert.} & \textbf{Offset} & \textbf{Crop} & \textbf{Insert.} \\
\hline
KGW  & 14.96 & 5.38 & 2.76 & 14.73 & 10.06 & 9.98 & 10.51 & 8.14 & 7.26 & 12.71 & 13.12 & 9.31 \\
Ours & \bestm{15.39} & \bestm{16.19} & \bestm{9.16} & \bestm{17.30} & \bestm{13.77} & \bestm{13.14} & \bestm{17.10} & \bestm{16.09} & \bestm{13.31} & \bestm{14.55} & \bestm{16.72} & \bestm{11.88} \\
\hline
\hline
\end{tabular}%
}
\end{table*}
\begin{table*}[h!]
\centering
\setlength{\tabcolsep}{3pt}
\renewcommand{\arraystretch}{1.13}
\caption{Partition rule comparison. $z$-scores under 5\% attacks. Best results are in bold.}
\label{tab:partition_zscore_5}
\resizebox{\textwidth}{!}{%
\begin{tabular}{c|ccc|ccc|ccc|ccc}
\hline
\hline
\rowcolor{gray!20}
 & \multicolumn{3}{c|}{\textbf{Stocks}} & \multicolumn{3}{c|}{\textbf{Energy}} & \multicolumn{3}{c|}{\textbf{ETTh}} & \multicolumn{3}{c}{\textbf{fMRI}} \\
\rowcolor{gray!20}
\hline
\textbf{Partition} & \textbf{Offset} & \textbf{Crop} & \textbf{Insert} & \textbf{Offset} & \textbf{Crop} & \textbf{Insert} & \textbf{Offset} & \textbf{Crop} & \textbf{Insert} & \textbf{Offset} & \textbf{Crop} & \textbf{Insert} \\
\hline
KGW  & 12.95 & 8.62 & 8.44 & 14.80 & 13.30 & 14.27 & 11.09 & 9.02 & 10.55 & 12.77 & 13.14 & 12.66 \\
Ours & \bestm{19.51} & \bestm{15.80} & \bestm{16.37} & \bestm{17.35} & \bestm{15.74} & \bestm{16.98} & \bestm{17.65} & \bestm{16.53} & \bestm{17.80} & \bestm{14.68} & \bestm{15.19} & \bestm{14.49} \\
\hline
\hline
\end{tabular}%
}
\end{table*}

\section{Receptive-Field Sweep}
\label{app:rf}

The models with $R \in \{8, 16, 32, 64\}$ in Table~\ref{tab:rf} use stride $s=R$, so the windows tile the sequence without overlap, and each window is summarized by $R/4$ codes from the same codebook, so that the token sequence keeps 16 tokens at $T=64$ and the codebook size of Table~\ref{tab:hp-lvqvae} is unchanged. Each model is trained from scratch with the four-stage procedure of the main experiments, and detection uses a robust encoder trained for that $R$ with the augmentation set of Table~\ref{tab:hp-lvqvae}. % [INSERT: confirm that the robust encoder was retrained for every R]
Table~\ref{tab:rf_appendix} reports clean detection and Context-FID across $R$. At $R=8$ and $16$ on Stocks and at $R \geq 32$ on fMRI the watermark is not recoverable on clean samples, so the zeros in Table~\ref{tab:rf} at those settings reflect the representation rather than the attack. Context-FID rises with $R$ for both non-watermarked and watermarked generation, gradually at moderate fields and sharply at $R=64$.

\begin{table}[h]
\centering
\setlength{\tabcolsep}{4pt}
\renewcommand{\arraystretch}{1.13}
\caption{Clean detection and generation quality across the receptive field $R$ for 64-length sequences. No-WM / WM: Context-FID of non-watermarked and watermarked generation.}
\label{tab:rf_appendix}
\resizebox{\textwidth}{!}{%
\begin{tabular}{c|l|ccccc}
\hline
\hline
\rowcolor{gray!20}
\textbf{Dataset} & \textbf{Metric} & $R{=}4$ & $R{=}8$ & $R{=}16$ & $R{=}32$ & $R{=}64$ \\
\hline
\multirow{2}{*}{Stocks} & Clean TPR & 1.00 & 0.00 & 0.00 & 1.00 & 1.00 \\
 & C-FID (No-WM / WM) & 0.04 / 0.07 & 0.03 / 0.15 & 0.03 / 0.32 & 0.08 / 0.39 & 0.10 / 0.22 \\
\hline
\multirow{2}{*}{ETTh} & Clean TPR & 1.00 & 1.00 & 1.00 & 1.00 & 0.03 \\
 & C-FID (No-WM / WM) & 0.02 / 0.03 & 0.03 / 0.06 & 0.03 / 0.08 & 0.07 / 0.09 & 11.44 / 11.90 \\
\hline
\multirow{2}{*}{Energy} & Clean TPR & 1.00 & 1.00 & 1.00 & 1.00 & 0.10 \\
 & C-FID (No-WM / WM) & 0.03 / 0.04 & 0.07 / 0.14 & 0.20 / 0.28 & 0.54 / 0.59 & 1.39 / 1.52 \\
\hline
\multirow{2}{*}{fMRI} & Clean TPR & 1.00 & 1.00 & 1.00 & 0.00 & 0.00 \\
 & C-FID (No-WM / WM) & 0.10 / 0.13 & 0.15 / 0.37 & 3.24 / 3.22 & 10.51 / 10.74 & 26.82 / 27.33 \\
\hline
\hline
\end{tabular}
}
\end{table}

\section{Experimental Results For Other Lengths}
This section provides additional results for other sequence lengths ($T=24,128$) and complements the main analysis at $T=64$. The overall trend is consistent with the main results.

\subsection{Watermark Detection and Generation Quality}

\paragraph{False-positive behavior under post-editing attacks}
\label{app:short}
Tables~\ref{tab:main_24} and~\ref{tab:main_128} extend the analysis of bidirectional null shift in Section~\ref{sec:exp-drift} to other sequence lengths.
Consistent with the main results, the same qualitative pattern persists across other sequence lengths, further supporting that the attacked null remains stable under L-VQVAE with LVQMark while existing baselines exhibit substantial drift.

\begin{table*}[h!]
\centering
\setlength{\tabcolsep}{3pt}
\renewcommand{\arraystretch}{1.13}
\caption{Results of synthetic time series watermark detection and quality. Watermarked (TPR) and Non-watermarked (FPR, mean $z$-score) detections are evaluated under 30\% attacks. Quality metrics are for 24-length. Best results are in bold, and second-best are underlined. $z$-scores with $|z|>3.09$ are marked in \textcolor{red}{red}.}
\label{tab:main_24}
\resizebox{\textwidth}{!}{%
\begin{tabular}{c|l|l|ccc|cr|cr|cr|cccc}
\hline
\hline
\rowcolor{gray!20}
\multicolumn{3}{c|}{\textbf{Setting}}
& \multicolumn{3}{c|}{\textbf{Watermark (TPR $\uparrow$)}}
& \multicolumn{6}{c|}{\textbf{Non-watermarked (FPR $\downarrow$ | $z$-score)}}
& \multicolumn{4}{c}{\textbf{Quality Metric ($\downarrow$)}} \\
\hline
\rowcolor{gray!20}
\multicolumn{1}{c|}{\textbf{Dataset}}& \multicolumn{1}{c|}{\textbf{Model}} & \multicolumn{1}{c|}{\textbf{Method}}
& \textbf{Offset} & \textbf{Crop} & \textbf{Insert}
& \multicolumn{2}{c|}{\textbf{Offset}}
& \multicolumn{2}{c|}{\textbf{Crop}}
& \multicolumn{2}{c|}{\textbf{Insert}}
 & \textbf{C-FID} & \textbf{Corr.} & \textbf{Disc.} & \textbf{Pred.}  \\
\hline
\multirow{5}{*}{{Stocks}} & \multirow{3}{*}{DiffusionTS} & TR & 0.00 & \bestm{1.00} & 0.00 & \bestm{0.00} & $+0.19$ & 1.00 & \textcolor{red}{$+63.26$} & \bestm{0.00} & $+1.91$ & 0.96 & 0.10 & 0.20 & \bestm{0.04}  \\
 & &  GS & \bestm{1.00} & 0.45 & \bestm{1.00} & \bestm{0.00} & $-1.15$ & \bestm{0.00} & \textcolor{red}{$-6.32$} & \bestm{0.00} & $+0.16$ & 8.88 & 0.09 & 0.43 & \bestm{0.04}  \\
 & &  TimeWak & \bestm{1.00} & \bestm{1.00} & \bestm{1.00} & \bestm{0.00} & $-0.99$ & 0.02 & $+1.39$ & \bestm{0.00} & $-1.77$ & 0.33 & \secondm{0.02} & 0.16 & \bestm{0.04}  \\
\cline{2-16}
 & SDformer & LVQMark & 0.13 & 0.81 & 0.22 & \bestm{0.00} & $+0.51$ & 0.86 & \textcolor{red}{$+4.49$} & 0.01 & $+1.33$ & \secondm{0.11} & \bestm{0.01} & \bestm{0.02} & \bestm{0.04}  \\
\cline{2-16}
 & L-VQVAE & LVQMark & \bestm{1.00} & 0.99 & 0.98 & \bestm{0.00} & $-0.25$ & 0.04 & $+1.10$ & 0.02 & $+0.38$ & \bestm{0.07} & \bestm{0.01} & \secondm{0.11} & \bestm{0.04}  \\
\hline
\hline
\multirow{5}{*}{{ETTh}} & \multirow{3}{*}{DiffusionTS} & TR & \bestm{1.00} & \bestm{1.00} & \bestm{1.00} & \bestm{0.00} & $+0.95$ & 1.00 & \textcolor{red}{$+10.84$} & 1.00 & \textcolor{red}{$+15.97$} & 1.56 & 0.18 & 0.27 & \secondm{0.14}  \\
 & &  GS & \bestm{1.00} & \bestm{1.00} & \bestm{1.00} & 1.00 & \textcolor{red}{$+12.01$} & 1.00 & \textcolor{red}{$+18.58$} & 1.00 & \textcolor{red}{$+6.96$} & 4.66 & 0.42 & 0.38 & 0.19  \\
 & &  TimeWak & \bestm{1.00} & 0.48 & \bestm{1.00} & \bestm{0.00} & $-0.34$ & 0.02 & $+1.05$ & \bestm{0.00} & $-1.68$ & 0.23 & 0.21 & 0.08 & \bestm{0.12}  \\
\cline{2-16}
 & SDformer & LVQMark & 0.98 & 0.03 & 0.07 & 0.01 & $+0.79$ & \bestm{0.01} & $+0.18$ & \bestm{0.00} & $+0.70$ & \secondm{0.13} & \secondm{0.08} & \secondm{0.04} & \bestm{0.12}  \\
\cline{2-16}
 & L-VQVAE & LVQMark & \bestm{1.00} & \bestm{1.00} & \bestm{1.00} & \bestm{0.00} & $+0.03$ & \bestm{0.01} & $+0.41$ & \bestm{0.00} & $-0.15$ & \bestm{0.05} & \bestm{0.05} & \bestm{0.02} & \bestm{0.12}  \\
\hline
\hline
\multirow{5}{*}{{Energy}} & \multirow{3}{*}{DiffusionTS} & TR & 0.00 & \bestm{1.00} & \bestm{1.00} & \bestm{0.00} & $+0.43$ & 1.00 & \textcolor{red}{$+39.25$} & 1.00 & \textcolor{red}{$+26.77$} & 0.43 & 2.66 & 0.41 & \secondm{0.30}  \\
 & &  GS & \bestm{1.00} & \bestm{1.00} & \bestm{1.00} & 1.00 & \textcolor{red}{$+13.17$} & 1.00 & \textcolor{red}{$+65.49$} & 1.00 & \textcolor{red}{$+14.31$} & 1.58 & 3.37 & 0.49 & 0.33  \\
 & &  TimeWak & \bestm{1.00} & \bestm{1.00} & \bestm{1.00} & \bestm{0.00} & $+0.33$ & 0.99 & \textcolor{red}{$+6.03$} & \bestm{0.00} & $-1.22$ & \secondm{0.09} & 1.53 & \secondm{0.14} & \bestm{0.25}  \\
\cline{2-16}
 & SDformer & LVQMark & \bestm{1.00} & 0.76 & \bestm{1.00} & \bestm{0.00} & $+0.29$ & 0.07 & $+1.48$ & \bestm{0.00} & $+0.32$ & 0.12 & \secondm{1.27} & 0.23 & \bestm{0.25}  \\
\cline{2-16}
 & L-VQVAE & LVQMark & \bestm{1.00} & \bestm{1.00} & \bestm{1.00} & \bestm{0.00} & $-0.16$ & \bestm{0.00} & $-0.19$ & \bestm{0.00} & $-0.03$ & \bestm{0.03} & \bestm{1.02} & \bestm{0.09} & \bestm{0.25}  \\
\hline
\hline
\multirow{5}{*}{{fMRI}} & \multirow{3}{*}{DiffusionTS} & TR & 0.04 & 0.00 & 0.00 & \bestm{0.00} & $+0.42$ & \bestm{0.00} & $+1.00$ & 1.00 & \textcolor{red}{$+4.89$} & 2.26 & 13.42 & 0.50 & 0.15  \\
 & &  GS & \bestm{1.00} & \bestm{1.00} & \bestm{1.00} & 0.24 & $+2.33$ & \bestm{0.00} & $-2.26$ & \bestm{0.00} & \textcolor{red}{$-6.16$} & 0.71 & 15.21 & 0.50 & \secondm{0.11}  \\
 & &  TimeWak & \bestm{1.00} & \bestm{1.00} & \bestm{1.00} & \bestm{0.00} & $+0.19$ & 0.22 & $+2.39$ & \bestm{0.00} & $+0.26$ & \bestm{0.18} & \secondm{1.98} & \bestm{0.09} & \bestm{0.10}  \\
\cline{2-16}
 & SDformer & LVQMark & \bestm{1.00} & 0.99 & \bestm{1.00} & \bestm{0.00} & $+0.14$ & 0.01 & $-0.41$ & 0.01 & $+0.01$ & 0.94 & 3.25 & \secondm{0.21} & \bestm{0.10}  \\
\cline{2-16}
 & L-VQVAE & LVQMark & \bestm{1.00} & \bestm{1.00} & \bestm{1.00} & \bestm{0.00} & $+0.11$ & 0.02 & $-0.23$ & \bestm{0.00} & $-0.05$ & \secondm{0.20} & \bestm{1.92} & 0.23 & \bestm{0.10}  \\
\hline
\hline
\end{tabular}%
}
\end{table*}

\begin{table*}[h!]
\centering
\setlength{\tabcolsep}{3pt}
\renewcommand{\arraystretch}{1.13}
\caption{Results of synthetic time series watermark detection and quality. Watermarked (TPR) and Non-watermarked (FPR, mean $z$-score) detections are evaluated under 30\% attacks. Quality metrics are for 128-length. Best results are in bold, and second-best are underlined. $z$-scores with $|z|>3.09$ are marked in \textcolor{red}{red}.}
\label{tab:main_128}
\resizebox{\textwidth}{!}{%
\begin{tabular}{c|l|l|ccc|cr|cr|cr|cccc}
\hline
\hline
\rowcolor{gray!20}
\multicolumn{3}{c|}{\textbf{Setting}}
& \multicolumn{3}{c|}{\textbf{Watermark (TPR $\uparrow$)}}
& \multicolumn{6}{c|}{\textbf{Non-watermarked (FPR $\downarrow$ | $z$-score)}}
& \multicolumn{4}{c}{\textbf{Quality Metric ($\downarrow$)}} \\
\hline
\rowcolor{gray!20}
\multicolumn{1}{c|}{\textbf{Dataset}}& \multicolumn{1}{c|}{\textbf{Model}} & \multicolumn{1}{c|}{\textbf{Method}}
& \textbf{Offset} & \textbf{Crop} & \textbf{Insert}
& \multicolumn{2}{c|}{\textbf{Offset}}
& \multicolumn{2}{c|}{\textbf{Crop}}
& \multicolumn{2}{c|}{\textbf{Insert}}
 & \textbf{C-FID} & \textbf{Corr.} & \textbf{Disc.} & \textbf{Pred.}  \\
\hline
\multirow{5}{*}{{Stocks}} & \multirow{3}{*}{DiffusionTS} & TR & 0.00 & \bestm{1.00} & \bestm{1.00} & \bestm{0.00} & $+0.30$ & 1.00 & \textcolor{red}{$+84.66$} & 1.00 & \textcolor{red}{$+20.60$} & 3.05 & 0.09 & 0.24 & \bestm{0.04}  \\
 & &  GS & \bestm{1.00} & 0.04 & \bestm{1.00} & \bestm{0.00} & \textcolor{red}{$-4.40$} & \bestm{0.00} & \textcolor{red}{$-10.04$} & \bestm{0.00} & \textcolor{red}{$-5.22$} & 2.63 & \secondm{0.04} & 0.19 & \bestm{0.04}  \\
 & &  TimeWak & \bestm{1.00} & \bestm{1.00} & \bestm{1.00} & \bestm{0.00} & $-0.16$ & \bestm{0.00} & \textcolor{red}{$-4.18$} & 0.69 & \textcolor{red}{$+3.74$} & 0.34 & \bestm{0.01} & \secondm{0.15} & \bestm{0.04}  \\
\cline{2-16}
 & SDformer & LVQMark & 0.00 & 0.00 & 0.00 & \bestm{0.00} & $-0.54$ & \bestm{0.00} & $-0.89$ & \bestm{0.00} & $-1.20$ & \secondm{0.12} & \bestm{0.01} & \bestm{0.03} & \bestm{0.04}  \\
\cline{2-16}
 & L-VQVAE & LVQMark & \bestm{1.00} & \bestm{1.00} & \bestm{1.00} & \bestm{0.00} & $+0.09$ & \bestm{0.00} & $-0.09$ & \bestm{0.00} & $+0.23$ & \bestm{0.08} & \bestm{0.01} & 0.18 & \bestm{0.04}  \\
\hline
\hline
\multirow{5}{*}{{ETTh}} & \multirow{3}{*}{DiffusionTS} & TR & 0.00 & \bestm{1.00} & \bestm{1.00} & \bestm{0.00} & $+1.52$ & 1.00 & \textcolor{red}{$+12.19$} & 1.00 & \textcolor{red}{$+36.77$} & 2.52 & 0.26 & 0.30 & \secondm{0.13}  \\
 & &  GS & \bestm{1.00} & \bestm{1.00} & \bestm{1.00} & \bestm{0.00} & \textcolor{red}{$-6.09$} & \bestm{0.00} & \textcolor{red}{$-13.71$} & \bestm{0.00} & \textcolor{red}{$-6.37$} & 5.59 & 0.23 & 0.39 & 0.15  \\
 & &  TimeWak & \bestm{1.00} & 0.92 & \bestm{1.00} & 0.02 & $+1.09$ & \bestm{0.00} & $-1.63$ & \bestm{0.00} & $+0.77$ & \secondm{1.08} & \secondm{0.18} & 0.15 & \bestm{0.11}  \\
\cline{2-16}
 & SDformer & LVQMark & 0.14 & 0.04 & 0.04 & \bestm{0.00} & $+0.53$ & 0.01 & $+0.76$ & \bestm{0.00} & $+1.01$ & \bestm{0.04} & \bestm{0.05} & \secondm{0.03} & \bestm{0.11}  \\
\cline{2-16}
 & L-VQVAE & LVQMark & \bestm{1.00} & \bestm{1.00} & \bestm{1.00} & \bestm{0.00} & $-0.04$ & \bestm{0.00} & $-0.05$ & \bestm{0.00} & $-0.30$ & \bestm{0.04} & \bestm{0.05} & \bestm{0.02} & \bestm{0.11}  \\
\hline
\hline
\multirow{5}{*}{{Energy}} & \multirow{3}{*}{DiffusionTS} & TR & 0.00 & \bestm{1.00} & \bestm{1.00} & \bestm{0.00} & $+2.20$ & 1.00 & \textcolor{red}{$+61.47$} & 1.00 & \textcolor{red}{$+98.65$} & 0.50 & 1.65 & 0.49 & 0.28  \\
 & &  GS & \bestm{1.00} & \bestm{1.00} & \bestm{1.00} & 1.00 & \textcolor{red}{$+15.43$} & 1.00 & \textcolor{red}{$+48.30$} & 1.00 & \textcolor{red}{$+27.89$} & 3.34 & 3.59 & 0.48 & 0.29  \\
 & &  TimeWak & \bestm{1.00} & 0.10 & \bestm{1.00} & \bestm{0.00} & $-1.57$ & \bestm{0.00} & $-0.78$ & \bestm{0.00} & $+0.66$ & 0.17 & 1.50 & 0.24 & \secondm{0.25}  \\
\cline{2-16}
 & SDformer & LVQMark & \bestm{1.00} & 0.00 & 0.32 & \bestm{0.00} & $+0.06$ & \bestm{0.00} & $-0.65$ & \bestm{0.00} & $-0.08$ & \secondm{0.04} & \bestm{0.67} & \bestm{0.05} & \secondm{0.25}  \\
\cline{2-16}
 & L-VQVAE & LVQMark & \bestm{1.00} & 0.99 & \bestm{1.00} & \bestm{0.00} & $-0.03$ & 0.02 & $+0.00$ & \bestm{0.00} & $+0.13$ & \bestm{0.03} & \secondm{0.74} & \secondm{0.19} & \bestm{0.24}  \\
\hline
\hline
\multirow{5}{*}{{fMRI}} & \multirow{3}{*}{DiffusionTS} & TR & 0.00 & \bestm{1.00} & \bestm{1.00} & \bestm{0.00} & $+0.02$ & 1.00 & \textcolor{red}{$+6.33$} & 1.00 & \textcolor{red}{$+21.25$} & 4.57 & 14.15 & 0.45 & 0.17  \\
 & &  GS & \bestm{1.00} & \bestm{1.00} & \bestm{1.00} & \bestm{0.00} & $-0.70$ & \bestm{0.00} & \textcolor{red}{$-57.42$} & 0.07 & $+0.52$ & 1.05 & 5.97 & 0.50 & 0.11  \\
 & &  TimeWak & \bestm{1.00} & \bestm{1.00} & \bestm{1.00} & \bestm{0.00} & $-0.07$ & \bestm{0.00} & $-3.03$ & \bestm{0.00} & $-1.95$ & 0.81 & 1.81 & 0.36 & \secondm{0.10}  \\
\cline{2-16}
 & SDformer & LVQMark & 0.10 & 0.02 & 0.03 & \bestm{0.00} & $-0.03$ & 0.01 & $-0.25$ & \bestm{0.00} & $-0.15$ & \bestm{0.09} & \bestm{0.84} & \secondm{0.16} & \bestm{0.08}  \\
\cline{2-16}
 & L-VQVAE & LVQMark & \bestm{1.00} & \bestm{1.00} & \bestm{1.00} & \bestm{0.00} & $-0.09$ & \bestm{0.00} & $+0.16$ & \bestm{0.00} & $-0.01$ & \secondm{0.13} & \secondm{0.90} & \bestm{0.06} & \bestm{0.08}  \\
\hline
\hline
\end{tabular}%
}
\end{table*}

\paragraph{Generation quality}

Tables~\ref{tab:quality_24} and~\ref{tab:quality_128} show the generation quality results for sequence lengths 24 and 128, respectively. L-VQVAE remains competitive in quality while improving watermark reliability under attack.

\begin{table*}[h!]
\centering
\scriptsize
\setlength{\tabcolsep}{3pt}
\renewcommand{\arraystretch}{1.13}
\caption{Results of synthetic time series quality for 24-length sequences. Best results are in bold, and second-best are underlined.}
\label{tab:quality_24}
\resizebox{\textwidth}{!}{%
\begin{tabular}{l|cccc|cccc|cccc|cccc}
\hline
\hline
\rowcolor{gray!20}
 & \multicolumn{4}{c|}{\textbf{Stocks}} & \multicolumn{4}{c|}{\textbf{Energy}} & \multicolumn{4}{c|}{\textbf{ETTh}} & \multicolumn{4}{c}{\textbf{fMRI}} \\
\rowcolor{gray!20}
\hline
\textbf{Model} & \textbf{C-FID} & \textbf{Corr.} & \textbf{Disc.} & \textbf{Pred.} & \textbf{C-FID} & \textbf{Corr.} & \textbf{Disc.} & \textbf{Pred.} & \textbf{C-FID} & \textbf{Corr.} & \textbf{Disc.} & \textbf{Pred.} & \textbf{C-FID} & \textbf{Corr.} & \textbf{Disc.} & \textbf{Pred.} \\
\hline
TimeVQVAE & 0.507 & 0.049 & 0.361 & \secondm{0.047} & 2.527 & 6.632 & 0.476 & 0.383 & 4.417 & 0.263 & 0.415 & 0.199 & 12.532 & 61.599 & 0.459 & 0.156 \\
DiffusionTS & 0.329 & 0.022 & 0.160 & \bestm{0.037} & 0.087 & 1.534 & 0.137 & \secondm{0.253} & 0.231 & 0.212 & 0.078 & 0.121 & 0.181 & 1.976 & \secondm{0.085} & 0.100 \\
SDformer & \bestm{0.028} & \bestm{0.009} & \bestm{0.023} & \bestm{0.037} & \secondm{0.030} & \secondm{1.088} & \secondm{0.093} & \bestm{0.252} & \secondm{0.036} & \secondm{0.066} & \bestm{0.010} & \secondm{0.120} & \secondm{0.114} & \bestm{1.833} & \bestm{0.056} & \bestm{0.092} \\
\hline
L-VQVAE & \secondm{0.034} & \secondm{0.010} & \secondm{0.140} & \bestm{0.037} & \bestm{0.018} & \bestm{1.008} & \bestm{0.033} & \bestm{0.252} & \bestm{0.029} & \bestm{0.052} & \secondm{0.018} & \bestm{0.117} & \bestm{0.105} & \secondm{1.875} & 0.209 & \secondm{0.095} \\
\hline
\hline
\end{tabular}%
}
\end{table*}

\begin{table*}[h!]
\centering
\scriptsize
\setlength{\tabcolsep}{3pt}
\renewcommand{\arraystretch}{1.13}
\caption{Results of synthetic time series quality for 128-length sequences. Best results are in bold, and second-best are underlined.}
\label{tab:quality_128}
\resizebox{\textwidth}{!}{%
\begin{tabular}{l|cccc|cccc|cccc|cccc}
\hline
\hline
\rowcolor{gray!20}
 & \multicolumn{4}{c|}{\textbf{Stocks}} & \multicolumn{4}{c|}{\textbf{Energy}} & \multicolumn{4}{c|}{\textbf{ETTh}} & \multicolumn{4}{c}{\textbf{fMRI}} \\
\rowcolor{gray!20}
\hline
\textbf{Model} & \textbf{C-FID} & \textbf{Corr.} & \textbf{Disc.} & \textbf{Pred.} & \textbf{C-FID} & \textbf{Corr.} & \textbf{Disc.} & \textbf{Pred.} & \textbf{C-FID} & \textbf{Corr.} & \textbf{Disc.} & \textbf{Pred.} & \textbf{C-FID} & \textbf{Corr.} & \textbf{Disc.} & \textbf{Pred.} \\
\hline
TimeVQVAE & 0.293 & 0.087 & 0.135 & 0.039 & 7.738 & 6.631 & 0.498 & 0.344 & 2.205 & 0.251 & 0.256 & 0.143 & 30.672 & 45.692 & 0.250 & 0.149 \\
DiffusionTS & 0.344 & \secondm{0.014} & 0.152 & \secondm{0.037} & 0.169 & 1.505 & 0.235 & 0.249 & 1.075 & 0.175 & 0.145 & \secondm{0.113} & 0.810 & 1.812 & 0.357 & 0.100 \\
SDformer & \secondm{0.128} & \bestm{0.013} & \bestm{0.032} & \bestm{0.036} & \secondm{0.032} & \bestm{0.708} & \bestm{0.057} & \bestm{0.244} & \bestm{0.032} & \bestm{0.050} & \secondm{0.062} & \secondm{0.113} & \bestm{0.064} & \bestm{0.801} & \secondm{0.176} & \bestm{0.080} \\
\hline
L-VQVAE & \bestm{0.038} & \bestm{0.013} & \secondm{0.035} & \bestm{0.036} & \bestm{0.031} & \secondm{0.733} & \secondm{0.155} & \secondm{0.246} & \secondm{0.039} & \secondm{0.052} & \bestm{0.025} & \bestm{0.107} & \secondm{0.096} & \secondm{0.895} & \bestm{0.054} & \secondm{0.082} \\
\hline
\hline
\end{tabular}%
}
\end{table*}

\subsection{Component Analysis and Ablations}

\paragraph{Effect of watermark scheduling}
\label{app:abl-short}

Tables~\ref{tab:sampling_method_24} and~\ref{tab:sampling_method_128} extend the scheduling ablation in Section~\ref{sec:exp-abl} to other sequence lengths. We compare the default configuration against variants without unbiased warm-up (\textsc{no-warmup}) and without both alternating partition and unbiased warm-up (\textsc{no-alt-warmup}). Overall, the default configuration yields the most consistent quality across datasets, and the degradation is typically larger when both components are removed. These trends are broadly consistent with the main results at 64-length.

\begin{table*}[h!]
\centering
\setlength{\tabcolsep}{3pt}
\renewcommand{\arraystretch}{1.13}
\caption{Ablation of watermark scheduling components for 24-length sequences (LVQMark). \textsc{no-warmup}: unbiased warm-up removed. \textsc{no-alt-warmup}: both alternating partition and unbiased warm-up removed. Best results are in bold, and second-best are underlined.}
\label{tab:sampling_method_24}
\resizebox{\textwidth}{!}{%
\begin{tabular}{l|cccc|cccc|cccc|cccc}
\hline
\hline
\rowcolor{gray!20}
 & \multicolumn{4}{c|}{\textbf{Stocks}} & \multicolumn{4}{c|}{\textbf{Energy}} & \multicolumn{4}{c|}{\textbf{ETTh}} & \multicolumn{4}{c}{\textbf{fMRI}} \\
\rowcolor{gray!20}
\hline
\textbf{Variant} & \textbf{C-FID} & \textbf{Corr.} & \textbf{Disc.} & \textbf{Pred.} & \textbf{C-FID} & \textbf{Corr.} & \textbf{Disc.} & \textbf{Pred.} & \textbf{C-FID} & \textbf{Corr.} & \textbf{Disc.} & \textbf{Pred.} & \textbf{C-FID} & \textbf{Corr.} & \textbf{Disc.} & \textbf{Pred.} \\
\hline
\textsc{no-alt-warmup} & 0.211 & 0.014 & \secondm{0.141} & \bestm{0.037} & 0.081 & 1.172 & 0.151 & 0.254 & 0.180 & 0.113 & 0.066 & \secondm{0.123} & 0.419 & 2.350 & 0.298 & 0.101 \\
\textsc{no-warmup} & \secondm{0.160} & \secondm{0.013} & 0.224 & \bestm{0.037} & \secondm{0.060} & \secondm{1.063} & \secondm{0.130} & \secondm{0.253} & \secondm{0.093} & \secondm{0.059} & \secondm{0.039} & \secondm{0.123} & \secondm{0.346} & \secondm{2.067} & \secondm{0.296} & \secondm{0.098} \\
LVQMark & \bestm{0.074} & \bestm{0.007} & \bestm{0.110} & \bestm{0.037} & \bestm{0.028} & \bestm{1.022} & \bestm{0.090} & \bestm{0.252} & \bestm{0.046} & \bestm{0.048} & \bestm{0.021} & \bestm{0.117} & \bestm{0.200} & \bestm{1.920} & \bestm{0.229} & \bestm{0.097} \\
\hline
\hline
\end{tabular}%
}
\end{table*}

\begin{table*}[h!]
\centering
\setlength{\tabcolsep}{3pt}
\renewcommand{\arraystretch}{1.13}
\caption{Ablation of watermark scheduling components for 128-length sequences (LVQMark). \textsc{no-warmup}: unbiased warm-up removed. \textsc{no-alt-warmup}: both alternating partition and unbiased warm-up removed. Best results are in bold, and second-best are underlined.}
\label{tab:sampling_method_128}
\resizebox{\textwidth}{!}{%
\begin{tabular}{l|cccc|cccc|cccc|cccc}
\hline
\hline
\rowcolor{gray!20}
 & \multicolumn{4}{c|}{\textbf{Stocks}} & \multicolumn{4}{c|}{\textbf{Energy}} & \multicolumn{4}{c|}{\textbf{ETTh}} & \multicolumn{4}{c}{\textbf{fMRI}} \\
\rowcolor{gray!20}
\hline
\textbf{Variant} & \textbf{C-FID} & \textbf{Corr.} & \textbf{Disc.} & \textbf{Pred.} & \textbf{C-FID} & \textbf{Corr.} & \textbf{Disc.} & \textbf{Pred.} & \textbf{C-FID} & \textbf{Corr.} & \textbf{Disc.} & \textbf{Pred.} & \textbf{C-FID} & \textbf{Corr.} & \textbf{Disc.} & \textbf{Pred.} \\
\hline
\textsc{no-alt-warmup} & 0.185 & \secondm{0.008} & \secondm{0.148} & \bestm{0.036} & 0.093 & 0.869 & 0.235 & 0.247 & 0.166 & 0.083 & 0.035 & \secondm{0.111} & 0.257 & 1.023 & 0.134 & \secondm{0.084} \\
\textsc{no-warmup} & \secondm{0.095} & \bestm{0.007} & \bestm{0.120} & \bestm{0.036} & \secondm{0.066} & \secondm{0.774} & \secondm{0.200} & \secondm{0.246} & \secondm{0.073} & \secondm{0.063} & \secondm{0.027} & \bestm{0.108} & \secondm{0.198} & \secondm{0.900} & \secondm{0.086} & \secondm{0.084} \\
LVQMark & \bestm{0.079} & 0.009 & 0.179 & \bestm{0.036} & \bestm{0.033} & \bestm{0.740} & \bestm{0.190} & \bestm{0.244} & \bestm{0.042} & \bestm{0.050} & \bestm{0.021} & \bestm{0.108} & \bestm{0.127} & \bestm{0.896} & \bestm{0.060} & \bestm{0.082} \\
\hline
\hline
\end{tabular}%
}
\end{table*}

%==================
\paragraph{Effect of local tokenization on robust detection}

Tables~\ref{tab:robust_encoder_24} and~\ref{tab:robust_encoder_128} extend the analysis of Section~\ref{sec:local_tokenization} to other sequence lengths by comparing models with and without robust encoder. These results show that robust re-encoding alone is insufficient without a locally recoverable token representation.

\begin{table*}[h!]
\centering
\setlength{\tabcolsep}{3pt}
\renewcommand{\arraystretch}{1.13}
\caption{Results of watermark detection under attack for 24-length sequences. LVQMark is applied to different variants, and watermark detection performance (TPR) is evaluated under 30\% attacks. Best results are in bold, and second-best are underlined.}
\label{tab:robust_encoder_24}
\resizebox{\textwidth}{!}{%
\begin{tabular}{c|c|ccc|ccc|ccc|ccc}
\hline
\hline
\rowcolor{gray!20}
 &  & \multicolumn{3}{c|}{\textbf{Stocks}} & \multicolumn{3}{c|}{\textbf{Energy}} & \multicolumn{3}{c|}{\textbf{ETTh}} & \multicolumn{3}{c}{\textbf{fMRI}} \\
\rowcolor{gray!20}
\hline
\textbf{Model} & \textbf{Type} & \textbf{Offset} & \textbf{Crop} & \textbf{Insert} & \textbf{Offset} & \textbf{Crop} & \textbf{Insert} & \textbf{Offset} & \textbf{Crop} & \textbf{Insert} & \textbf{Offset} & \textbf{Crop} & \textbf{Insert} \\
\hline
\multirow{2}{*}{SDformer} & w/o Robust & \secondm{0.13} & \secondm{0.81} & 0.22 & \bestm{1.00} & \secondm{0.76} & \bestm{1.00} & \secondm{0.98} & 0.03 & 0.07 & \bestm{1.00} & \secondm{0.99} & \bestm{1.00} \\
 & w/ Robust & 0.07 & 0.20 & 0.08 & \bestm{1.00} & 0.68 & \bestm{1.00} & \bestm{1.00} & \secondm{0.57} & \secondm{0.96} & \bestm{1.00} & \bestm{1.00} & \bestm{1.00} \\
\hline
\multirow{2}{*}{L-VQVAE} & w/o Robust & \bestm{1.00} & 0.21 & \secondm{0.35} & \bestm{1.00} & 0.00 & \bestm{1.00} & \bestm{1.00} & 0.01 & \bestm{1.00} & \bestm{1.00} & \bestm{1.00} & \bestm{1.00} \\
 & w/ Robust & \bestm{1.00} & \bestm{0.99} & \bestm{0.98} & \bestm{1.00} & \bestm{1.00} & \bestm{1.00} & \bestm{1.00} & \bestm{1.00} & \bestm{1.00} & \bestm{1.00} & \bestm{1.00} & \bestm{1.00} \\
\hline
\hline
\end{tabular}%
}
\end{table*}

\begin{table*}[h!]
\centering
\setlength{\tabcolsep}{3pt}
\renewcommand{\arraystretch}{1.13}
\caption{Results of watermark detection under attack for 128-length sequences. LVQMark is applied to different variants, and watermark detection performance (TPR) is evaluated under 30\% attacks. Best results are in bold, and second-best are underlined.}
\label{tab:robust_encoder_128}
\resizebox{\textwidth}{!}{%
\begin{tabular}{c|c|ccc|ccc|ccc|ccc}
\hline
\hline
\rowcolor{gray!20}
 &  & \multicolumn{3}{c|}{\textbf{Stocks}} & \multicolumn{3}{c|}{\textbf{Energy}} & \multicolumn{3}{c|}{\textbf{ETTh}} & \multicolumn{3}{c}{\textbf{fMRI}} \\
\rowcolor{gray!20}
\hline
\textbf{Model} & \textbf{Type} & \textbf{Offset} & \textbf{Crop} & \textbf{Insert} & \textbf{Offset} & \textbf{Crop} & \textbf{Insert} & \textbf{Offset} & \textbf{Crop} & \textbf{Insert} & \textbf{Offset} & \textbf{Crop} & \textbf{Insert} \\
\hline
\multirow{2}{*}{SDformer} & w/o Robust & \secondm{0.00} & \secondm{0.00} & \secondm{0.00} & \bestm{1.00} & 0.00 & 0.32 & 0.14 & 0.04 & 0.04 & \secondm{0.10} & \secondm{0.02} & \secondm{0.03} \\
 & w/ Robust & \secondm{0.00} & \secondm{0.00} & \secondm{0.00} & \bestm{1.00} & \secondm{0.20} & \secondm{0.85} & \secondm{0.41} & \secondm{0.11} & \secondm{0.29} & 0.05 & 0.01 & 0.01 \\
\hline
\multirow{2}{*}{L-VQVAE} & w/o Robust & \bestm{1.00} & 0 & 0.99 & 1.00 & 0.00 & 0.81 & 1.00 & 0.00 & 1.00 & 1.00 & 1.00 & 0.62 \\
 & w/ Robust & \bestm{1.00} & \bestm{1.00} & \bestm{1.00} & \bestm{1.00} & \bestm{0.99} & \bestm{1.00} & \bestm{1.00} & \bestm{1.00} & \bestm{1.00} & \bestm{1.00} & \bestm{1.00} & \bestm{1.00} \\
\hline
\hline
\end{tabular}%
}
\end{table*}

\section{Experimental Results Under Mild Attack}
\label{app:mild_attack}

While the main paper reports results under the stronger 30\% attack setting, this section provides the corresponding results under milder 5\% attacks. We include the full detection and quality results, the robust-encoder ablations in order to examine whether the main observations are specific to strong perturbations. Overall, the results are broadly consistent with the main findings: although the attack-induced distortion is often smaller than in the 30\% setting, the relative pattern across methods remains largely unchanged.

\subsection{Watermark Detection}
Tables~\ref{tab:main_24_5pct}--\ref{tab:main_128_5pct} show that the qualitative conclusions of the main paper remain unchanged under the milder 5\% attack setting. Compared with the 30\% setting, the attack-induced distortion is generally smaller in magnitude, but the relative pattern across methods remains largely the same. Across sequence lengths, L-VQVAE with LVQMark continues to achieve strong watermark detection on watermarked samples while keeping the false-positive rate on attacked non-watermarked samples consistently low. The corresponding non-watermarked $z$-scores also remain close to the intended null region, indicating that the detector also stays well calibrated under mild post-editing attacks. By contrast, the baseline methods still exhibit post-editing-dependent instability, showing that the calibration problem is not limited to severe attacks but can already arise under small perturbations. Overall, these results suggest that mild attacks reduce, but do not eliminate, the robustness gap observed in the 30\% setting.

\begin{table*}[h!]
\centering
\setlength{\tabcolsep}{3pt}
\renewcommand{\arraystretch}{1.13}
\caption{Results of synthetic time series watermark detection and quality. Watermarked (TPR) and Non-watermarked (FPR, mean $z$-score) detections are evaluated under 5\% attacks. Quality metrics are for 24-length. Best results are in bold, and second-best are underlined. $z$-scores with $|z|>3.09$ are marked in \textcolor{red}{red}.}
\label{tab:main_24_5pct}
\resizebox{\textwidth}{!}{%
\begin{tabular}{c|l|l|ccc|cr|cr|cr|cccc}
\hline
\hline
\rowcolor{gray!20}
\multicolumn{3}{c|}{\textbf{Setting}}
& \multicolumn{3}{c|}{\textbf{Watermark (TPR $\uparrow$)}}
& \multicolumn{6}{c|}{\textbf{Non-watermarked (FPR $\downarrow$ | $z$-score)}}
& \multicolumn{4}{c}{\textbf{Quality Metric ($\downarrow$)}} \\
\hline
\rowcolor{gray!20}
\multicolumn{1}{c|}{\textbf{Dataset}}& \multicolumn{1}{c|}{\textbf{Model}} & \multicolumn{1}{c|}{\textbf{Method}}
& \textbf{Offset} & \textbf{Crop} & \textbf{Insert}
& \multicolumn{2}{c|}{\textbf{Offset}}
& \multicolumn{2}{c|}{\textbf{Crop}}
& \multicolumn{2}{c|}{\textbf{Insert}}
 & \textbf{C-FID} & \textbf{Corr.} & \textbf{Disc.} & \textbf{Pred.}  \\
\hline
\multirow{5}{*}{{Stocks}} & \multirow{3}{*}{DiffusionTS} & TR & 0.00 & \bestm{1.00} & 0.00 & \bestm{0.00} & $+0.06$ & 1.00 & \textcolor{red}{$+52.15$} & \bestm{0.00} & $+0.41$ & 0.96 & 0.10 & 0.20 & \bestm{0.04}  \\
 & &  GS & \bestm{1.00} & \bestm{1.00} & \bestm{1.00} & \bestm{0.00} & $-0.27$ & 1.00 & \textcolor{red}{$+22.14$} & 0.43 & $+2.88$ & 8.88 & 0.09 & 0.43 & \bestm{0.04}  \\
 & &  TimeWak & \bestm{1.00} & \bestm{1.00} & \bestm{1.00} & \bestm{0.00} & $-0.20$ & 1.00 & \textcolor{red}{$+9.36$} & \bestm{0.00} & $+0.58$ & 0.33 & \secondm{0.02} & 0.16 & \bestm{0.04}  \\
\cline{2-16}
 & SDformer & LVQMark & 0.19 & 0.44 & 0.72 & \bestm{0.00} & $+0.19$ & 0.53 & \textcolor{red}{$+3.11$} & \bestm{0.00} & $+0.22$ & \secondm{0.11} & \bestm{0.01} & \bestm{0.02} & \bestm{0.04}  \\
\cline{2-16}
 & L-VQVAE & LVQMark & \bestm{1.00} & \bestm{1.00} & \bestm{1.00} & \bestm{0.00} & $+0.11$ & \bestm{0.02} & $+0.43$ & \bestm{0.00} & $+0.09$ & \bestm{0.07} & \bestm{0.01} & \secondm{0.11} & \bestm{0.04}  \\
\hline
\hline
\multirow{5}{*}{{ETTh}} & \multirow{3}{*}{DiffusionTS} & TR & \bestm{1.00} & \bestm{1.00} & \bestm{1.00} & \bestm{0.00} & $+0.19$ & 1.00 & \textcolor{red}{$+10.11$} & 1.00 & \textcolor{red}{$+4.78$} & 1.56 & 0.18 & 0.27 & \secondm{0.14}  \\
 & &  GS & \bestm{1.00} & \bestm{1.00} & \bestm{1.00} & 0.18 & $+2.17$ & \bestm{0.00} & \textcolor{red}{$-10.53$} & 0.14 & $+2.02$ & 4.66 & 0.42 & 0.38 & 0.18  \\
 & &  TimeWak & \bestm{1.00} & \bestm{1.00} & \bestm{1.00} & \bestm{0.00} & $-0.26$ & \bestm{0.00} & $+0.77$ & \bestm{0.00} & $-0.34$ & 0.23 & 0.21 & 0.08 & \bestm{0.12}  \\
\cline{2-16}
 & SDformer & LVQMark & \bestm{1.00} & 0.01 & \bestm{1.00} & \bestm{0.00} & $+0.15$ & \bestm{0.00} & $+0.10$ & \bestm{0.00} & $+0.29$ & \secondm{0.13} & \secondm{0.08} & \secondm{0.04} & \bestm{0.12}  \\
\cline{2-16}
 & L-VQVAE & LVQMark & \bestm{1.00} & \bestm{1.00} & \bestm{1.00} & \bestm{0.00} & $+0.28$ & \bestm{0.00} & $+0.13$ & \bestm{0.00} & $-0.14$ & \bestm{0.05} & \bestm{0.05} & \bestm{0.02} & \bestm{0.12}  \\
\hline
\hline
\multirow{5}{*}{{Energy}} & \multirow{3}{*}{DiffusionTS} & TR & 0.00 & \bestm{1.00} & \bestm{1.00} & \bestm{0.00} & $+0.03$ & 1.00 & \textcolor{red}{$+36.66$} & 1.00 & \textcolor{red}{$+9.01$} & 0.43 & 2.66 & 0.41 & \secondm{0.30}  \\
 & &  GS & \bestm{1.00} & \bestm{1.00} & \bestm{1.00} & 0.59 & \textcolor{red}{$+3.33$} & 1.00 & \textcolor{red}{$+43.84$} & 0.48 & \textcolor{red}{$+3.17$} & 1.58 & 3.37 & 0.49 & 0.33  \\
 & &  TimeWak & \bestm{1.00} & \bestm{1.00} & \bestm{1.00} & \bestm{0.00} & $-1.43$ & 1.00 & \textcolor{red}{$+5.79$} & \bestm{0.00} & $-0.44$ & \secondm{0.09} & 1.53 & \secondm{0.14} & \bestm{0.25}  \\
\cline{2-16}
 & SDformer & LVQMark & \bestm{1.00} & \bestm{1.00} & \bestm{1.00} & \bestm{0.00} & $-0.22$ & \bestm{0.00} & $+0.15$ & \bestm{0.00} & $+0.42$ & 0.12 & \secondm{1.27} & 0.23 & \bestm{0.25}  \\
\cline{2-16}
 & L-VQVAE & LVQMark & \bestm{1.00} & \bestm{1.00} & \bestm{1.00} & \bestm{0.00} & $-0.11$ & 0.01 & $-0.02$ & \bestm{0.00} & $-0.05$ & \bestm{0.03} & \bestm{1.02} & \bestm{0.09} & \bestm{0.25}  \\
\hline
\hline
\multirow{5}{*}{{fMRI}} & \multirow{3}{*}{DiffusionTS} & TR & \bestm{1.00} & 0.00 & 0.00 & \bestm{0.00} & $+0.06$ & \bestm{0.00} & $+0.20$ & \bestm{0.00} & $+0.70$ & 2.26 & 13.42 & 0.50 & 0.15  \\
 & &  GS & \bestm{1.00} & \bestm{1.00} & \bestm{1.00} & \bestm{0.00} & $+0.58$ & 0.20 & $+2.37$ & \bestm{0.00} & $-0.42$ & 0.71 & 15.21 & 0.50 & \secondm{0.11}  \\
 & &  TimeWak & \bestm{1.00} & \bestm{1.00} & \bestm{1.00} & \bestm{0.00} & $-0.09$ & \bestm{0.00} & $-1.07$ & \bestm{0.00} & $-0.28$ & \bestm{0.18} & \secondm{1.98} & \bestm{0.08} & \bestm{0.10}  \\
\cline{2-16}
 & SDformer & LVQMark & \bestm{1.00} & \bestm{1.00} & \bestm{1.00} & \bestm{0.00} & $-0.08$ & \bestm{0.00} & $-0.36$ & \bestm{0.00} & $-0.07$ & 0.94 & 3.24 & \secondm{0.21} & \bestm{0.10}  \\
\cline{2-16}
 & L-VQVAE & LVQMark & \bestm{1.00} & \bestm{1.00} & \bestm{1.00} & \bestm{0.00} & $+0.07$ & \bestm{0.00} & $-0.11$ & \bestm{0.00} & $-0.14$ & \secondm{0.20} & \bestm{1.92} & 0.23 & \bestm{0.10}  \\
\hline
\hline
\end{tabular}%
}
\end{table*}

\begin{table*}[h!]
\centering
\setlength{\tabcolsep}{3pt}
\renewcommand{\arraystretch}{1.13}
\caption{Results of synthetic time series watermark detection and quality. Watermarked (TPR) and Non-watermarked (FPR, mean $z$-score) detections are evaluated under 5\% attacks. Quality metrics are for 64-length. Best results are in bold, and second-best are underlined. $z$-scores with $|z|>3.09$ are marked in \textcolor{red}{red}.}
\label{tab:main_64_5pct}
\resizebox{\textwidth}{!}{%
\begin{tabular}{c|l|l|ccc|cr|cr|cr|cccc}
\hline
\hline
\rowcolor{gray!20}
\multicolumn{3}{c|}{\textbf{Setting}}
& \multicolumn{3}{c|}{\textbf{Watermark (TPR $\uparrow$)}}
& \multicolumn{6}{c|}{\textbf{Non-watermarked (FPR $\downarrow$ | $z$-score)}}
& \multicolumn{4}{c}{\textbf{Quality Metric ($\downarrow$)}} \\
\hline
\rowcolor{gray!20}
\multicolumn{1}{c|}{\textbf{Dataset}}& \multicolumn{1}{c|}{\textbf{Model}} & \multicolumn{1}{c|}{\textbf{Method}}
& \textbf{Offset} & \textbf{Crop} & \textbf{Insert}
& \multicolumn{2}{c|}{\textbf{Offset}}
& \multicolumn{2}{c|}{\textbf{Crop}}
& \multicolumn{2}{c|}{\textbf{Insert}}
 & \textbf{C-FID} & \textbf{Corr.} & \textbf{Disc.} & \textbf{Pred.}  \\
\hline
\multirow{5}{*}{{Stocks}} & \multirow{3}{*}{DiffusionTS} & TR & 0.00 & \bestm{1.00} & 0.00 & \bestm{0.00} & $+0.06$ & 1.00 & \textcolor{red}{$+55.65$} & \bestm{0.00} & $+2.06$ & 1.52 & 0.07 & 0.15 & \bestm{0.04}  \\
 & &  GS & \bestm{1.00} & \bestm{1.00} & \bestm{1.00} & \bestm{0.00} & $-0.35$ & \bestm{0.00} & \textcolor{red}{$-4.62$} & 0.05 & $+1.55$ & 1.50 & \secondm{0.02} & 0.22 & \bestm{0.04}  \\
 & &  TimeWak & \bestm{1.00} & \bestm{1.00} & \bestm{1.00} & \bestm{0.00} & $-0.12$ & 0.29 & $+2.56$ & \bestm{0.00} & $+0.11$ & 0.29 & \bestm{0.01} & 0.13 & \bestm{0.04}  \\
\cline{2-16}
 & SDformer & LVQMark & 0.01 & 0.03 & 0.01 & \bestm{0.00} & $-0.17$ & \bestm{0.00} & $+0.58$ & 0.01 & $+0.32$ & \secondm{0.08} & \bestm{0.01} & \bestm{0.01} & \bestm{0.04}  \\
\cline{2-16}
 & L-VQVAE & LVQMark & \bestm{1.00} & \bestm{1.00} & \bestm{1.00} & \bestm{0.00} & $-0.33$ & \bestm{0.00} & $+0.08$ & \bestm{0.00} & $+0.30$ & \bestm{0.07} & \bestm{0.01} & \secondm{0.06} & \bestm{0.04}  \\
\hline
\hline
\multirow{5}{*}{{ETTh}} & \multirow{3}{*}{DiffusionTS} & TR & \bestm{1.00} & \bestm{1.00} & \bestm{1.00} & \bestm{0.00} & $+0.21$ & 1.00 & \textcolor{red}{$+14.18$} & 1.00 & \textcolor{red}{$+12.70$} & 2.17 & 0.22 & 0.29 & \secondm{0.14}  \\
 & &  GS & \bestm{1.00} & \bestm{1.00} & \bestm{1.00} & \bestm{0.00} & $-0.29$ & \bestm{0.00} & \textcolor{red}{$-29.58$} & \bestm{0.00} & $-1.60$ & 3.43 & 0.25 & 0.36 & 0.16  \\
 & &  TimeWak & \bestm{1.00} & \bestm{1.00} & \bestm{1.00} & 0.01 & $+0.19$ & 0.90 & \textcolor{red}{$+4.08$} & \bestm{0.00} & $+0.23$ & 0.37 & 0.13 & 0.11 & \bestm{0.12}  \\
\cline{2-16}
 & SDformer & LVQMark & 0.27 & 0.00 & 0.06 & 0.01 & $+0.25$ & \bestm{0.00} & $-0.27$ & \bestm{0.00} & $-0.11$ & \secondm{0.04} & \bestm{0.05} & \bestm{0.00} & \bestm{0.12}  \\
\cline{2-16}
 & L-VQVAE & LVQMark & \bestm{1.00} & \bestm{1.00} & \bestm{1.00} & \bestm{0.00} & $-0.17$ & \bestm{0.00} & $-0.07$ & 0.01 & $+0.22$ & \bestm{0.03} & \secondm{0.06} & \secondm{0.01} & \bestm{0.12}  \\
\hline
\hline
\multirow{5}{*}{{Energy}} & \multirow{3}{*}{DiffusionTS} & TR & 0.00 & \bestm{1.00} & \bestm{1.00} & \bestm{0.00} & $+0.05$ & 1.00 & \textcolor{red}{$+50.50$} & 1.00 & \textcolor{red}{$+21.61$} & 0.58 & 1.98 & 0.43 & \secondm{0.28}  \\
 & &  GS & \bestm{1.00} & \bestm{1.00} & \bestm{1.00} & 0.07 & $+1.21$ & 1.00 & \textcolor{red}{$+11.15$} & 0.37 & $+2.71$ & 1.78 & 2.72 & 0.48 & 0.31  \\
 & &  TimeWak & \bestm{1.00} & \bestm{1.00} & \bestm{1.00} & \bestm{0.00} & $-1.26$ & \bestm{0.00} & $-1.42$ & \bestm{0.00} & $-0.84$ & \secondm{0.14} & 1.52 & \secondm{0.14} & \bestm{0.25}  \\
\cline{2-16}
 & SDformer & LVQMark & \bestm{1.00} & 0.46 & \bestm{1.00} & \bestm{0.00} & $+0.06$ & \bestm{0.00} & $+0.23$ & \bestm{0.00} & $+0.07$ & \bestm{0.04} & \secondm{1.04} & \bestm{0.08} & \bestm{0.25}  \\
\cline{2-16}
 & L-VQVAE & LVQMark & \bestm{1.00} & \bestm{1.00} & \bestm{1.00} & \bestm{0.00} & $-0.06$ & 0.01 & $+0.02$ & 0.01 & $+0.17$ & \bestm{0.04} & \bestm{0.94} & 0.15 & \bestm{0.25}  \\
\hline
\hline
\multirow{5}{*}{{fMRI}} & \multirow{3}{*}{DiffusionTS} & TR & \bestm{1.00} & \bestm{1.00} & \bestm{1.00} & \bestm{0.00} & $+0.03$ & \bestm{0.00} & $+0.30$ & \bestm{0.00} & $+2.25$ & 3.63 & 12.83 & 0.40 & 0.14  \\
 & &  GS & \bestm{1.00} & \bestm{1.00} & \bestm{1.00} & \bestm{0.00} & $-0.55$ & 1.00 & \textcolor{red}{$+20.44$} & 1.00 & \textcolor{red}{$+8.32$} & 0.74 & 8.31 & 0.50 & \secondm{0.10}  \\
 & &  TimeWak & \bestm{1.00} & \bestm{1.00} & \bestm{1.00} & \bestm{0.00} & $+0.09$ & 0.13 & $+1.96$ & 0.01 & $+0.60$ & \secondm{0.45} & 1.87 & \secondm{0.25} & \secondm{0.10}  \\
\cline{2-16}
 & SDformer & LVQMark & 0.24 & 0.29 & 0.24 & \bestm{0.00} & $+0.04$ & 0.01 & $+0.43$ & \bestm{0.00} & $+0.10$ & \bestm{0.13} & \secondm{1.19} & \bestm{0.12} & \bestm{0.09}  \\
\cline{2-16}
 & L-VQVAE & LVQMark & \bestm{1.00} & \bestm{1.00} & \bestm{1.00} & \bestm{0.00} & $-0.18$ & \bestm{0.00} & $+0.09$ & \bestm{0.00} & $-0.04$ & \bestm{0.13} & \bestm{1.14} & 0.27 & \bestm{0.09}  \\
\hline
\hline
\end{tabular}%
}
\end{table*}

\begin{table*}[h!]
\centering
\setlength{\tabcolsep}{3pt}
\renewcommand{\arraystretch}{1.13}
\caption{Results of synthetic time series watermark detection and quality. Watermarked (TPR) and Non-watermarked (FPR, mean $z$-score) detections are evaluated under 5\% attacks. Quality metrics are for 128-length. Best results are in bold, and second-best are underlined. $z$-scores with $|z|>3.09$ are marked in \textcolor{red}{red}.}
\label{tab:main_128_5pct}
\resizebox{\textwidth}{!}{%
\begin{tabular}{c|l|l|ccc|cr|cr|cr|cccc}
\hline
\hline
\rowcolor{gray!20}
\multicolumn{3}{c|}{\textbf{Setting}}
& \multicolumn{3}{c|}{\textbf{Watermark (TPR $\uparrow$)}}
& \multicolumn{6}{c|}{\textbf{Non-watermarked (FPR $\downarrow$ | $z$-score)}}
& \multicolumn{4}{c}{\textbf{Quality Metric ($\downarrow$)}} \\
\hline
\rowcolor{gray!20}
\multicolumn{1}{c|}{\textbf{Dataset}}& \multicolumn{1}{c|}{\textbf{Model}} & \multicolumn{1}{c|}{\textbf{Method}}
& \textbf{Offset} & \textbf{Crop} & \textbf{Insert}
& \multicolumn{2}{c|}{\textbf{Offset}}
& \multicolumn{2}{c|}{\textbf{Crop}}
& \multicolumn{2}{c|}{\textbf{Insert}}
 & \textbf{C-FID} & \textbf{Corr.} & \textbf{Disc.} & \textbf{Pred.}  \\
\hline
\multirow{5}{*}{{Stocks}} & \multirow{3}{*}{DiffusionTS} & TR & 0.00 & \bestm{1.00} & \bestm{1.00} & \bestm{0.00} & $+0.09$ & 1.00 & \textcolor{red}{$+66.42$} & 1.00 & \textcolor{red}{$+5.55$} & 3.05 & 0.09 & 0.24 & \bestm{0.04}  \\
 & &  GS & \bestm{1.00} & \bestm{1.00} & \bestm{1.00} & \bestm{0.00} & $-1.32$ & \bestm{0.00} & \textcolor{red}{$-3.91$} & \bestm{0.00} & $-0.47$ & 2.63 & \secondm{0.04} & 0.19 & \bestm{0.04}  \\
 & &  TimeWak & \bestm{1.00} & \bestm{1.00} & \bestm{1.00} & \bestm{0.00} & $-0.17$ & 1.00 & \textcolor{red}{$+11.97$} & 0.16 & $+2.05$ & 0.34 & \bestm{0.01} & \secondm{0.15} & \bestm{0.04}  \\
\cline{2-16}
 & SDformer & LVQMark & 0.00 & 0.00 & 0.00 & \bestm{0.00} & $-0.82$ & \bestm{0.00} & $-0.22$ & \bestm{0.00} & $-0.02$ & \secondm{0.12} & \bestm{0.01} & \bestm{0.03} & \bestm{0.04}  \\
\cline{2-16}
 & L-VQVAE & LVQMark & \bestm{1.00} & \bestm{1.00} & \bestm{1.00} & \bestm{0.00} & $+0.02$ & \bestm{0.00} & $+0.16$ & 0.01 & $+0.15$ & \bestm{0.08} & \bestm{0.01} & 0.18 & \bestm{0.04}  \\
\hline
\hline
\multirow{5}{*}{{ETTh}} & \multirow{3}{*}{DiffusionTS} & TR & 0.00 & \bestm{1.00} & \bestm{1.00} & \bestm{0.00} & $+0.28$ & 1.00 & \textcolor{red}{$+11.00$} & 1.00 & \textcolor{red}{$+11.88$} & 2.52 & 0.26 & 0.30 & \secondm{0.13}  \\
 & &  GS & \bestm{1.00} & \bestm{1.00} & \bestm{1.00} & \bestm{0.00} & $-0.93$ & \bestm{0.00} & \textcolor{red}{$-13.63$} & \bestm{0.00} & $-1.08$ & 5.59 & 0.23 & 0.39 & 0.15  \\
 & &  TimeWak & \bestm{1.00} & \bestm{1.00} & \bestm{1.00} & \bestm{0.00} & $+0.03$ & 0.02 & $+0.94$ & \bestm{0.00} & $+0.40$ & \secondm{1.08} & \secondm{0.17} & 0.15 & \bestm{0.11}  \\
\cline{2-16}
 & SDformer & LVQMark & 0.52 & 0.03 & 0.10 & \bestm{0.00} & $-0.07$ & 0.03 & $+1.19$ & \bestm{0.00} & $+0.53$ & \bestm{0.04} & \bestm{0.05} & \secondm{0.03} & \bestm{0.11}  \\
\cline{2-16}
 & L-VQVAE & LVQMark & \bestm{1.00} & \bestm{1.00} & \bestm{1.00} & \bestm{0.00} & $-0.10$ & \bestm{0.00} & $+0.16$ & \bestm{0.00} & $+0.04$ & \bestm{0.04} & \bestm{0.05} & \bestm{0.02} & \bestm{0.11}  \\
\hline
\hline
\multirow{5}{*}{{Energy}} & \multirow{3}{*}{DiffusionTS} & TR & 0.00 & \bestm{1.00} & \bestm{1.00} & \bestm{0.00} & $+0.04$ & 1.00 & \textcolor{red}{$+58.72$} & 1.00 & \textcolor{red}{$+37.10$} & 0.50 & 1.65 & 0.49 & 0.28  \\
 & &  GS & \bestm{1.00} & \bestm{1.00} & \bestm{1.00} & \bestm{0.00} & $+0.84$ & 1.00 & \textcolor{red}{$+32.87$} & 1.00 & \textcolor{red}{$+6.63$} & 3.34 & 3.59 & 0.48 & 0.29  \\
 & &  TimeWak & \bestm{1.00} & \bestm{1.00} & \bestm{1.00} & \bestm{0.00} & $-0.82$ & \bestm{0.00} & $-1.89$ & \bestm{0.00} & $+0.58$ & 0.17 & 1.50 & 0.24 & \secondm{0.25}  \\
\cline{2-16}
 & SDformer & LVQMark & \bestm{1.00} & 0.14 & \bestm{1.00} & \bestm{0.00} & $-0.06$ & \bestm{0.00} & $-0.61$ & \bestm{0.00} & $+0.28$ & \secondm{0.04} & \bestm{0.67} & \bestm{0.05} & \secondm{0.25}  \\
\cline{2-16}
 & L-VQVAE & LVQMark & \bestm{1.00} & \bestm{1.00} & \bestm{1.00} & \bestm{0.00} & $+0.15$ & \bestm{0.00} & $-0.09$ & \bestm{0.00} & $-0.01$ & \bestm{0.03} & \secondm{0.74} & \secondm{0.19} & \bestm{0.24}  \\
\hline
\hline
\multirow{5}{*}{{fMRI}} & \multirow{3}{*}{DiffusionTS} & TR & 0.00 & 0.00 & \bestm{1.00} & \bestm{0.00} & $+0.02$ & \bestm{0.00} & $+1.37$ & 1.00 & \textcolor{red}{$+3.98$} & 4.57 & 14.15 & 0.45 & 0.17  \\
 & &  GS & \bestm{1.00} & \bestm{1.00} & \bestm{1.00} & \bestm{0.00} & $+0.04$ & \bestm{0.00} & \textcolor{red}{$-38.23$} & 0.29 & $+2.54$ & 1.05 & 5.97 & 0.50 & 0.11  \\
 & &  TimeWak & \bestm{1.00} & \bestm{1.00} & \bestm{1.00} & \bestm{0.00} & $-0.08$ & \bestm{0.00} & $-0.90$ & \bestm{0.00} & $-0.47$ & 0.81 & 1.81 & 0.36 & \secondm{0.10}  \\
\cline{2-16}
 & SDformer & LVQMark & 0.10 & 0.12 & 0.11 & \bestm{0.00} & $+0.06$ & \bestm{0.00} & $+0.02$ & \bestm{0.00} & $-0.00$ & \bestm{0.09} & \bestm{0.84} & \secondm{0.16} & \bestm{0.08}  \\
\cline{2-16}
 & L-VQVAE & LVQMark & \bestm{1.00} & \bestm{1.00} & \bestm{1.00} & \bestm{0.00} & $-0.10$ & \bestm{0.00} & $+0.25$ & \bestm{0.00} & $-0.03$ & \secondm{0.13} & \secondm{0.90} & \bestm{0.06} & \bestm{0.08}  \\
\hline
\hline
\end{tabular}%
}
\end{table*}

\subsection{Effect of Local Tokenization on Robust Detection}
Tables~\ref{tab:robust_encoder_24_5pct}--\ref{tab:robust_encoder_128_5pct} further examine whether robust re-encoding alone is sufficient in the mild-attack regime. The results are consistent with the main-paper ablation: adding a robust encoder can partially improve detection in some settings, but the gains remain limited and inconsistent when the underlying representation is not locally recoverable. By contrast, when robust re-encoding is combined with the local tokenization of L-VQVAE, detection performance remains uniformly strong across datasets, attacks, and sequence lengths. This again supports our central claim that robust recovery is not solely a matter of attack-aware training, but depends fundamentally on a representation whose recovered units remain locally stable under perturbation.

\begin{table*}[h!]
\centering
\setlength{\tabcolsep}{3pt}
\renewcommand{\arraystretch}{1.13}
\caption{Results of watermark detection under attack for 24-length sequences. LVQMark is applied to different variants, and watermark detection performance (TPR) is evaluated under 5\% attacks. Best results are in bold, and second-best are underlined.}
\label{tab:robust_encoder_24_5pct}
\resizebox{\textwidth}{!}{%
\begin{tabular}{c|c|ccc|ccc|ccc|ccc}
\hline
\hline
\rowcolor{gray!20}
 &  & \multicolumn{3}{c|}{\textbf{Stocks}} & \multicolumn{3}{c|}{\textbf{Energy}} & \multicolumn{3}{c|}{\textbf{ETTh}} & \multicolumn{3}{c}{\textbf{fMRI}} \\
\rowcolor{gray!20}
\hline
\textbf{Model} & \textbf{Type} & \textbf{Offset} & \textbf{Crop} & \textbf{Insert} & \textbf{Offset} & \textbf{Crop} & \textbf{Insert} & \textbf{Offset} & \textbf{Crop} & \textbf{Insert} & \textbf{Offset} & \textbf{Crop} & \textbf{Insert} \\
\hline
\multirow{2}{*}{SDformer} & w/o Robust & \secondm{0.19} & 0.44 & 0.72 & \bestm{1.00} & \bestm{1.00} & \bestm{1.00} & \bestm{1.00} & 0.01 & \bestm{1.00} & \bestm{1.00} & \bestm{1.00} & \bestm{1.00} \\
 & w/ Robust & 0.08 & 0.57 & 0.24 & \bestm{1.00} & \bestm{1.00} & \bestm{1.00} & \bestm{1.00} & \bestm{1.00} & \bestm{1.00} & \bestm{1.00} & \bestm{1.00} & \bestm{1.00} \\
\hline
\multirow{2}{*}{L-VQVAE} & w/o Robust & \bestm{1.00} & \secondm{0.78} & \secondm{0.86} & \bestm{1.00} & \secondm{0.76} & \bestm{1.00} & \bestm{1.00} & \secondm{0.90} & \bestm{1.00} & \bestm{1.00} & \bestm{1.00} & \bestm{1.00} \\
 & w/ Robust & \bestm{1.00} & \bestm{1.00} & \bestm{1.00} & \bestm{1.00} & \bestm{1.00} & \bestm{1.00} & \bestm{1.00} & \bestm{1.00} & \bestm{1.00} & \bestm{1.00} & \bestm{1.00} & \bestm{1.00} \\
\hline
\hline
\end{tabular}%
}
\end{table*}

\begin{table*}[h!]
\centering
\setlength{\tabcolsep}{3pt}
\renewcommand{\arraystretch}{1.13}
\caption{Results of watermark detection under attack for 64-length sequences. LVQMark is applied to different variants, and watermark detection performance (TPR) is evaluated under 5\% attacks. Best results are in bold, and second-best are underlined.}
\label{tab:robust_encoder_64_5pct}
\resizebox{\textwidth}{!}{%
\begin{tabular}{c|c|ccc|ccc|ccc|ccc}
\hline
\hline
\rowcolor{gray!20}
 &  & \multicolumn{3}{c|}{\textbf{Stocks}} & \multicolumn{3}{c|}{\textbf{Energy}} & \multicolumn{3}{c|}{\textbf{ETTh}} & \multicolumn{3}{c}{\textbf{fMRI}} \\
\rowcolor{gray!20}
\hline
\textbf{Model} & \textbf{Type} & \textbf{Offset} & \textbf{Crop} & \textbf{Insert} & \textbf{Offset} & \textbf{Crop} & \textbf{Insert} & \textbf{Offset} & \textbf{Crop} & \textbf{Insert} & \textbf{Offset} & \textbf{Crop} & \textbf{Insert} \\
\hline
\multirow{2}{*}{SDformer} & w/o Robust & \secondm{0.01} & \secondm{0.03} & \secondm{0.01} & \bestm{1.00} & 0.46 & \bestm{1.00} & \secondm{0.27} & 0.00 & 0.06 & \secondm{0.24} & \secondm{0.29} & \secondm{0.24} \\
 & w/ Robust & 0.00 & 0.00 & 0.00 & \bestm{1.00} & \secondm{0.95} & \bestm{1.00} & 0.14 & 0.09 & \secondm{0.10} & 0.04 & 0.12 & 0.10 \\
\hline
\multirow{2}{*}{L-VQVAE} & w/o Robust & \bestm{1.00} & 0.00 & \bestm{1.00} & \bestm{1.00} & 0.03 & \bestm{1.00} & \bestm{1.00} & \secondm{0.93} & \bestm{1.00} & \bestm{1.00} & \bestm{1.00} & \bestm{1.00} \\
 & w/ Robust & \bestm{1.00} & \bestm{1.00} & \bestm{1.00} & \bestm{1.00} & \bestm{1.00} & \bestm{1.00} & \bestm{1.00} & \bestm{1.00} & \bestm{1.00} & \bestm{1.00} & \bestm{1.00} & \bestm{1.00} \\
\hline
\hline
\end{tabular}%
}
\end{table*}

\begin{table*}[h!]
\centering
\setlength{\tabcolsep}{3pt}
\renewcommand{\arraystretch}{1.13}
\caption{Results of watermark detection under attack for 128-length sequences. LVQMark is applied to different variants, and watermark detection performance (TPR) is evaluated under 5\% attacks. Best results are in bold, and second-best are underlined.}
\label{tab:robust_encoder_128_5pct}
\resizebox{\textwidth}{!}{%
\begin{tabular}{c|c|ccc|ccc|ccc|ccc}
\hline
\hline
\rowcolor{gray!20}
 &  & \multicolumn{3}{c|}{\textbf{Stocks}} & \multicolumn{3}{c|}{\textbf{Energy}} & \multicolumn{3}{c|}{\textbf{ETTh}} & \multicolumn{3}{c}{\textbf{fMRI}} \\
\rowcolor{gray!20}
\hline
\textbf{Model} & \textbf{Type} & \textbf{Offset} & \textbf{Crop} & \textbf{Insert} & \textbf{Offset} & \textbf{Crop} & \textbf{Insert} & \textbf{Offset} & \textbf{Crop} & \textbf{Insert} & \textbf{Offset} & \textbf{Crop} & \textbf{Insert} \\
\hline
\multirow{2}{*}{SDformer} & w/o Robust & \secondm{0.00} & 0.00 & \secondm{0.00} & \bestm{1.00} & 0.14 & \bestm{1.00} & \secondm{0.52} & 0.03 & 0.10 & \secondm{0.10} & \secondm{0.12} & \secondm{0.11} \\
 & w/ Robust & \secondm{0.00} & 0.00 & \secondm{0.00} & \bestm{1.00} & \secondm{0.85} & \bestm{1.00} & 0.34 & 0.31 & \secondm{0.35} & 0.06 & 0.04 & 0.02 \\
\hline
\multirow{2}{*}{L-VQVAE} & w/o Robust & \bestm{1.00} & \secondm{0.94} & \bestm{1.00} & \bestm{1.00} & 0.02 & \bestm{1.00} & \bestm{1.00} & \secondm{0.50} & \bestm{1.00} & \bestm{1.00} & \bestm{1.00} & \bestm{1.00} \\
 & w/ Robust & \bestm{1.00} & \bestm{1.00} & \bestm{1.00} & \bestm{1.00} & \bestm{1.00} & \bestm{1.00} & \bestm{1.00} & \bestm{1.00} & \bestm{1.00} & \bestm{1.00} & \bestm{1.00} & \bestm{1.00} \\
\hline
\hline
\end{tabular}%
}
\end{table*}

\section{Detailed Watermark Detection and Quality Statistics}
\label{statisticalanalysis}

Tables~\ref{tab:main_24_std}--\ref{tab:main_128_std} complement the main results by reporting both the mean and the standard deviation of the detection and quality metrics. For detection, the reported standard deviations summarize the spread of sample-level $z$-scores under each setting, showing how tightly the detection statistics concentrate around their mean values. For quality, the reported standard deviations reflect variability across the same five-run protocol used in the main experiments. These additional statistics support the same qualitative conclusions as the mean-only results. In particular, L-VQVAE with LVQMark keeps the non-watermarked $z$-scores centered near the intended null region across attacks and sequence lengths, while the corresponding standard deviations remain moderate, indicating stable calibration at the sample level rather than a mean obtained from offsetting extreme cases. At the same time, the watermark $z$-scores remain clearly positive and well separated from zero, showing that strong detectability is preserved together with null stability. By contrast, several baselines continue to exhibit large mean shifts on attacked non-watermarked samples, and these shifts are typically much larger in magnitude than their corresponding standard deviations, indicating that the observed null drift is systematic rather than attributable to a small subset of atypical samples. Taken together, these results reinforce the main finding of the paper: the advantage of L-VQVAE with LVQMark lies not only in achieving favorable mean performance, but also in maintaining a more stable and interpretable detection behavior under attack.

\begin{table*}[h!]
\centering
\setlength{\tabcolsep}{3pt}
\renewcommand{\arraystretch}{1.13}
\caption{Results of synthetic time series watermark detection and quality. Watermark and Non-watermarked detections ($z$-score) are evaluated under 30\% attacks. Detection and quality metrics (mean $\pm$ std) are for 24-length.}
\label{tab:main_24_std}
\resizebox{\textwidth}{!}{%
\begin{tabular}{c|l|l|r|r|r|r|r|r|r|r|r|r}
\hline
\hline
\rowcolor{gray!20}
\multicolumn{3}{c|}{\textbf{Setting}}
& \multicolumn{3}{c|}{\textbf{Watermark ($z$-score $\uparrow$)}}
& \multicolumn{3}{c|}{\textbf{Non-watermarked ($z$-score)}}
& \multicolumn{4}{c}{\textbf{Quality Metric ($\downarrow$)}} \\
\hline
\rowcolor{gray!20}
\textbf{Dataset} & \multicolumn{1}{c|}{\textbf{Model}} & \multicolumn{1}{c|}{\textbf{Method}}
& \multicolumn{1}{c|}{\textbf{Offset}} & \multicolumn{1}{c|}{\textbf{Crop}} & \multicolumn{1}{c|}{\textbf{Insert}}
& \multicolumn{1}{c|}{\textbf{Offset}} & \multicolumn{1}{c|}{\textbf{Crop}} & \multicolumn{1}{c|}{\textbf{Insert}}
& \multicolumn{1}{c|}{\textbf{C-FID}} & \multicolumn{1}{c|}{\textbf{Corr.}} & \multicolumn{1}{c|}{\textbf{Disc.}} & \multicolumn{1}{c}{\textbf{Pred.}} \\
\hline
\multirow{5}{*}{{Stocks}} & \multirow{3}{*}{DiffusionTS} & TR
& $0.59_{\pm 0.03}$ & $51.54_{\pm 0.66}$ & $0.22_{\pm 0.04}$
& $0.19_{\pm 0.03}$ & $63.26_{\pm 0.69}$ & $1.91_{\pm 0.09}$
& $0.96_{\pm 0.15}$ & $0.10_{\pm 0.01}$ & $0.20_{\pm 0.02}$ & $0.04_{\pm 0.00}$ \\
 & & GS
& $85.77_{\pm 0.80}$ & $3.11_{\pm 0.67}$ & $52.58_{\pm 0.80}$
& $-1.15_{\pm 1.22}$ & $-6.32_{\pm 0.69}$ & $0.16_{\pm 0.81}$
& $8.88_{\pm 0.62}$ & $0.09_{\pm 0.01}$ & $0.43_{\pm 0.01}$ & $0.04_{\pm 0.00}$ \\
 & & TimeWak
& $180.18_{\pm 0.86}$ & $11.24_{\pm 0.94}$ & $56.48_{\pm 1.13}$
& $-0.99_{\pm 0.99}$ & $1.39_{\pm 0.86}$ & $-1.77_{\pm 1.03}$
& $0.33_{\pm 0.03}$ & $0.02_{\pm 0.01}$ & $0.16_{\pm 0.02}$ & $0.04_{\pm 0.00}$ \\
\cline{2-13}
 & SDformer & LVQMark
& $2.14_{\pm 0.96}$ & $4.21_{\pm 1.19}$ & $2.39_{\pm 0.94}$
& $0.51_{\pm 0.91}$ & $4.49_{\pm 1.28}$ & $1.33_{\pm 0.82}$
& $0.11_{\pm 0.02}$ & $0.01_{\pm 0.00}$ & $0.02_{\pm 0.02}$ & $0.04_{\pm 0.00}$ \\
\cline{2-13}
 & L-VQVAE & LVQMark
& $11.16_{\pm 1.12}$ & $6.97_{\pm 1.40}$ & $5.45_{\pm 0.88}$
& $-0.25_{\pm 0.85}$ & $1.10_{\pm 1.21}$ & $0.38_{\pm 1.18}$
& $0.07_{\pm 0.01}$ & $0.01_{\pm 0.01}$ & $0.11_{\pm 0.09}$ & $0.04_{\pm 0.00}$ \\
\hline
\hline
\multirow{5}{*}{{ETTh}} & \multirow{3}{*}{DiffusionTS} & TR
& $3.70_{\pm 0.10}$ & $11.62_{\pm 0.25}$ & $18.23_{\pm 0.32}$
& $0.95_{\pm 0.04}$ & $10.84_{\pm 0.22}$ & $15.97_{\pm 0.28}$
& $1.56_{\pm 0.16}$ & $0.18_{\pm 0.02}$ & $0.27_{\pm 0.01}$ & $0.14_{\pm 0.00}$ \\
 & & GS
& $53.08_{\pm 1.14}$ & $23.49_{\pm 1.00}$ & $36.63_{\pm 1.00}$
& $12.01_{\pm 0.79}$ & $18.58_{\pm 1.11}$ & $6.96_{\pm 0.94}$
& $4.66_{\pm 0.47}$ & $0.42_{\pm 0.02}$ & $0.38_{\pm 0.02}$ & $0.18_{\pm 0.01}$ \\
 & & TimeWak
& $115.86_{\pm 0.96}$ & $3.07_{\pm 0.97}$ & $19.44_{\pm 1.06}$
& $-0.34_{\pm 0.97}$ & $1.05_{\pm 0.96}$ & $-1.68_{\pm 1.00}$
& $0.23_{\pm 0.01}$ & $0.21_{\pm 0.04}$ & $0.08_{\pm 0.03}$ & $0.12_{\pm 0.00}$ \\
\cline{2-13}
 & SDformer & LVQMark
& $5.15_{\pm 0.94}$ & $0.50_{\pm 1.26}$ & $1.82_{\pm 0.89}$
& $0.79_{\pm 0.96}$ & $0.18_{\pm 1.22}$ & $0.70_{\pm 0.85}$
& $0.13_{\pm 0.02}$ & $0.08_{\pm 0.03}$ & $0.04_{\pm 0.02}$ & $0.12_{\pm 0.00}$ \\
\cline{2-13}
 & L-VQVAE & LVQMark
& $20.66_{\pm 1.12}$ & $17.01_{\pm 1.24}$ & $13.48_{\pm 0.95}$
& $0.03_{\pm 1.10}$ & $0.41_{\pm 1.45}$ & $-0.15_{\pm 1.01}$
& $0.05_{\pm 0.00}$ & $0.05_{\pm 0.01}$ & $0.02_{\pm 0.01}$ & $0.12_{\pm 0.00}$ \\
\hline
\hline
\multirow{5}{*}{{Energy}} & \multirow{3}{*}{DiffusionTS} & TR
& $0.84_{\pm 0.03}$ & $37.42_{\pm 0.25}$ & $26.12_{\pm 0.16}$
& $0.43_{\pm 0.03}$ & $39.25_{\pm 0.32}$ & $26.77_{\pm 0.16}$
& $0.43_{\pm 0.03}$ & $2.66_{\pm 0.30}$ & $0.41_{\pm 0.01}$ & $0.30_{\pm 0.00}$ \\
 & & GS
& $53.58_{\pm 1.00}$ & $75.56_{\pm 1.24}$ & $49.04_{\pm 1.07}$
& $13.17_{\pm 1.19}$ & $65.49_{\pm 1.36}$ & $14.31_{\pm 1.14}$
& $1.58_{\pm 0.06}$ & $3.37_{\pm 0.13}$ & $0.49_{\pm 0.00}$ & $0.33_{\pm 0.00}$ \\
 & & TimeWak
& $192.68_{\pm 1.64}$ & $10.87_{\pm 0.89}$ & $55.08_{\pm 1.08}$
& $0.33_{\pm 0.94}$ & $6.03_{\pm 1.07}$ & $-1.22_{\pm 1.01}$
& $0.09_{\pm 0.01}$ & $1.53_{\pm 0.19}$ & $0.14_{\pm 0.02}$ & $0.25_{\pm 0.00}$ \\
\cline{2-13}
 & SDformer & LVQMark
& $24.84_{\pm 0.84}$ & $3.99_{\pm 1.27}$ & $14.32_{\pm 0.90}$
& $0.29_{\pm 1.02}$ & $1.48_{\pm 0.99}$ & $0.32_{\pm 0.95}$
& $0.12_{\pm 0.00}$ & $1.27_{\pm 0.13}$ & $0.23_{\pm 0.01}$ & $0.25_{\pm 0.00}$ \\
\cline{2-13}
 & L-VQVAE & LVQMark
& $28.18_{\pm 1.09}$ & $21.94_{\pm 1.19}$ & $21.55_{\pm 1.00}$
& $-0.16_{\pm 0.97}$ & $-0.19_{\pm 1.21}$ & $-0.03_{\pm 0.99}$
& $0.03_{\pm 0.00}$ & $1.02_{\pm 0.10}$ & $0.09_{\pm 0.01}$ & $0.25_{\pm 0.00}$ \\
\hline
\hline
\multirow{5}{*}{{fMRI}} & \multirow{3}{*}{DiffusionTS} & TR
& $2.99_{\pm 0.06}$ & $2.47_{\pm 0.05}$ & $1.74_{\pm 0.07}$
& $0.42_{\pm 0.03}$ & $0.55_{\pm 0.04}$ & $4.89_{\pm 0.04}$
& $2.26_{\pm 0.10}$ & $13.42_{\pm 0.14}$ & $0.50_{\pm 0.00}$ & $0.15_{\pm 0.00}$ \\
 & & GS
& $418.07_{\pm 1.19}$ & $173.30_{\pm 0.84}$ & $262.45_{\pm 1.32}$
& $2.33_{\pm 0.88}$ & $-3.89_{\pm 0.87}$ & $-6.16_{\pm 1.01}$
& $0.71_{\pm 0.05}$ & $15.21_{\pm 0.06}$ & $0.50_{\pm 0.00}$ & $0.11_{\pm 0.00}$ \\
 & & TimeWak
& $380.87_{\pm 0.92}$ & $77.01_{\pm 0.92}$ & $134.32_{\pm 0.87}$
& $0.19_{\pm 1.07}$ & $2.39_{\pm 0.91}$ & $0.26_{\pm 1.04}$
& $0.18_{\pm 0.01}$ & $1.98_{\pm 0.07}$ & $0.08_{\pm 0.02}$ & $0.10_{\pm 0.00}$ \\
\cline{2-13}
 & SDformer & LVQMark
& $22.63_{\pm 0.95}$ & $6.01_{\pm 1.23}$ & $9.43_{\pm 0.97}$
& $0.14_{\pm 0.98}$ & $-0.41_{\pm 1.22}$ & $0.01_{\pm 0.91}$
& $0.94_{\pm 0.07}$ & $3.24_{\pm 0.15}$ & $0.21_{\pm 0.03}$ & $0.10_{\pm 0.00}$ \\
\cline{2-13}
 & L-VQVAE & LVQMark
& $20.28_{\pm 0.91}$ & $24.23_{\pm 1.32}$ & $17.16_{\pm 0.96}$
& $0.11_{\pm 0.91}$ & $-0.23_{\pm 1.32}$ & $-0.05_{\pm 1.02}$
& $0.20_{\pm 0.01}$ & $1.92_{\pm 0.03}$ & $0.23_{\pm 0.01}$ & $0.10_{\pm 0.00}$ \\
\hline
\hline
\end{tabular}%
}
\end{table*}

\begin{table*}[h!]
\centering
\setlength{\tabcolsep}{3pt}
\renewcommand{\arraystretch}{1.13}
\caption{Results of synthetic time series watermark detection and quality. Watermark and Non-watermarked detections ($z$-score) are evaluated under 30\% attacks. Detection and quality metrics (mean $\pm$ std) are for 64-length.}
\label{tab:main_64_std}
\resizebox{\textwidth}{!}{%
\begin{tabular}{c|l|l|r|r|r|r|r|r|r|r|r|r}
\hline
\hline
\rowcolor{gray!20}
\multicolumn{3}{c|}{\textbf{Setting}}
& \multicolumn{3}{c|}{\textbf{Watermark ($z$-score $\uparrow$)}}
& \multicolumn{3}{c|}{\textbf{Non-watermarked ($z$-score)}}
& \multicolumn{4}{c}{\textbf{Quality Metric ($\downarrow$)}} \\
\hline
\rowcolor{gray!20}
\textbf{Dataset} & \multicolumn{1}{c|}{\textbf{Model}} & \multicolumn{1}{c|}{\textbf{Method}}
& \multicolumn{1}{c|}{\textbf{Offset}} & \multicolumn{1}{c|}{\textbf{Crop}} & \multicolumn{1}{c|}{\textbf{Insert}}
& \multicolumn{1}{c|}{\textbf{Offset}} & \multicolumn{1}{c|}{\textbf{Crop}} & \multicolumn{1}{c|}{\textbf{Insert}}
& \multicolumn{1}{c|}{\textbf{C-FID}} & \multicolumn{1}{c|}{\textbf{Corr.}} & \multicolumn{1}{c|}{\textbf{Disc.}} & \multicolumn{1}{c}{\textbf{Pred.}} \\
\hline
\multirow{5}{*}{{Stocks}} & \multirow{3}{*}{DiffusionTS} & TR
& $0.88_{\pm 0.03}$ & $56.36_{\pm 0.84}$ & $4.60_{\pm 0.17}$
& $0.21_{\pm 0.03}$ & $69.43_{\pm 0.77}$ & $8.11_{\pm 0.21}$
& $1.52_{\pm 0.11}$ & $0.07_{\pm 0.02}$ & $0.15_{\pm 0.05}$ & $0.04_{\pm 0.00}$ \\
 & & GS
& $143.09_{\pm 1.43}$ & $-14.28_{\pm 0.94}$ & $64.00_{\pm 0.80}$
& $-1.40_{\pm 0.78}$ & $-14.87_{\pm 0.92}$ & $1.52_{\pm 1.01}$
& $1.50_{\pm 0.32}$ & $0.02_{\pm 0.01}$ & $0.22_{\pm 0.03}$ & $0.04_{\pm 0.00}$ \\
 & & TimeWak
& $335.49_{\pm 0.81}$ & $16.23_{\pm 0.92}$ & $79.44_{\pm 1.28}$
& $-0.19_{\pm 0.95}$ & $-4.32_{\pm 0.91}$ & $1.35_{\pm 0.91}$
& $0.29_{\pm 0.03}$ & $0.01_{\pm 0.00}$ & $0.13_{\pm 0.03}$ & $0.04_{\pm 0.00}$ \\
\cline{2-13}
 & SDformer & LVQMark
& $0.59_{\pm 0.81}$ & $0.99_{\pm 0.87}$ & $0.42_{\pm 0.93}$
& $-0.16_{\pm 0.89}$ & $1.19_{\pm 1.05}$ & $0.31_{\pm 1.04}$
& $0.08_{\pm 0.01}$ & $0.01_{\pm 0.00}$ & $0.01_{\pm 0.01}$ & $0.04_{\pm 0.00}$ \\
\cline{2-13}
 & L-VQVAE & LVQMark
& $15.21_{\pm 1.05}$ & $28.43_{\pm 1.33}$ & $8.23_{\pm 0.85}$
& $-0.12_{\pm 0.95}$ & $-0.05_{\pm 1.28}$ & $0.32_{\pm 1.06}$
& $0.07_{\pm 0.01}$ & $0.01_{\pm 0.01}$ & $0.06_{\pm 0.02}$ & $0.04_{\pm 0.00}$ \\
\hline
\hline
\multirow{5}{*}{{ETTh}} & \multirow{3}{*}{DiffusionTS} & TR
& $4.69_{\pm 0.09}$ & $17.54_{\pm 0.29}$ & $39.34_{\pm 0.52}$
& $0.89_{\pm 0.04}$ & $17.16_{\pm 0.27}$ & $38.55_{\pm 0.60}$
& $2.17_{\pm 0.10}$ & $0.22_{\pm 0.01}$ & $0.29_{\pm 0.01}$ & $0.14_{\pm 0.00}$ \\
 & & GS
& $160.44_{\pm 2.50}$ & $56.44_{\pm 1.67}$ & $127.53_{\pm 1.28}$
& $-1.53_{\pm 0.85}$ & $-5.44_{\pm 1.43}$ & $-5.82_{\pm 1.42}$
& $3.43_{\pm 0.24}$ & $0.25_{\pm 0.02}$ & $0.36_{\pm 0.01}$ & $0.16_{\pm 0.00}$ \\
 & & TimeWak
& $194.83_{\pm 1.53}$ & $3.42_{\pm 0.81}$ & $25.92_{\pm 1.13}$
& $1.17_{\pm 1.00}$ & $-0.28_{\pm 0.85}$ & $1.20_{\pm 0.88}$
& $0.37_{\pm 0.02}$ & $0.13_{\pm 0.01}$ & $0.11_{\pm 0.00}$ & $0.12_{\pm 0.00}$ \\
\cline{2-13}
 & SDformer & LVQMark
& $-0.16_{\pm 0.92}$ & $0.03_{\pm 1.16}$ & $0.07_{\pm 0.97}$
& $-0.94_{\pm 0.83}$ & $0.09_{\pm 1.17}$ & $-0.09_{\pm 0.98}$
& $0.04_{\pm 0.00}$ & $0.05_{\pm 0.02}$ & $0.00_{\pm 0.01}$ & $0.12_{\pm 0.01}$ \\
\cline{2-13}
 & L-VQVAE & LVQMark
& $16.81_{\pm 0.99}$ & $17.48_{\pm 1.13}$ & $14.46_{\pm 0.97}$
& $-0.12_{\pm 1.06}$ & $-0.22_{\pm 1.08}$ & $0.35_{\pm 0.84}$
& $0.03_{\pm 0.00}$ & $0.06_{\pm 0.03}$ & $0.01_{\pm 0.01}$ & $0.12_{\pm 0.01}$ \\
\hline
\hline
\multirow{5}{*}{{Energy}} & \multirow{3}{*}{DiffusionTS} & TR
& $2.14_{\pm 0.04}$ & $48.74_{\pm 0.34}$ & $58.22_{\pm 0.28}$
& $1.18_{\pm 0.04}$ & $53.73_{\pm 0.38}$ & $58.55_{\pm 0.29}$
& $0.58_{\pm 0.04}$ & $1.98_{\pm 0.09}$ & $0.43_{\pm 0.02}$ & $0.28_{\pm 0.00}$ \\
 & & GS
& $31.04_{\pm 0.88}$ & $2.78_{\pm 1.06}$ & $37.14_{\pm 1.38}$
& $14.51_{\pm 0.86}$ & $0.87_{\pm 1.08}$ & $11.77_{\pm 1.18}$
& $1.78_{\pm 0.14}$ & $2.72_{\pm 0.15}$ & $0.48_{\pm 0.01}$ & $0.31_{\pm 0.00}$ \\
 & & TimeWak
& $167.49_{\pm 2.09}$ & $5.02_{\pm 0.93}$ & $28.04_{\pm 1.01}$
& $-9.69_{\pm 0.82}$ & $-0.42_{\pm 0.93}$ & $-1.57_{\pm 0.98}$
& $0.14_{\pm 0.01}$ & $1.52_{\pm 0.24}$ & $0.14_{\pm 0.01}$ & $0.25_{\pm 0.00}$ \\
\cline{2-13}
 & SDformer & LVQMark
& $7.41_{\pm 0.98}$ & $1.14_{\pm 0.87}$ & $3.29_{\pm 0.88}$
& $-0.02_{\pm 1.10}$ & $0.34_{\pm 0.87}$ & $0.34_{\pm 0.90}$
& $0.04_{\pm 0.00}$ & $1.04_{\pm 0.15}$ & $0.08_{\pm 0.02}$ & $0.25_{\pm 0.00}$ \\
\cline{2-13}
 & L-VQVAE & LVQMark
& $15.24_{\pm 1.16}$ & $13.22_{\pm 1.12}$ & $11.31_{\pm 1.02}$
& $-0.06_{\pm 1.09}$ & $-0.25_{\pm 1.14}$ & $0.04_{\pm 1.07}$
& $0.04_{\pm 0.00}$ & $0.94_{\pm 0.34}$ & $0.15_{\pm 0.01}$ & $0.25_{\pm 0.00}$ \\
\hline
\hline
\multirow{5}{*}{{fMRI}} & \multirow{3}{*}{DiffusionTS} & TR
& $5.25_{\pm 0.05}$ & $3.36_{\pm 0.06}$ & $20.02_{\pm 0.11}$
& $0.05_{\pm 0.03}$ & $2.27_{\pm 0.04}$ & $12.14_{\pm 0.05}$
& $3.63_{\pm 0.41}$ & $12.83_{\pm 0.08}$ & $0.40_{\pm 0.17}$ & $0.14_{\pm 0.00}$ \\
 & & GS
& $543.72_{\pm 1.11}$ & $275.99_{\pm 1.72}$ & $421.92_{\pm 3.12}$
& $-1.78_{\pm 0.91}$ & $-38.77_{\pm 1.70}$ & $-0.01_{\pm 2.12}$
& $0.74_{\pm 0.05}$ & $8.31_{\pm 0.03}$ & $0.50_{\pm 0.00}$ & $0.10_{\pm 0.00}$ \\
 & & TimeWak
& $574.43_{\pm 0.98}$ & $84.45_{\pm 0.98}$ & $185.86_{\pm 0.95}$
& $0.08_{\pm 0.97}$ & $-2.74_{\pm 0.99}$ & $0.18_{\pm 0.90}$
& $0.45_{\pm 0.01}$ & $1.87_{\pm 0.05}$ & $0.25_{\pm 0.11}$ & $0.10_{\pm 0.00}$ \\
\cline{2-13}
 & SDformer & LVQMark
& $2.39_{\pm 0.97}$ & $1.13_{\pm 1.09}$ & $1.48_{\pm 0.95}$
& $0.03_{\pm 0.86}$ & $0.51_{\pm 1.16}$ & $-0.40_{\pm 0.85}$
& $0.13_{\pm 0.01}$ & $1.19_{\pm 0.04}$ & $0.12_{\pm 0.02}$ & $0.09_{\pm 0.00}$ \\
\cline{2-13}
 & L-VQVAE & LVQMark
& $12.90_{\pm 0.88}$ & $15.19_{\pm 1.23}$ & $10.45_{\pm 0.97}$
& $-0.03_{\pm 0.99}$ & $-0.24_{\pm 1.15}$ & $-0.17_{\pm 0.99}$
& $0.13_{\pm 0.00}$ & $1.14_{\pm 0.06}$ & $0.27_{\pm 0.11}$ & $0.09_{\pm 0.00}$ \\
\hline
\hline
\end{tabular}%
}
\end{table*}

\begin{table*}[h!]
\centering
\setlength{\tabcolsep}{3pt}
\renewcommand{\arraystretch}{1.13}
\caption{Results of synthetic time series watermark detection and quality. Watermark and Non-watermarked detections ($z$-score) are evaluated under 30\% attacks. Detection and quality metrics (mean $\pm$ std) are for 128-length.}
\label{tab:main_128_std}
\resizebox{\textwidth}{!}{%
\begin{tabular}{c|l|l|r|r|r|r|r|r|r|r|r|r}
\hline
\hline
\rowcolor{gray!20}
\multicolumn{3}{c|}{\textbf{Setting}}
& \multicolumn{3}{c|}{\textbf{Watermark ($z$-score $\uparrow$)}}
& \multicolumn{3}{c|}{\textbf{Non-watermarked ($z$-score)}}
& \multicolumn{4}{c}{\textbf{Quality Metric ($\downarrow$)}} \\
\hline
\rowcolor{gray!20}
\textbf{Dataset} & \textbf{Model} & \textbf{Method}
& \textbf{Offset} & \textbf{Crop} & \textbf{Insert}
& \textbf{Offset} & \textbf{Crop} & \textbf{Insert}
& \textbf{C-FID} & \textbf{Corr.} & \textbf{Disc.} & \textbf{Pred.} \\
\hline
\multirow{5}{*}{{Stocks}} & \multirow{3}{*}{DiffusionTS} & TR
& $0.96_{\pm 0.03}$ & $60.85_{\pm 0.68}$ & $19.13_{\pm 0.39}$
& $0.30_{\pm 0.03}$ & $84.66_{\pm 0.65}$ & $20.60_{\pm 0.43}$
& $3.05_{\pm 0.78}$ & $0.09_{\pm 0.01}$ & $0.24_{\pm 0.05}$ & $0.04_{\pm 0.00}$ \\
 & & GS
& $158.85_{\pm 1.94}$ & $1.08_{\pm 0.95}$ & $56.98_{\pm 0.95}$
& $-4.40_{\pm 1.00}$ & $-10.04_{\pm 0.81}$ & $-5.22_{\pm 1.01}$
& $2.63_{\pm 0.16}$ & $0.04_{\pm 0.02}$ & $0.19_{\pm 0.03}$ & $0.04_{\pm 0.00}$ \\
 & & TimeWak
& $466.00_{\pm 1.19}$ & $10.37_{\pm 1.01}$ & $76.52_{\pm 1.29}$
& $-0.16_{\pm 0.90}$ & $-4.18_{\pm 0.93}$ & $3.74_{\pm 1.03}$
& $0.34_{\pm 0.08}$ & $0.01_{\pm 0.00}$ & $0.15_{\pm 0.04}$ & $0.04_{\pm 0.00}$ \\
\cline{2-13}
 & SDformer & LVQMark
& $-0.57_{\pm 0.94}$ & $-0.77_{\pm 0.66}$ & $-0.78_{\pm 0.95}$
& $-0.54_{\pm 0.87}$ & $-0.89_{\pm 0.98}$ & $-1.20_{\pm 1.10}$
& $0.12_{\pm 0.01}$ & $0.01_{\pm 0.01}$ & $0.03_{\pm 0.03}$ & $0.04_{\pm 0.00}$ \\
\cline{2-13}
 & L-VQVAE & LVQMark
& $19.79_{\pm 1.51}$ & $13.17_{\pm 1.64}$ & $13.82_{\pm 1.20}$
& $0.09_{\pm 0.92}$ & $-0.09_{\pm 1.24}$ & $0.23_{\pm 0.96}$
& $0.08_{\pm 0.01}$ & $0.01_{\pm 0.01}$ & $0.18_{\pm 0.12}$ & $0.04_{\pm 0.00}$ \\
\hline
\hline
\multirow{5}{*}{{ETTh}} & \multirow{3}{*}{DiffusionTS} & TR
& $0.61_{\pm 0.04}$ & $10.99_{\pm 0.23}$ & $35.91_{\pm 0.46}$
& $1.52_{\pm 0.04}$ & $12.19_{\pm 0.23}$ & $36.77_{\pm 0.38}$
& $2.52_{\pm 0.12}$ & $0.26_{\pm 0.01}$ & $0.30_{\pm 0.01}$ & $0.13_{\pm 0.00}$ \\
 & & GS
& $229.36_{\pm 2.44}$ & $51.50_{\pm 1.15}$ & $155.83_{\pm 1.52}$
& $-6.09_{\pm 0.98}$ & $-13.71_{\pm 0.85}$ & $-6.37_{\pm 1.18}$
& $5.59_{\pm 0.36}$ & $0.23_{\pm 0.01}$ & $0.39_{\pm 0.00}$ & $0.15_{\pm 0.00}$ \\
 & & TimeWak
& $257.53_{\pm 2.12}$ & $4.62_{\pm 1.03}$ & $40.33_{\pm 0.98}$
& $1.09_{\pm 0.91}$ & $-1.63_{\pm 0.86}$ & $0.77_{\pm 0.97}$
& $1.08_{\pm 0.08}$ & $0.17_{\pm 0.02}$ & $0.15_{\pm 0.01}$ & $0.11_{\pm 0.00}$ \\
\cline{2-13}
 & SDformer & LVQMark
& $2.01_{\pm 1.01}$ & $1.12_{\pm 1.15}$ & $1.43_{\pm 0.85}$
& $0.53_{\pm 0.85}$ & $0.76_{\pm 1.11}$ & $1.01_{\pm 0.87}$
& $0.04_{\pm 0.00}$ & $0.05_{\pm 0.03}$ & $0.03_{\pm 0.01}$ & $0.11_{\pm 0.01}$ \\
\cline{2-13}
 & L-VQVAE & LVQMark
& $13.82_{\pm 0.93}$ & $13.76_{\pm 1.07}$ & $11.98_{\pm 1.02}$
& $-0.04_{\pm 0.81}$ & $-0.05_{\pm 1.09}$ & $-0.30_{\pm 0.83}$
& $0.04_{\pm 0.00}$ & $0.05_{\pm 0.01}$ & $0.02_{\pm 0.01}$ & $0.11_{\pm 0.01}$ \\
\hline
\hline
\multirow{5}{*}{{Energy}} & \multirow{3}{*}{DiffusionTS} & TR
& $2.34_{\pm 0.04}$ & $56.95_{\pm 0.34}$ & $95.70_{\pm 0.35}$
& $2.20_{\pm 0.03}$ & $61.47_{\pm 0.39}$ & $98.65_{\pm 0.37}$
& $0.50_{\pm 0.05}$ & $1.65_{\pm 0.19}$ & $0.49_{\pm 0.00}$ & $0.28_{\pm 0.00}$ \\
 & & GS
& $56.04_{\pm 0.78}$ & $61.55_{\pm 1.05}$ & $65.92_{\pm 1.13}$
& $15.43_{\pm 1.09}$ & $48.30_{\pm 0.98}$ & $27.89_{\pm 1.32}$
& $3.34_{\pm 0.25}$ & $3.59_{\pm 0.12}$ & $0.48_{\pm 0.01}$ & $0.29_{\pm 0.01}$ \\
 & & TimeWak
& $172.41_{\pm 1.57}$ & $1.82_{\pm 0.90}$ & $18.91_{\pm 0.96}$
& $-1.57_{\pm 0.89}$ & $-0.78_{\pm 0.88}$ & $0.66_{\pm 0.86}$
& $0.17_{\pm 0.01}$ & $1.50_{\pm 0.25}$ & $0.24_{\pm 0.10}$ & $0.25_{\pm 0.00}$ \\
\cline{2-13}
 & SDformer & LVQMark
& $7.48_{\pm 1.03}$ & $-0.29_{\pm 0.86}$ & $2.74_{\pm 0.89}$
& $0.06_{\pm 0.96}$ & $-0.65_{\pm 0.69}$ & $-0.08_{\pm 0.90}$
& $0.04_{\pm 0.00}$ & $0.67_{\pm 0.19}$ & $0.05_{\pm 0.01}$ & $0.25_{\pm 0.00}$ \\
\cline{2-13}
 & L-VQVAE & LVQMark
& $8.48_{\pm 1.01}$ & $6.16_{\pm 1.26}$ & $6.57_{\pm 1.08}$
& $-0.03_{\pm 1.02}$ & $0.00_{\pm 1.26}$ & $0.13_{\pm 0.82}$
& $0.03_{\pm 0.00}$ & $0.74_{\pm 0.22}$ & $0.19_{\pm 0.02}$ & $0.24_{\pm 0.00}$ \\
\hline
\hline
\multirow{5}{*}{{fMRI}} & \multirow{3}{*}{DiffusionTS} & TR
& $0.22_{\pm 0.04}$ & $4.73_{\pm 0.05}$ & $25.33_{\pm 0.12}$
& $0.02_{\pm 0.02}$ & $6.33_{\pm 0.04}$ & $21.25_{\pm 0.06}$
& $4.57_{\pm 0.51}$ & $14.15_{\pm 0.05}$ & $0.45_{\pm 0.09}$ & $0.17_{\pm 0.00}$ \\
 & & GS
& $801.62_{\pm 0.81}$ & $148.15_{\pm 1.57}$ & $499.87_{\pm 2.71}$
& $-0.70_{\pm 0.97}$ & $-57.42_{\pm 0.88}$ & $0.52_{\pm 1.78}$
& $1.05_{\pm 0.06}$ & $5.97_{\pm 0.05}$ & $0.50_{\pm 0.00}$ & $0.11_{\pm 0.00}$ \\
 & & TimeWak
& $739.70_{\pm 1.24}$ & $97.66_{\pm 0.98}$ & $204.66_{\pm 0.96}$
& $-0.07_{\pm 0.96}$ & $-3.03_{\pm 0.95}$ & $-1.95_{\pm 1.00}$
& $0.81_{\pm 0.04}$ & $1.81_{\pm 0.03}$ & $0.36_{\pm 0.10}$ & $0.10_{\pm 0.00}$ \\
\cline{2-13}
 & SDformer & LVQMark
& $1.95_{\pm 0.94}$ & $0.73_{\pm 1.15}$ & $0.93_{\pm 0.90}$
& $-0.03_{\pm 0.88}$ & $-0.25_{\pm 1.11}$ & $-0.15_{\pm 0.86}$
& $0.09_{\pm 0.01}$ & $0.84_{\pm 0.03}$ & $0.16_{\pm 0.03}$ & $0.08_{\pm 0.00}$ \\
\cline{2-13}
 & L-VQVAE & LVQMark
& $10.47_{\pm 0.97}$ & $10.75_{\pm 1.25}$ & $8.05_{\pm 0.96}$
& $-0.09_{\pm 0.87}$ & $0.16_{\pm 1.24}$ & $-0.01_{\pm 0.96}$
& $0.13_{\pm 0.01}$ & $0.90_{\pm 0.02}$ & $0.06_{\pm 0.05}$ & $0.08_{\pm 0.00}$ \\
\hline
\hline
\end{tabular}%
}
\end{table*}

% visualization generation result
\section{Visualization of Generation Results}
\label{app:qualitative}

\begin{figure}[h!]
    \centering
    \begin{subfigure}{0.32\textwidth}
        \centering
        \includegraphics[width=\linewidth]{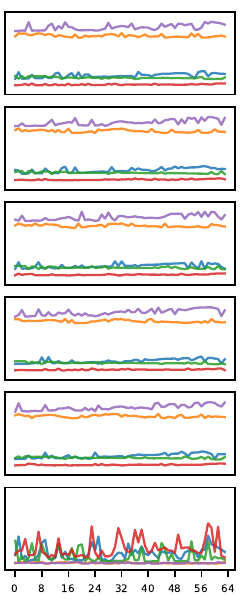}
        \caption{Original}
    \end{subfigure}
    \hspace{0.001\textwidth}
    \begin{subfigure}{0.32\textwidth}
        \centering
        \includegraphics[width=\linewidth]{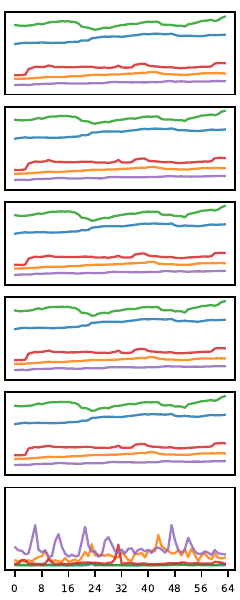}
        \caption{Watermarked}
    \end{subfigure}
    \hspace{0.001\textwidth}
    \begin{subfigure}{0.32\textwidth}
        \centering
        \includegraphics[width=\linewidth]{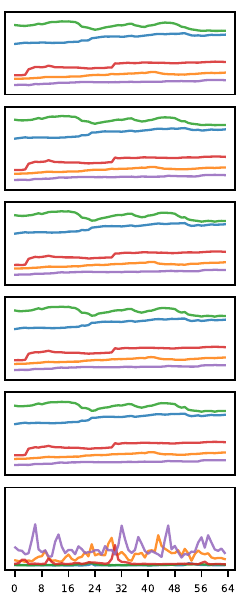}
        \caption{Non-watermarked}
    \end{subfigure}
    \caption{Visualization of original and generated samples on the Stocks dataset with length 64.}
    \label{fig:vis_stocks}
\end{figure}

\begin{figure}[h!]
    \centering
    \begin{subfigure}{0.305\textwidth}
        \centering
        \includegraphics[width=\linewidth]{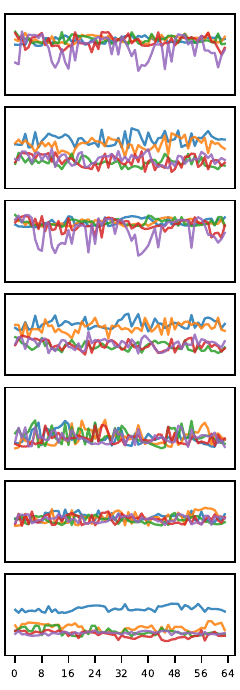}
        \caption{Original}
    \end{subfigure}
    \hspace{0.005\textwidth}
    \begin{subfigure}{0.305\textwidth}
        \centering
        \includegraphics[width=\linewidth]{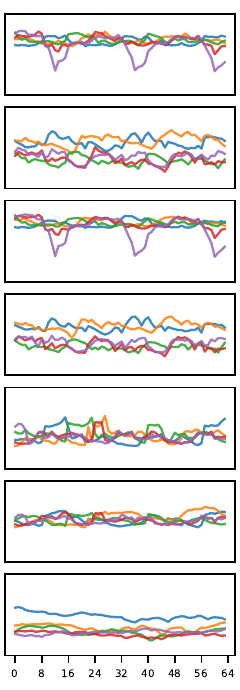}
        \caption{Watermarked}
    \end{subfigure}
    \hspace{0.005\textwidth}
    \begin{subfigure}{0.305\textwidth}
        \centering
        \includegraphics[width=\linewidth]{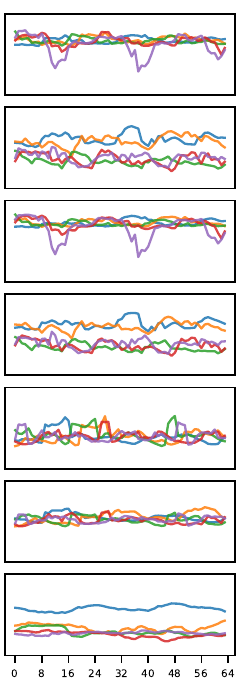}
        \caption{Non-watermarked}
    \end{subfigure}
    \caption{Visualization of original and generated samples on the ETTh dataset with length 64.}
    \label{fig:vis_etth}
\end{figure}

\begin{figure}[h!]
    \centering
    \begin{subfigure}{0.305\textwidth}
        \centering
        \includegraphics[width=\linewidth]{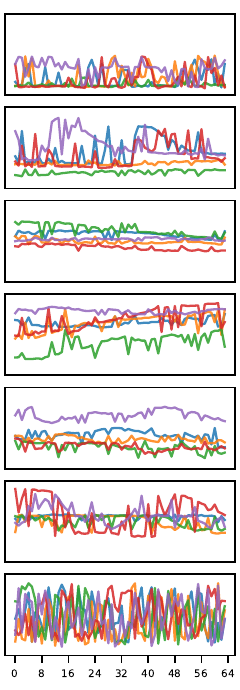}
        \caption{Original}
    \end{subfigure}
    \hspace{0.005\textwidth}
    \begin{subfigure}{0.305\textwidth}
        \centering
        \includegraphics[width=\linewidth]{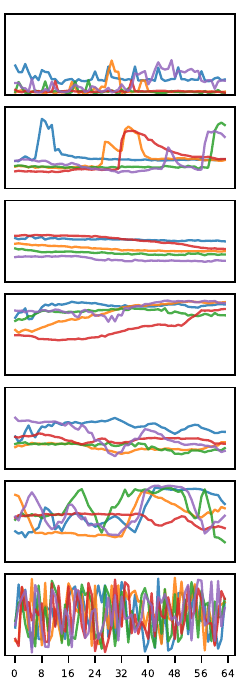}
        \caption{Watermarked}
    \end{subfigure}
    \hspace{0.005\textwidth}
    \begin{subfigure}{0.305\textwidth}
        \centering
        \includegraphics[width=\linewidth]{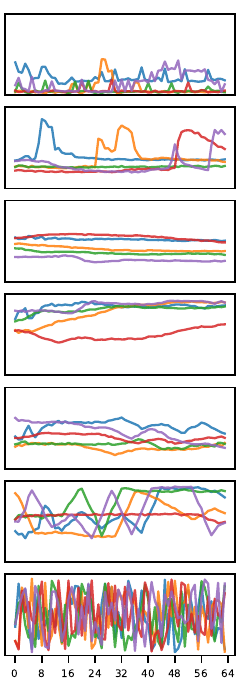}
        \caption{Non-watermarked}
    \end{subfigure}
    \caption{Visualization of original and generated samples on the Energy dataset with length 64.}
    \label{fig:vis_energy}
\end{figure}

\begin{figure}[h!]
    \centering
    \begin{subfigure}{0.305\textwidth}
        \centering
        \includegraphics[width=\linewidth]{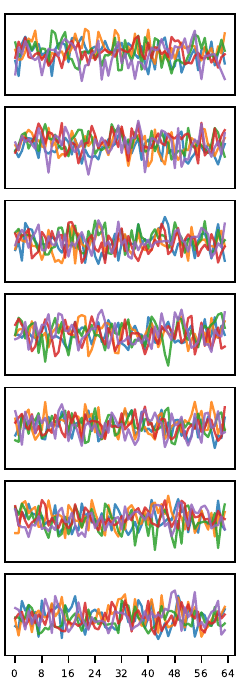}
        \caption{Original}
    \end{subfigure}
    \hspace{0.005\textwidth}
    \begin{subfigure}{0.305\textwidth}
        \centering
        \includegraphics[width=\linewidth]{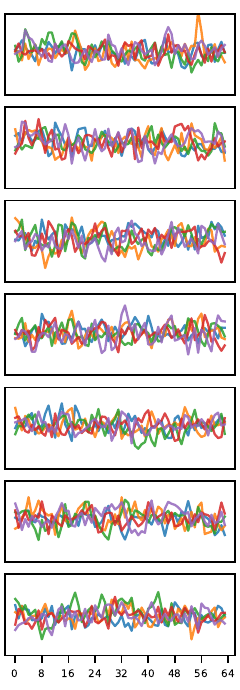}
        \caption{Watermarked}
    \end{subfigure}
    \hspace{0.005\textwidth}
    \begin{subfigure}{0.305\textwidth}
        \centering
        \includegraphics[width=\linewidth]{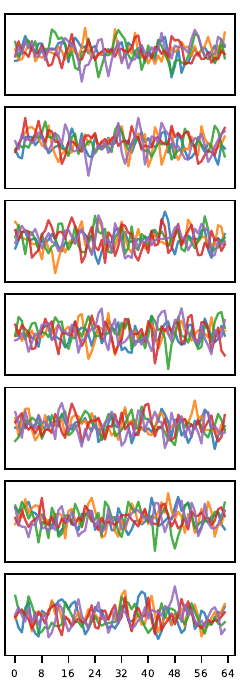}
        \caption{Non-watermarked}
    \end{subfigure}
    \caption{Visualization of original and generated samples on the fMRI dataset with length 64.}
    \label{fig:vis_fMRI}
\end{figure}

We provide qualitative visualizations of generation-time watermarked time series data on Stocks, ETTh, Energy, and fMRI for 64-length sequences.
Figures~\ref{fig:vis_stocks}-\ref{fig:vis_fMRI} compare original samples with LVQMark generations.
The left column shows original samples, while the middle and right columns show watermarked generations produced with LVQMark and non-watermarked generations produced without watermarking, respectively.

Each row corresponds to a different channel, with five representative samples shown in each column. The generated samples capture dataset-specific temporal patterns, and the watermarked samples remain qualitatively similar to the non-watermarked ones, suggesting that LVQMark does not introduce noticeable qualitative degradation. This observation is also consistent with the low Context-FID values reported in Table~\ref{tab:main_64}.

% \section{Technical appendices and supplementary material}
% Technical appendices with additional results, figures, graphs, and proofs may be submitted with the paper submission before the full submission deadline (see above). You can upload a ZIP file for videos or code, but do not upload a separate PDF file for the appendix. There is no page limit for the technical appendices. 

% Note: Think of the appendix as ``optional reading'' for reviewers. The paper must be able to stand alone without the appendix; for example, adding critical experiments that support the main claims to an appendix is inappropriate. 

%%%%%%%%%%%%%%%%%%%%%%%%%%%%%%%%%%%%%%%%%%%%%%%%%%%%%%%%%%%%

\clearpage
\input{checklist.tex}

\end{document}

%% file: checklist.tex
\section*{NeurIPS Paper Checklist}

\begin{enumerate}

\item {\bf Claims}
    \item[] Question: Do the main claims made in the abstract and introduction accurately reflect the paper's contributions and scope?
    \item[] Answer: \answerYes{} % Replace by \answerYes{}, \answerNo{}, or \answerNA{}.
    \item[] Justification: The claims in the abstract and introduction accurately reflect the paper's contributions. The proposed L-VQVAE and LVQMark are described with their scope and assumptions, and the experimental results in Sections 3–4 support the stated claims.
    \item[] Guidelines:
    \begin{itemize}
        \item The answer \answerNA{} means that the abstract and introduction do not include the claims made in the paper.
        \item The abstract and/or introduction should clearly state the claims made, including the contributions made in the paper and important assumptions and limitations. A \answerNo{} or \answerNA{} answer to this question will not be perceived well by the reviewers. 
        \item The claims made should match theoretical and experimental results, and reflect how much the results can be expected to generalize to other settings. 
        \item It is fine to include aspirational goals as motivation as long as it is clear that these goals are not attained by the paper. 
    \end{itemize}

\item {\bf Limitations}
    \item[] Question: Does the paper discuss the limitations of the work performed by the authors?
    \item[] Answer: \answerYes{} % Replace by \answerYes{}, \answerNo{}, or \answerNA{}.
    \item[] Justification: Section 5 contains an explicit Limitations paragraph discussing the restricted number of datasets, sequence lengths, attack types, and the increased complexity of the multi-stage training pipeline compared to single-stage baselines.
    \item[] Guidelines:
    \begin{itemize}
        \item The answer \answerNA{} means that the paper has no limitation while the answer \answerNo{} means that the paper has limitations, but those are not discussed in the paper. 
        \item The authors are encouraged to create a separate ``Limitations'' section in their paper.
        \item The paper should point out any strong assumptions and how robust the results are to violations of these assumptions (e.g., independence assumptions, noiseless settings, model well-specification, asymptotic approximations only holding locally). The authors should reflect on how these assumptions might be violated in practice and what the implications would be.
        \item The authors should reflect on the scope of the claims made, e.g., if the approach was only tested on a few datasets or with a few runs. In general, empirical results often depend on implicit assumptions, which should be articulated.
        \item The authors should reflect on the factors that influence the performance of the approach. For example, a facial recognition algorithm may perform poorly when image resolution is low or images are taken in low lighting. Or a speech-to-text system might not be used reliably to provide closed captions for online lectures because it fails to handle technical jargon.
        \item The authors should discuss the computational efficiency of the proposed algorithms and how they scale with dataset size.
        \item If applicable, the authors should discuss possible limitations of their approach to address problems of privacy and fairness.
        \item While the authors might fear that complete honesty about limitations might be used by reviewers as grounds for rejection, a worse outcome might be that reviewers discover limitations that aren't acknowledged in the paper. The authors should use their best judgment and recognize that individual actions in favor of transparency play an important role in developing norms that preserve the integrity of the community. Reviewers will be specifically instructed to not penalize honesty concerning limitations.
    \end{itemize}

\item {\bf Theory assumptions and proofs}
    \item[] Question: For each theoretical result, does the paper provide the full set of assumptions and a complete (and correct) proof?
    \item[] Answer: \answerNA{} % Replace by \answerYes{}, \answerNo{}, or \answerNA{}.
    \item[] Justification: The paper does not include formal theorems or proofs; it is an empirical contribution.
    \item[] Guidelines:
    \begin{itemize}
        \item The answer \answerNA{} means that the paper does not include theoretical results. 
        \item All the theorems, formulas, and proofs in the paper should be numbered and cross-referenced.
        \item All assumptions should be clearly stated or referenced in the statement of any theorems.
        \item The proofs can either appear in the main paper or the supplemental material, but if they appear in the supplemental material, the authors are encouraged to provide a short proof sketch to provide intuition. 
        \item Inversely, any informal proof provided in the core of the paper should be complemented by formal proofs provided in appendix or supplemental material.
        \item Theorems and Lemmas that the proof relies upon should be properly referenced. 
    \end{itemize}

    \item {\bf Experimental result reproducibility}
    \item[] Question: Does the paper fully disclose all the information needed to reproduce the main experimental results of the paper to the extent that it affects the main claims and/or conclusions of the paper (regardless of whether the code and data are provided or not)?
    \item[] Answer: \answerYes{} % Replace by \answerYes{}, \answerNo{}, or \answerNA{}.
    \item[] Justification: Appendix B provides full details on datasets, preprocessing, hyperparameters (Tables 9--10), attack configurations, evaluation metrics, and cost measurement (Appendix B.6). All results are averaged over five runs with fixed random seeds.
    \item[] Guidelines:
    \begin{itemize}
        \item The answer \answerNA{} means that the paper does not include experiments.
        \item If the paper includes experiments, a \answerNo{} answer to this question will not be perceived well by the reviewers: Making the paper reproducible is important, regardless of whether the code and data are provided or not.
        \item If the contribution is a dataset and\slash or model, the authors should describe the steps taken to make their results reproducible or verifiable. 
        \item Depending on the contribution, reproducibility can be accomplished in various ways. For example, if the contribution is a novel architecture, describing the architecture fully might suffice, or if the contribution is a specific model and empirical evaluation, it may be necessary to either make it possible for others to replicate the model with the same dataset, or provide access to the model. In general. releasing code and data is often one good way to accomplish this, but reproducibility can also be provided via detailed instructions for how to replicate the results, access to a hosted model (e.g., in the case of a large language model), releasing of a model checkpoint, or other means that are appropriate to the research performed.
        \item While NeurIPS does not require releasing code, the conference does require all submissions to provide some reasonable avenue for reproducibility, which may depend on the nature of the contribution. For example
        \begin{enumerate}
            \item If the contribution is primarily a new algorithm, the paper should make it clear how to reproduce that algorithm.
            \item If the contribution is primarily a new model architecture, the paper should describe the architecture clearly and fully.
            \item If the contribution is a new model (e.g., a large language model), then there should either be a way to access this model for reproducing the results or a way to reproduce the model (e.g., with an open-source dataset or instructions for how to construct the dataset).
            \item We recognize that reproducibility may be tricky in some cases, in which case authors are welcome to describe the particular way they provide for reproducibility. In the case of closed-source models, it may be that access to the model is limited in some way (e.g., to registered users), but it should be possible for other researchers to have some path to reproducing or verifying the results.
        \end{enumerate}
    \end{itemize}

\item {\bf Open access to data and code}
    \item[] Question: Does the paper provide open access to the data and code, with sufficient instructions to faithfully reproduce the main experimental results, as described in supplemental material?
    \item[] Answer: \answerYes{} % Replace by \answerYes{}, \answerNo{}, or \answerNA{}.
    \item[] Justification: The code is publicly available at \url{https://github.com/dongbeank/LVQMark}. All datasets used are publicly available.
    \item[] Guidelines:
    \begin{itemize}
        \item The answer \answerNA{} means that paper does not include experiments requiring code.
        \item Please see the NeurIPS code and data submission guidelines (\url{https://neurips.cc/public/guides/CodeSubmissionPolicy}) for more details.
        \item While we encourage the release of code and data, we understand that this might not be possible, so \answerNo{} is an acceptable answer. Papers cannot be rejected simply for not including code, unless this is central to the contribution (e.g., for a new open-source benchmark).
        \item The instructions should contain the exact command and environment needed to run to reproduce the results. See the NeurIPS code and data submission guidelines (\url{https://neurips.cc/public/guides/CodeSubmissionPolicy}) for more details.
        \item The authors should provide instructions on data access and preparation, including how to access the raw data, preprocessed data, intermediate data, and generated data, etc.
        \item The authors should provide scripts to reproduce all experimental results for the new proposed method and baselines. If only a subset of experiments are reproducible, they should state which ones are omitted from the script and why.
        \item At submission time, to preserve anonymity, the authors should release anonymized versions (if applicable).
        \item Providing as much information as possible in supplemental material (appended to the paper) is recommended, but including URLs to data and code is permitted.
    \end{itemize}

\item {\bf Experimental setting/details}
    \item[] Question: Does the paper specify all the training and test details (e.g., data splits, hyperparameters, how they were chosen, type of optimizer) necessary to understand the results?
    \item[] Answer: \answerYes{} % Replace by \answerYes{}, \answerNo{}, or \answerNA{}.
    \item[] Justification: Appendix B specifies data splits (80\%/20\%), optimizer, batch sizes, learning rates, training steps per stage, and attack parameters in full detail.
    \item[] Guidelines:
    \begin{itemize}
        \item The answer \answerNA{} means that the paper does not include experiments.
        \item The experimental setting should be presented in the core of the paper to a level of detail that is necessary to appreciate the results and make sense of them.
        \item The full details can be provided either with the code, in appendix, or as supplemental material.
    \end{itemize}

\item {\bf Experiment statistical significance}
    \item[] Question: Does the paper report error bars suitably and correctly defined or other appropriate information about the statistical significance of the experiments?
    \item[] Answer: \answerYes{} % Replace by \answerYes{}, \answerNo{}, or \answerNA{}.
    \item[] Justification: All reported results are averaged over five runs with fixed random seeds. Mean and standard deviation of all detection and quality metrics are reported in Appendix G (Tables across sequence lengths 24, 64, and 128).

    \item[] Guidelines:
    \begin{itemize}
        \item The answer \answerNA{} means that the paper does not include experiments.
        \item The authors should answer \answerYes{} if the results are accompanied by error bars, confidence intervals, or statistical significance tests, at least for the experiments that support the main claims of the paper.
        \item The factors of variability that the error bars are capturing should be clearly stated (for example, train/test split, initialization, random drawing of some parameter, or overall run with given experimental conditions).
        \item The method for calculating the error bars should be explained (closed form formula, call to a library function, bootstrap, etc.)
        \item The assumptions made should be given (e.g., Normally distributed errors).
        \item It should be clear whether the error bar is the standard deviation or the standard error of the mean.
        \item It is OK to report 1-sigma error bars, but one should state it. The authors should preferably report a 2-sigma error bar than state that they have a 96\% CI, if the hypothesis of Normality of errors is not verified.
        \item For asymmetric distributions, the authors should be careful not to show in tables or figures symmetric error bars that would yield results that are out of range (e.g., negative error rates).
        \item If error bars are reported in tables or plots, the authors should explain in the text how they were calculated and reference the corresponding figures or tables in the text.
    \end{itemize}

\item {\bf Experiments compute resources}
    \item[] Question: For each experiment, does the paper provide sufficient information on the computer resources (type of compute workers, memory, time of execution) needed to reproduce the experiments?
    \item[] Answer: \answerYes{} % Replace by \answerYes{}, \answerNo{}, or \answerNA{}.
    \item[] Justification: Compute resources are described in Appendix B.5, including GPU models (NVIDIA H100 PCIe 80 GB, RTX PRO 6000 Blackwell 96 GB $\times$2) and CPU. Per-stage training steps are provided in Tables 9--10, and the training time, generation throughput, and verification cost of LVQMark and TimeWak are measured in Appendix B.6 on a single RTX PRO 6000.
    \item[] Guidelines:
    \begin{itemize}
        \item The answer \answerNA{} means that the paper does not include experiments.
        \item The paper should indicate the type of compute workers CPU or GPU, internal cluster, or cloud provider, including relevant memory and storage.
        \item The paper should provide the amount of compute required for each of the individual experimental runs as well as estimate the total compute. 
        \item The paper should disclose whether the full research project required more compute than the experiments reported in the paper (e.g., preliminary or failed experiments that didn't make it into the paper). 
    \end{itemize}
    
\item {\bf Code of ethics}
    \item[] Question: Does the research conducted in the paper conform, in every respect, with the NeurIPS Code of Ethics \url{https://neurips.cc/public/EthicsGuidelines}?
    \item[] Answer: \answerYes{} % Replace by \answerYes{}, \answerNo{}, or \answerNA{}.
    \item[] Justification: The research uses only publicly available datasets, involves no human subjects or sensitive data collection, and conforms to the NeurIPS Code of Ethics.
    \item[] Guidelines:
    \begin{itemize}
        \item The answer \answerNA{} means that the authors have not reviewed the NeurIPS Code of Ethics.
        \item If the authors answer \answerNo, they should explain the special circumstances that require a deviation from the Code of Ethics.
        \item The authors should make sure to preserve anonymity (e.g., if there is a special consideration due to laws or regulations in their jurisdiction).
    \end{itemize}

\item {\bf Broader impacts}
    \item[] Question: Does the paper discuss both potential positive societal impacts and negative societal impacts of the work performed?
    \item[] Answer: \answerYes{} % Replace by \answerYes{}, \answerNo{}, or \answerNA{}.
    \item[] Justification: The proposed method supports provenance verification of synthetic time series, which has a positive societal impact in combating misattribution of AI-generated content. A potential negative impact is that advances in watermarking robustness may motivate stronger evasion attacks; this is acknowledged in Section 4.5.
    \item[] Guidelines:
    \begin{itemize}
        \item The answer \answerNA{} means that there is no societal impact of the work performed.
        \item If the authors answer \answerNA{} or \answerNo, they should explain why their work has no societal impact or why the paper does not address societal impact.
        \item Examples of negative societal impacts include potential malicious or unintended uses (e.g., disinformation, generating fake profiles, surveillance), fairness considerations (e.g., deployment of technologies that could make decisions that unfairly impact specific groups), privacy considerations, and security considerations.
        \item The conference expects that many papers will be foundational research and not tied to particular applications, let alone deployments. However, if there is a direct path to any negative applications, the authors should point it out. For example, it is legitimate to point out that an improvement in the quality of generative models could be used to generate Deepfakes for disinformation. On the other hand, it is not needed to point out that a generic algorithm for optimizing neural networks could enable people to train models that generate Deepfakes faster.
        \item The authors should consider possible harms that could arise when the technology is being used as intended and functioning correctly, harms that could arise when the technology is being used as intended but gives incorrect results, and harms following from (intentional or unintentional) misuse of the technology.
        \item If there are negative societal impacts, the authors could also discuss possible mitigation strategies (e.g., gated release of models, providing defenses in addition to attacks, mechanisms for monitoring misuse, mechanisms to monitor how a system learns from feedback over time, improving the efficiency and accessibility of ML).
    \end{itemize}
    
\item {\bf Safeguards}
    \item[] Question: Does the paper describe safeguards that have been put in place for responsible release of data or models that have a high risk for misuse (e.g., pre-trained language models, image generators, or scraped datasets)?
    \item[] Answer: \answerYes{} % Replace by \answerYes{}, \answerNo{}, or \answerNA{}.
    \item[] Justification: The paper presents a detection framework and does not release high-risk pretrained generative models or scraped datasets that would require additional safeguards.
    \item[] Guidelines:
    \begin{itemize}
        \item The answer \answerNA{} means that the paper poses no such risks.
        \item Released models that have a high risk for misuse or dual-use should be released with necessary safeguards to allow for controlled use of the model, for example by requiring that users adhere to usage guidelines or restrictions to access the model or implementing safety filters. 
        \item Datasets that have been scraped from the Internet could pose safety risks. The authors should describe how they avoided releasing unsafe images.
        \item We recognize that providing effective safeguards is challenging, and many papers do not require this, but we encourage authors to take this into account and make a best faith effort.
    \end{itemize}

\item {\bf Licenses for existing assets}
    \item[] Question: Are the creators or original owners of assets (e.g., code, data, models), used in the paper, properly credited and are the license and terms of use explicitly mentioned and properly respected?
    \item[] Answer: \answerYes{} % Replace by \answerYes{}, \answerNo{}, or \answerNA{}.
    \item[] Justification: All datasets (Stocks, ETTh, Energy, fMRI) and baseline methods (DiffusionTS, TimeVQVAE, SDformer, TimeWak, KGW, Tree-Ring, Gaussian Shading) are cited with their original papers. Dataset sources are listed in Table 11.
    \item[] Guidelines:
    \begin{itemize}
        \item The answer \answerNA{} means that the paper does not use existing assets.
        \item The authors should cite the original paper that produced the code package or dataset.
        \item The authors should state which version of the asset is used and, if possible, include a URL.
        \item The name of the license (e.g., CC-BY 4.0) should be included for each asset.
        \item For scraped data from a particular source (e.g., website), the copyright and terms of service of that source should be provided.
        \item If assets are released, the license, copyright information, and terms of use in the package should be provided. For popular datasets, \url{paperswithcode.com/datasets} has curated licenses for some datasets. Their licensing guide can help determine the license of a dataset.
        \item For existing datasets that are re-packaged, both the original license and the license of the derived asset (if it has changed) should be provided.
        \item If this information is not available online, the authors are encouraged to reach out to the asset's creators.
    \end{itemize}

\item {\bf New assets}
    \item[] Question: Are new assets introduced in the paper well documented and is the documentation provided alongside the assets?
    \item[] Answer: \answerNA{} % Replace by \answerYes{}, \answerNo{}, or \answerNA{}.
    \item[] Justification: The paper does not release new public datasets or pretrained model checkpoints.
    \item[] Guidelines:
    \begin{itemize}
        \item The answer \answerNA{} means that the paper does not release new assets.
        \item Researchers should communicate the details of the dataset\slash code\slash model as part of their submissions via structured templates. This includes details about training, license, limitations, etc. 
        \item The paper should discuss whether and how consent was obtained from people whose asset is used.
        \item At submission time, remember to anonymize your assets (if applicable). You can either create an anonymized URL or include an anonymized zip file.
    \end{itemize}

\item {\bf Crowdsourcing and research with human subjects}
    \item[] Question: For crowdsourcing experiments and research with human subjects, does the paper include the full text of instructions given to participants and screenshots, if applicable, as well as details about compensation (if any)? 
    \item[] Answer: \answerNA{} % Replace by \answerYes{}, \answerNo{}, or \answerNA{}.
    \item[] Justification: The paper does not involve crowdsourcing or research with human subjects.
    \item[] Guidelines:
    \begin{itemize}
        \item The answer \answerNA{} means that the paper does not involve crowdsourcing nor research with human subjects.
        \item Including this information in the supplemental material is fine, but if the main contribution of the paper involves human subjects, then as much detail as possible should be included in the main paper. 
        \item According to the NeurIPS Code of Ethics, workers involved in data collection, curation, or other labor should be paid at least the minimum wage in the country of the data collector. 
    \end{itemize}

\item {\bf Institutional review board (IRB) approvals or equivalent for research with human subjects}
    \item[] Question: Does the paper describe potential risks incurred by study participants, whether such risks were disclosed to the subjects, and whether Institutional Review Board (IRB) approvals (or an equivalent approval/review based on the requirements of your country or institution) were obtained?
    \item[] Answer: \answerNA{} % Replace by \answerYes{}, \answerNo{}, or \answerNA{}.
    \item[] Justification: No human subjects research is conducted, so IRB approval is not applicable.
    \item[] Guidelines:
    \begin{itemize}
        \item The answer \answerNA{} means that the paper does not involve crowdsourcing nor research with human subjects.
        \item Depending on the country in which research is conducted, IRB approval (or equivalent) may be required for any human subjects research. If you obtained IRB approval, you should clearly state this in the paper. 
        \item We recognize that the procedures for this may vary significantly between institutions and locations, and we expect authors to adhere to the NeurIPS Code of Ethics and the guidelines for their institution. 
        \item For initial submissions, do not include any information that would break anonymity (if applicable), such as the institution conducting the review.
    \end{itemize}

\item {\bf Declaration of LLM usage}
    \item[] Question: Does the paper describe the usage of LLMs if it is an important, original, or non-standard component of the core methods in this research? Note that if the LLM is used only for writing, editing, or formatting purposes and does \emph{not} impact the core methodology, scientific rigor, or originality of the research, declaration is not required.
    %this research? 
    \item[] Answer: \answerNA{} % Replace by \answerYes{}, \answerNo{}, or \answerNA{}.
    \item[] Justification: LLMs are not used as an important, original, or non-standard component of the core methodology in this work.
    \item[] Guidelines:
    \begin{itemize}
        \item The answer \answerNA{} means that the core method development in this research does not involve LLMs as any important, original, or non-standard components.
        \item Please refer to our LLM policy in the NeurIPS handbook for what should or should not be described.
    \end{itemize}

\end{enumerate}